\documentclass[12pt]{article}
\usepackage[margin=1.0in]{geometry}
\usepackage{cite}
\usepackage{graphics}
\usepackage{graphicx}
\usepackage{amsmath}
\usepackage{mathtools}
\usepackage{amsfonts}
\usepackage{caption}
\usepackage{subcaption}
\usepackage{rotating}

\title{Distributed Machine Learning in Materials that Couple Sensing, Actuation, Computation and Communication}
\author{Dana Hughes and Nikolaus Correll \\ 
        Department of Computer Science \\
        University of Colorado Boulder}

\date{}

\begin{document}

\maketitle



\begin{abstract}

This paper reviews machine learning applications and approaches to detection, classification and control of intelligent materials and structures with embedded distributed computation elements.  The purpose of this survey is to identify desired tasks to be performed in each type of material or structure (e.g., damage detection in composites), identify and compare common approaches to learning such tasks, and investigate models and training paradigms used.  Machine learning approaches and common temporal features used in the domains of structural health monitoring, morphable aircraft, wearable computing and robotic skins are explored.  As the ultimate goal of this research is to incorporate the approaches described in this survey into a robotic material paradigm, the potential for adapting the computational models used in these applications, and corresponding training algorithms, to an amorphous network of computing nodes is considered.  Distributed versions of support vector machines, graphical models and mixture models developed in the field of wireless sensor networks are reviewed.  Potential areas of investigation, including possible architectures for incorporating machine learning into robotic nodes, training approaches, and the possibility of using deep learning approaches for automatic feature extraction, are discussed.

\end{abstract}

\section{Introduction}
Over the last few decades, there has been growing interest in extending the functionality of engineered materials, such as composites, cement-based materials, polymers and textiles, beyond that of a purely structural material.  From a purely materials perspective, a class of materials (referred to in the literature as ``functional materials'' or ``smart materials'') has emerged to describe materials which respond to some external stimuli.  Examples of such materials include piezoelectric materials, shape memory alloys, chromic materials, and polycaprolactone.  During the same time, microelectromechanical systems (MEMS) were developed, which resulted in a suite of sensors and actuators, including accelerometers, gyroscopes, pressure sensors, LEDs, ultrasonic transducers and microphones, which are on the millimeter scale.  Small, inexpensive components such as these has allowed for active sensing to be embedded into engineering materials, to assist with monitoring the local properties of a structure \cite{hughes2001method,hughes2003near,hughes2005-mst1,hughes2005-mst2}.  Microcontrollers and microprocessors have also become available on a similar scale.  Combining these with sensors and actuators can extend the capabilities materials to be programmed to respond in a desired manner to external stimulus.

The advancement of these constituent parts has resulted in several concepts which incorporate one or more of these parts into a new class of material.  ``Intelligent Materials'' was introduced in 1990 as ``materials that respond to environmental changes at the most optimum condition and manifest their own functions according to these changes'' \cite{takagi1990-intelligentmaterials}.  This approach is proposed as a natural extension to functional materials:  by incorporating information or software directly into the material, the material itself behaves intelligently, rather than simply being reactive to its environment.  

\begin{figure}[!htb]
  \centering
  \includegraphics[width=0.8\textwidth]{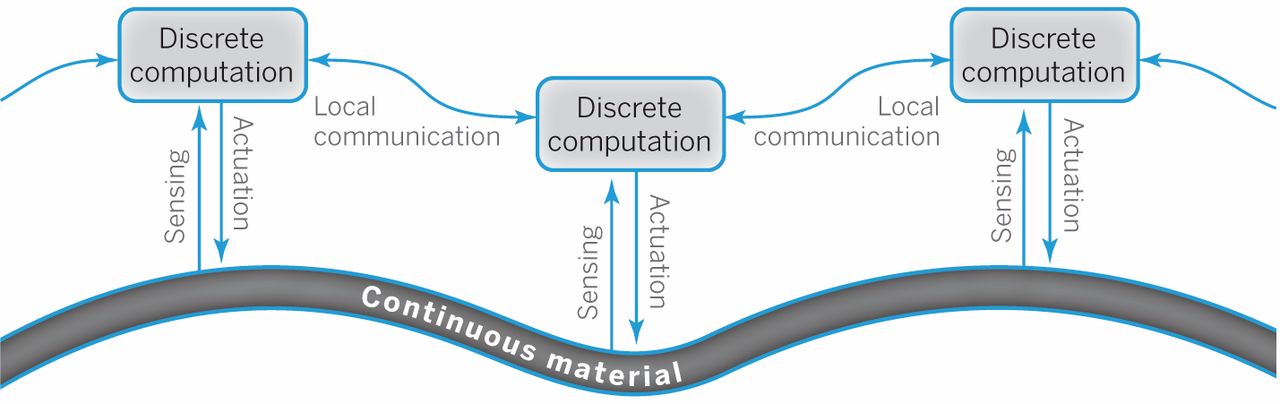}
  \caption{Relation between continuous physical material and computing network in robotic materials.  From \cite{mcevoy2015-science}).  Reprinted with permission from AAAS.}
  \label{fig:robotic_materials_continuous_discrete}
\end{figure}

Recently, ``robotic materials'' has been proposed as the combination of a network of sensing-computing-actuating elements with manufactured materials, such as polymers, composites, cement-based materials, and fabrics \cite{mcevoy2015-science}.  Robotic materials may be viewed as a continuous material, which responds to external physical stimuli, interacting with discrete elements, which can locally sense the state and locally modify some property of the material, as shown in Figure \ref{fig:robotic_materials_continuous_discrete}.  The ability to program the discrete elements allows robotic materials to approximate the goal of intelligent materials.  Several applications of robotic materials have already been explored, such as gesture recognition in architectural facades \cite{farrow2014-wall} and robotic skins \cite{hughes2015-affective_touch}, detection, localization and identification of tactile interaction with textured surfaces \cite{hughes2015-skin}, and garments capable of auditory scene monitoring \cite{profita2015-flutter}.

The above applications often assume a simplified model of the underlying physical process being measured.  In \cite{hughes2015-skin}, the skin is assumed to be a infinite thin plate of constant thickness, the analysis of which may not apply in the presence of connectors, joints or locations where the skin is excessively stretched.  Similarly, in \cite{profita2015-flutter}, determining which sound to respond to simply involves identifying when measured signal exceeds the ambient sound by some threshold, rather than attempting to identify specific sounds of interest (e.g., sirens, gun fire, etc.).  

Machine learning is one promising approach to developing intelligent behavior in robotic materials.  Broadly speaking, machine learning approaches attempt to learn a particular general task, such as identifying a pattern or learning a function, from a set of provided examples.  Typically, the physical response of a material to some stimulus, such as the vibrations generated from an impact, is only well defined for very simple structures (e.g., rectangular planar composite panels).  More complex geometries, such as those found in real world structures, require computationally intensive numerical modeling approaches, such as finite element methods (FEM) \cite{zienkiewicz2005finite}.  Machine learning approaches, in contrast, utilize a set of examples to \emph{learn} how a structure responds to an external stimulus from several sample stimuli.  For robotic materials, it is typically expected that both spatial and temporal aspects will be present in learning a task.

The goal of this paper is to explore the potential of using machine learning techniques for robotic material applications.  Specifically, this paper aims to address the following:

\begin{itemize}
  \item Identify specific fields and applications where robotic materials can be suitably utilized.
  \item Review the existing literature in these fields for examples where machine learning has been used for a particular task, identifying the features, models and learning approaches used.
  \item Investigate distributed machine learning algorithms and models, specifically those developed for sensor networks.
  \item Identify potential areas to investigate for intelligent robotic materials.  This includes determining appropriate models for data fusion and learning time series, automatic feature extraction, training issues, and the effects that limited memory, communication and computation abilities of individual nodes.
\end{itemize}

The remainder of this paper is organized in the following manner:  Section \ref{sec:robotic_materials} provides an overview of fields and concepts which motivated robotic materials, a detailed description of robotic materials, an reviews successful applications in this field.  Section \ref{sec:ml_robotic_materials} explores potential approaches to incorporating machine learning models and algorithms into a robotic material, specifically citing recent applications in the field of wireless sensor networks.  Section \ref{sec:ml_in_materials} surveys machine learning applications in fields which have been suggested as potential robotic materials, namely structural health monitoring, aerospace structures, wearable computing and robotic skins.  Section \ref{sec:discussion} discusses the potential research challenges to applying machine learning to robotic materials, and concludes in Section \ref{sec:conclusion}.  An overview of the machine learning models and features used are discussed in the appendix:  Appendix \ref{sec:appendix_machine_learning} reviews machine learning algorithms of interest for this paper, and Appendix \ref{sec:appendix_features} describes common features used in the literature reviewed.

\section{Overview of Robotic Materials}\label{sec:robotic_materials}

\subsection{Background}

Robotic materials are a recently introduced class of multi-functional materials which tightly integrate sensing, actuation, communication, and computation \cite{mcevoy2015-science}.  Robotic materials have been influenced by ideas presented in distributed MEMS \cite{berlin1997-distributedmems}, programmable matter \cite{goldstein2005-programmable_matter}, amorphous computation \cite{abelson2000-amorphous_computing}, morphological computation \cite{pfeifer2005-morphological_computation}, intelligent materials \cite{takagi1990-intelligentmaterials}, sensor networks \cite{akyildiz2002-sensornetworks} and pervasive intelligence \cite{servat2002-pervasive_intelligence}. 



The concept of distributed MEMS evolved from fabrication techniques which enable miniaturization and batch fabrication of sensors and actuators.  As the physical size and cost of sensors and actuators decrease, systems with thousands or millions of units become feasible.  Photolithographic processes provide a means to manufacture mechanical components in MEMS in conjunction with microelectronic components, which provide a means to control the resulting sensor and actuator arrays.  Conceptual applications include ``smart dust'', where MEMS sensors are diffused into an environment, active surfaces, where MEMS actuators influence the surrounding environment, and smart structures, where MEMS elements are fixed in place and interactions are coupled through the dynamics of the material to which they are attached.  

Programmable matter views miniaturization of computing components as a means of reproducing moving physical 3D objects, similar to how video and audio can reproduce sounds and moving images.  At the core, programmable matter consists of an ensemble of claytronic atoms (catoms), which are capable of moving in three dimensions, adhere to and communicate with other catoms, and compute state information.  Unlike microrobotics, it is unnecessary to make catoms completely autonomous; rather, catoms need only respond to a set of rules based on local state.  Shape formation becomes possible by incorporating holes in the ensemble, with random velocity, and enforcing shape goals at edge regions in the ensemble.

Amorphous computing is similar to the concept of programmable matter, though individual elements are influenced by biological cells.  One main concept in amorphous computing is that there is no underlying structure to the individual cells, and communication is not viewed as discrete events.  Rather, messages are diffused throughout the material, and individual cells respond to concentrations of received messages.  In this way, programs are designed such that state propagates through the material similar to wave propagation in physics, and based on markers within cells and local response rules, pattern formation is possible.

Morhpological computation attempts to shift intelligence in robotics from a concept embedded purely into computation to one in which the physical body and surrounding environment of the robot play some role in performing an intelligent action.  For example, perception for obstacle avoidance can be easily solved through non-homogeneous arrangement of light sensors, and fast and efficient locomotion, which is difficult to control using feedback control loops, is easily achieved through the interplay of simple oscillations, springed legs, a flexible spine and gravity.  This same concept is useful in robotic materials:  physical properties of a structure or material may be exploited to aid or perform certain types of computation.  For instance, the cross-section of a composite panel can be designed to be resonant at a particular frequency.  This lets the material act as a natural filter for vibration frequencies, which can simplify the signal analysis.

Intelligent Materials was introduced in 1990 as ``materials that respond to environmental changes at the most optimum condition and manifest their own functions according to these changes'' \cite{takagi1990-intelligentmaterials}.  This approach is proposed as a natural extension to functional materials:  by incorporating information or software directly into the material, the material itself behaves intelligently, rather than simply being reactive to its environment.  It is suggested that these types of materials be constructed from components at scale with the overall structure (e.g., drug delivery systems should have nano-scale computing and actuation elements, while structural materials can have meter-scale components) \cite{rogers1993-intelligentmaterials}, or computing, sensing and actuation components be directly incorporated into the material construction itself, using manufacturing processes used for electronic components.  From the perspective of information processing, holonic systems, neural networks and cellular automata have been suggested as computational models for intelligent materials \cite{yoshihito1994-informationprocessing}.

Sensor networks emerged in the late 1990s from the combination of advancements in wireless communication and a reduction in cost of electronics \cite{akyildiz2002-sensornetworks,rahman2010survey}.  Sensor networks consist of a large number of sensor nodes distributed in an environment, typically for monitoring some aspect of the environment, such as temperature.  Nodes in a sensor network are typically deployed densely in an environment, the topology changes frequently, and communicate through wireless broadcasting.  In contract, robotic materials assume a network with fixed topology, and communication between neighboring nodes.  The typical goal of sensor networks is to aggregate measurements and communicate the results to an external device through one or more sink nodes.  While machine learning approaches have been used with sensor networks for certain tasks, such as determining the location of a target in the environment, research is primarily concerned with system-level aspects, such as operating systems, communication protocols, routing and simulation frameworks.

The concept of pervasive intelligence provides a promising paradigm for incorporating intelligence into robotic materials \cite{servat2002-pervasive_intelligence}.  Pervasive intelligence argues for a combination of reactive multi-agent systems and amorphous computing:  the amorphous computing paradigm provides a suitable computing environment for which an ecosystem of reactive agents could evolve in a self-organized manner.  In the environment, communication between agents is not discrete; rather, agents can be viewed as receiving a continuum of information as it diffuses through the environment.  Based on this, agents are required to evolve or learn an appropriate response using few resources.  This paradigm shares several aspects with a robotic materials:  hypothetical agents in a robotic material (i.e., individual computing nodes) have limited resources to learn appropriate reactions, will potentially interact with a continuum of information from the physical dynamics of the underlying material, and communication between individual agents/nodes may be unreliable, based on communication bandwidth and possible node failure.  However, computing nodes are arranged in a lattice, which could provide some advantages over an amorphous medium:  nodes have a fixed number of neighboring agents which it can communicate with, rather than requiring an agent to accept messages from an arbitrary source.  Additionally, nodes located in a physically similar area of the material, such as would be the case with local neighbors, would be expected to react the same, and consequently could have the same learned behavior.

\subsection{Algorithmic Considerations}

There are several necessary considerations associated with algorithms implemented in a robotic material \cite{mcevoy2015-science}.  Algorithms must scale as the material grows in size.  As it is desired to keep the cost of individual nodes to a minimum, algorithms must be able to run on the limited computing capabilities and memory of the selected microcontroller.  Furthermore, algorithms must be robust to failure of individual nodes.  These considerations have many implications with regard to use machine learning approaches with robotic materials.

Scalability implies that any algorithm cannot have full access to all sensor data or all actuators in the system.  One approach is to ensure that nodes have limited support over the sensors, i.e., a learned task relies only on information gathered by nodes in a local neighborhood.  For many applications, it is reasonable to assume that information gathered from a local neighborhood of nodes would be sufficient for the task of interest.  For example, detecting the location of the source of a vibration requires only a small number of sensing nodes \cite{hughes2015-skin}.  The intensity of vibration due to an impact decreases as distance increases from the location of impact.  Consequently, nodes outside a sufficiently large enough radius from the source would not detect the impact, and only nodes within the radius need be used to calculate the location of the impact.

Computing and memory limitations place a limitation on the size or complexity of a machine learning model used on an individual sensing node.  For instance, while increasing the number of components in a GMM or the number of hidden units in a neural network may increase the representation power or classification accuracy of the model.  However, the increased number of parameters may exceed the memory capacity of an individual node.  This issue may at least be partially assuaged by the distributed nature of robotic materials.  Sensor nodes may reduce the amount of data collected locally, such as through filtering, transforms or encoding of the data.  In addition, as data is propagated through the network, a node can combine its measurements with data shared from neighboring nodes.

Finally, several possibilities exist to enable robustness to failure of individual nodes.  Data from a failed node can simply be treated as missing data in neighboring nodes.  Several techniques already exist for handling missing data \cite{garcia2010pattern}.  Furthermore, due to the spatial nature of robotic materials, the impact of missing data may not be as significant as in other machine learning tasks, as data from nearby nodes will likely be highly correlated with the data expected from the failed node.

\subsection{Applications}\label{sec:robotic_material_applications}

\begin{figure}[!htb]
  \centering
  \begin{subfigure}[b]{0.45\textwidth}
    \includegraphics[width=\textwidth]{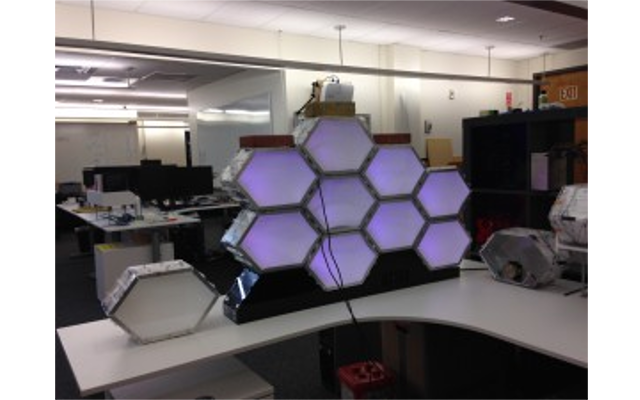}
    \caption{Smart Facade}
    \label{fig:robotic_material_applications_facade}
  \end{subfigure}
  \begin{subfigure}[b]{0.45\textwidth}
    \includegraphics[width=\textwidth]{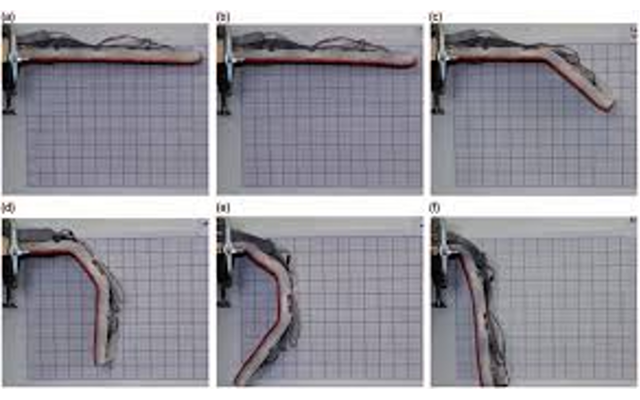}
    \caption{Morphable Beam}
    \label{fig:robotic_material_applications_morphable_beam}
  \end{subfigure}
  \begin{subfigure}[b]{0.45\textwidth}
    \includegraphics[width=\textwidth]{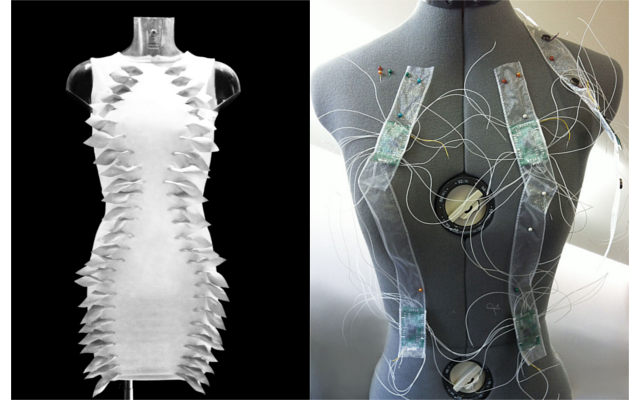}
    \caption{Flutter}
    \label{fig:robotic_material_applications_flutter}
  \end{subfigure}
  \begin{subfigure}[b]{0.45\textwidth}
    \includegraphics[width=\textwidth]{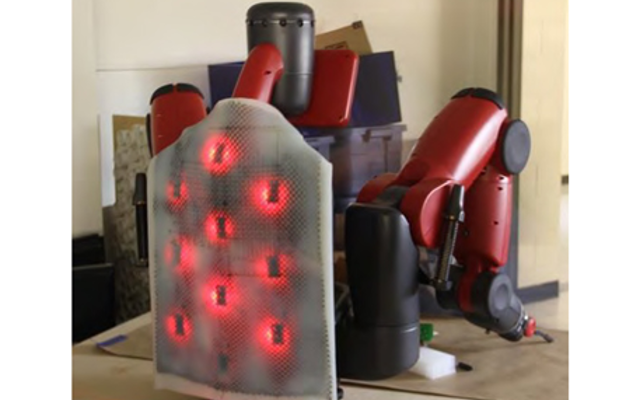}
    \caption{Robotic Skin}
    \label{fig:robotic_material_applications_skin}
  \end{subfigure}
  \caption{Robotic Materials.  b. \copyright 2015 SAGE Publications.  Reprinted, with permission, from \cite{mcevoy2014-variable_stiffness};  d. \copyright 2014 IEEE.  Reprinted, with permission, from \cite{hughes2014-skin}.}
  \label{fig:robotic_material_applications}
\end{figure}

Several recent applications across several disciplines have demonstrated the applicability of robotic materials, including smart facades \cite{farrow2014-wall, hosseinmardi2015-gesture}, variable stiffness materials \cite{mcevoy2014-variable_stiffness, mcevoy2014-shape_change, mcevoy2013-controllable_stiffness}, assistive garments \cite{profita2015-flutter} and robotic skin \cite{hughes2014-skin, hughes2015-skin}.  Each application highlights one or more aspects of robotic materials, and each demonstrate an application in the specific fields explored in Section \ref{sec:ml_in_materials}.  Prototypes of these applications are shown in Figure \ref{fig:robotic_material_applications}.

\subsubsection{Smart Facades}

Smart facades consist of configurable architectural components (i.e., ``smart bricks''), where each brick represents a node in an architectural robotic material, and bricks are assembled to form an architectural facade.  Each brick can detect environmental properties, such as ambient light or temperature, and can detect touch through capacitive sensing, allowing the facade to act as a user interface.  Bricks can respond to the environment or users by changing color or opacity, route air through the facade through a series of fans, or play sound (e.g., music).  Research on this material centers around developing touch interfaces, from menus and widgets \cite{farrow2014-wall} to recognizing letters from a set of predefined gestures \cite{hosseinmardi2015-gesture}.  Recognition of a gesture must be performed in a distributed manner, as touches span several individual blocks, and recognition algorithms must be robust to various scales of the gesture.  To identify a particular gesture, local neighborhoods of nodes interpolate a motion vector.  The sequence of motion vectors form a gesture chain, which can be used to calculate the nearest gesture in a dictionary using 1-nearest neighbor (1-NN).  This computation can be distributed throughout the network by distributing the dictionary throughout the network and performing 1-NN on the gestures in the local dataset, by performing 1-NN on the full dataset using only the local motion vector, or through a hybrid approach of the two.  This contribution demonstrates the utility of using local communication to reduce dimensionality of distributed measurements, as well as showing the capability of storing and computing over large datasets with limited node memory.  While this project does not specifically correlate to one of the applications of interest in this paper, it does demonstrate the suitability of robotic materials as civil and architectural structural components, and provides a physical platform for tasks such as health monitoring of civil structures.

\subsubsection{Variable Stiffness Composites}

Variable stiffness materials are composite materials consisting of a thermoplastic base material embedded in silicone rubber.  Individual portions of the material can be heated using nichrome heating wire wrapped around the base material, and local temperature can be monitored using thermistors \cite{mcevoy2013-controllable_stiffness}.  By creating the composite material from a lattice of variable stiffness bars and embedding computing nodes to monitor and control the local temperature, variable stiffness sheets and beams can be realized as a robotic material, where the local modulus of elasticity can be changed in response to the environment or for programmable configuration.  Performing a sequence of heating and cooling at specific portions of the beam or sheet enables changing the shape of the composite, either using gravity \cite{mcevoy2013-controllable_stiffness} or cables \cite{mcevoy2014-shape_change} to generate a distributed load on the structure.  Performing shape change in this manner provides a very high degree of freedom structure, which would normally require a prohibitive number of motors or other actuators to get equivalent behavior.  This application demonstrates the importance of understanding material and actuator properties, and the correspondence between the two, in robotic materials.  Shape-changing and variable stiffness materials have been proposed to be used as aerodynamic surfaces in cars, boats and airplanes, or for medical applications such as orthopedic casts \cite{mcevoy2014-variable_stiffness}.

\subsubsection{Textile Materials}

Flutter \cite{profita2015-flutter} is a prototype wearable assistive technology for the hearing impaired community.  Flutter is designed to augment hearing with vibrotactile feedback indicating the direction from which an alert, alarm, siren, or similar sound originates.  This garment is fabricated from textiles incorporating a network of computing nodes, each of which is able to detect sounds and control a vibration motor.  Locally, nodes continuously monitor the audio scene, calculating the spectrum of the scene and maintaining an estimate of the local ambient noise.  When a sound is detected which is sufficiently above the ambient levels (e.g., a siren), the energy of the spectrum is computed and compared with neighboring nodes.  In this way, individual nodes can determine if it has measured the highest spectral energy, and respond by pulsing its vibration motor.  This project shows that e-textiles are excellent candidates for robotic materials, specifically due to constraints on energy consumption and the size of components.

\subsubsection{Amorphous Robotic Skin}

Amorphous robotic skin is a final example of the potential of robotic materials \cite{hughes2014-skin, hughes2015-skin}.  The skin in this investigation is designed to localize tactile stimulation on the surface of the skin using a sparse network of sensing nodes, and identify the texture rubbed against the skin, by detecting vibrations propagating through the skin.  This architecture differs greatly from traditional skin approaches, where measurements from an array or matrix of densely packed sensors are performed by a central computer.  Vibration-based tactile sensing is very high bandwidth--sample rates for a single sensor are at least several kHz, whereas pressure-based sensing is sampled at most of up to several hundreds of Hz \cite{schmitz2011-tactile_sensors}.  Consequently, when using vibration as a sensing modality in robotic skin, the bandwidth of communication or computation limits the scalability of a centralized approach much more rapidly than when using pressure alone.  Local sensing nodes in the skin monitor vibrations and detects deviations from ambient vibrations in a similar manner to Flutter.  When a significantly large vibration is detected, the spectrum of the vibration is shared with local neighbors.  Spectral energy is used to elect a node to perform localization and identification of the source of the stimulus.  Using a plate model of the skin, gradient descent is used to estimate the position of the vibration source from the measurements.  A logistic regression model is used to identify which texture rubbed against the skin.  In this manner, the skin can provide information about interactions with the environment through sparse, event-driven communication, rather than continual monitoring of the individual sensors.  This approach shows the benefit of designing algorithms around the physical properties of the underlying material, and the capability of a robotic material to operate autonomously and interact with one or more external systems.


\section{Machine Learning Approaches for Robotic Materials}\label{sec:ml_robotic_materials}

The robotic material examples provided in Section \ref{sec:robotic_material_applications} describe several applications where local algorithms are typically designed with some simplifying assumption.  For example, in \cite{profita2015-flutter}, triggering the garment's feedback is based simply on sound levels which exceed some threshold of the ambient sound level, and in \cite{hughes2015-skin}, the underlying skin is modeled as a thin, homogeneous plate.  For several of these applications, it is preferable for the material to respond to a specific stimulus (e.g., a garment that responds to sirens and car horns), or one which is robust to variations in the underlying structure (e.g., fasteners present at various locations in the skin).  One approach to achieve this is to utilize machine learning to identify the specific patterns of interest, or to be able to adapt to local variations in the material.  This section describes machine learning approaches and features used in the examples in Section \ref{sec:robotic_material_applications} (and described in more detail in Section \ref{sec:ml_in_materials}), and describes how distributed implementations of several of these approaches have been achieved recently in the sensor networks domain.

\subsection{Machine Learning Algorithms}


Several machine learning approaches have been applied to the various fields described in Section \ref{sec:ml_in_materials}, as well as being adapted for use with sensor networks.  While specific tasks, models and feature selection vary among the applications, the desired tasks fall broadly into one of the following categories:  classification (identifying which class a set of measurements belongs to), regression (fitting a function to example input-output pairs), and novelty detection (identifying when a particular measurement occurs outside of a nominal mode of operation).  This section provides a brief overview of the machine learning approaches of interest; a complete description of these approaches are provided in Appendix \ref{sec:appendix_machine_learning}.  The models discussed here have been selected based on their how often they have been used in the reviewed literature, and their applicability for robotic materials.  For example, support vector machines and Bayesian models have been adapted for use with sensor networks, neural networks and Bayesian models have temporal implementations (RNNs, DBNs and HMMs), and convolutional neural networks are useful for automatically extracting robust features in time or spatial domains.

\subsubsection{Novelty Detection}\label{sec:anomaly_detection}

Novelty detection, also referred to as anomaly or outlier detection in the literature, is the task of identifying situations which occur outside of a nominal state or mode of operation.  Novelty detection has been utilized extensively in structural health monitoring to detect the presence of damage \cite{figueiredo2011-ml_damage_detection,worden1999-concordancy,worden2003-experiment1,figueiredo2011-ml_damage_detection}.  While a classification model could be trained to identify defective and non-defective cases, this approach is limited in its potential, as collecting data for the defective cases is often costly.  Consequently, it is more common to collect data under normal operating conditions, and determining if novel measurements lie outside some threshold of normal operation.  Impact detection and gait anomaly due to diseases such as Parkinson's are notable exceptions to this situation, as impacts can be performed on a structure without actually damaging the structure \cite{haywood2005-impact_monitor,leclerc2007-impact_detection,staszewski2000-impact_detection,sharif-khodaei2012-impact_location}, and gait data can be collected and diagnosed for individuals with and without a particular disease \cite{manap2011anomaly,manap2011anomalous,manap2013performance}.

Variation from a nominal state or operation can be measured as a novelty indicator (NI), also called a damage indicator (DI) in the structural health monitoring literature, a scalar measurement representing the deviation from normal operation \cite{figueiredo2011-ml_damage_detection}, based on extracted features.  Measurements can be based on the statistical distribution of feature sets from samples extracted from a training set, and relating the NI of a novel measurement to the likelihood that the measurement was generated from a nominal state.  Mahalanobis Squared Distance (MSD) assumes measurements from nominal states is accurately represented by a multivariate Gaussian distribution.  Kernel density estimate (KDE) is a more direct approach which estimates the probability density function directly from sampled data from a nominal state \cite{worden2003-experiment1}.  KDE generates a distribution as the sum of kernel functions centered around a set of training data points, where each kernel is smoothed by a bandwidth parameter.  

Novelty indicators may also be based on the residual error calculated from a novel measurement and the reconstruction of a model based on training data.  Autoencoder models can be trained to accurately reconstruct a set of training data from nominal states.  Factor analysis, which describes correlations between observed features and a small number of unobserved independent variables (factors), can be used to build models as linear combinations of the factors \cite{figueiredo2011-ml_damage_detection}.  Once models are constructed, novel measurements can be reconstructed from the models.  A new measurement would not likely be accurately reconstructed; the norm of the residual between the measurement and reconstruction can be used as a NI.  

Novelty indicators have also been computed by performing singular value decomposition (SVD) on a matrix of training values collected from a nominal state.  Assuming data from a nominal state is composed of a linear combination of a finite number of characteristic vectors, which represent particular properties of the structure, SVD can be used to detected novelty \cite{ruotolo1999using}.  Combining a set of training data into a training matrix, SVD can be used to calculate the rank of the training matrix (i.e., number of characteristic vectors), or, in the presence of additive noise, the rank can be estimates by only considering singular values above a predefined threshold.  Adding a measurement from a defective state to the training matrix, the presence of a defect would result in a novel characteristic vector in the matrix, increasing the estimated rank when a defect is present.  A novelty indicator which is robust to multiplicative noise can also be calculated, as described in Appendix \ref{sec:appendix_SVD}.  

\textbf{Calculation of Thresholds.}  In the approaches mentioned above, some threshold or critical value is required to distinguish between nominal and novel states.  This must be done using only data from nominal states, and often requires making some assumption about the underlying data.  A simple approach involves assuming novelty indicators follow a Gaussian distribution.  The NIs can be computed for each training point, and thresholds can be set such that a large, predefined percentage (i.e., confidence level) of the training data is within the threshold \cite{worden2002novelty,worden2003-experiment1,worden1999-concordancy}.

\subsubsection{Classification and Regression}

Classification is potentially the most common task performed in the domains investigated here, being used to classify types or broad locations of damage in structural health monitoring, classifying affective touch in robotic skin, and activity and gait recognition in wearable computing.  Regression is useful for localizing damage or impacts in structural health monitoring, or localizing stimulus in robotic skin.  Regression can also be used in control applications, such as morphable airfoils or exosuit control, for example, to approximate Q values in Q-Learning approaches, such as those used in \cite{tandale2004-preliminary_rl, valasek2005-rl, valasek2008-improved_rl,lampton2009-rl, lampton2010-rl}.  While technically two separate tasks, the same type of models can be used for both regression and classification by simply modifying a small aspect of the model or the training approach.

\textbf{Neural Networks.}  Artifical neural networks are common approaches to perform classification and regression in the literature cited \cite{song2006-shm_ica_svm,seo1999-cfrp_nn,oh2009-damage_svm,haywood2005-impact_monitor,staszewski2000-impact_detection,leclerc2007-impact_detection,sharif-khodaei2012-impact_location,manson2003experimental3,elkordy1993-neuralnetwork,flagg2013affective,van2014neural,manap2011anomaly,gagnon2013qualitative,dutta2009automated,manap2011statistical,vallejo2013artificial}.  Artificial neural networks are computational models inspired by the behavior of biological neurons.  Neural networks consist of layers of neurons and directed connections between them, as shown in Figure \ref{fig:neural_network}.  By selecting appropriate activation functions for the output layers, neural networks can be used for regression (e.g., using rectified linear units) or classification (e.g., using a softmax layer).  Training involves minimizing some cost function with respect to the connection weights and a set of training examples, typically using backpropagation \cite{werbos1974-backprop, rumelhart1986-backprop}.  Beyond use as a machine learning approach, three particular network architectures are of interest for the applications described in this paper:  recurrent networks, convolutional networks,  and autoencoders.


Recurrent Neural Networks (RNNs) consist of networks which contain cycles, which provide models which are more suitable for sequential or time-series models.  Recurrent connections allow the network to maintain and update an internal state at each step in the sequence, providing a suitable architecture to handle temporal aspects of a signal.  Long Short Term Memory (LSTM) \cite{hochreiter1997-LSTM} components are specifically useful for situations where there is a long time delay between input to the network and the corresponding effect on the output, where simple recurrent connections could not learn a relationship with such a large time delay \cite{hochreiter2001-rnngradient}.

Convolutional layers in neural networks utilize a weight sharing scheme to allow a network to learn local features which are invariant to translation and scaling, and have been shown to be especially useful for image recognition \cite{lecun1995-cnn}.  Convolutional layers take advantage of local structure of input data, which allows convolutional networks to automatically learn features of interest from training data, rather than requiring hand-developed features.  Despite the potential to automatically learn features of interest, there has only been a few investigations in gait recognition to utilize this approach\cite{ordonez2016deep,yang2015deep,zeng2014convolutional}.

Autoencoders are networks which have found several applications in the domains of interest.  Autoencoders are built and trained to perform the identity function--the output of the network is trained to reproduce the provided input.  A ``code'' layer of neurons, whose size is smaller than the input, provides an means of reducing the dimensionality of the data.  In this way, autoencoders behave in the same way as principle components analysis (PCA).  Because of the nonlinearity of the architecture, autoencoders with a single hidden layer have been shown to provide superior reconstruction of the input signal, when compared with principle components analysis (PCA) \cite{japkowicz2000-nonlinear}.  Finding a minimal representation directly from the data is useful to reduce the dimensionality of observations for dynamic Bayesian networks and HMMs, making learning more feasible \cite{hughes2015-affective_touch}.  Finally, autoencoders can be used to perform novelty detection, as discussed in Section \ref{sec:anomaly_detection}.

\textbf{Support Vector Machines.}  Like neural networks, Support Vector Machines (SVMs) have been extensively used for classification and regression tasks in the literature cited \cite{worden2007-ml_shm,song2006-shm_ica_svm,mechbal2015-probabilistic_svm,worden2001-damage_identification,jung2014touching,ta2015grenoble,kaboli2015humanoids,kaboli2014humanoids,barron2013discrimination,manap2011anomalous}.  In the original formulation, an SVM is a binary classifier trained on a linearly separable dataset, which is constructed by finding an optimal hyperplane which separates the two classes in the dataset \cite{cortes1995-svm}.  A subset of the training data, known as \emph{support vectors}, is used to determine the hyperplane by maximizing the margin between the hyperplane and the support vectors.  Nonlinear separation is possible by using nonlinear kernel functions to map the original features to a kernel space, and performing linear separation on the mapped features.  Regression can be performed by constructing a support vector regression (SVR) machine; the problem formulation is modified to minimize the distance between training examples and the optimal hyperplane \cite{smola2004-tutorial_svr}; in essence, the goal is simply to attempt to keep the training data points within the bounds of the margins, as opposed to outside of the margins. 

\begin{figure}[!htb]
  \centering
    \includegraphics[width=0.8\textwidth]{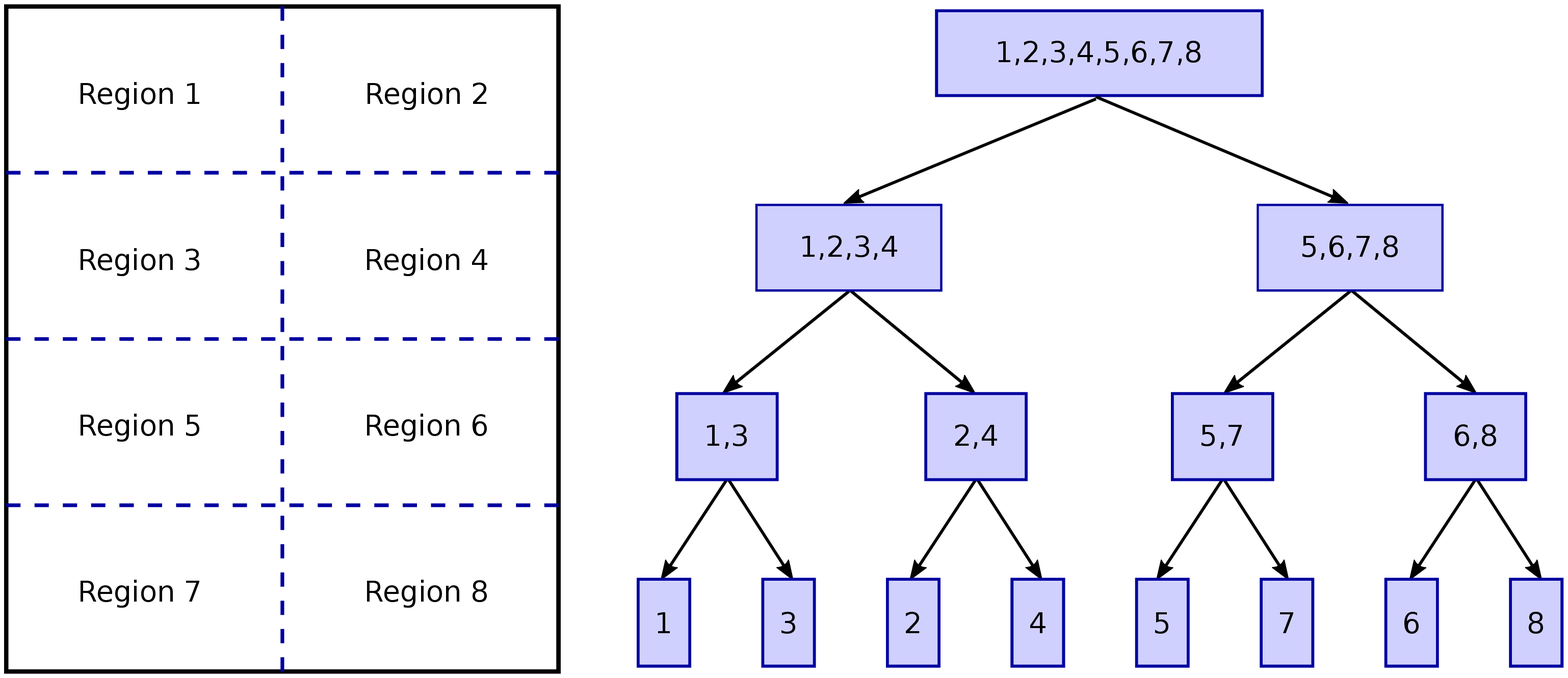}
    \caption{Division of a structure and associated SVM-GPDT classifier, as used in \cite{mechbal2015-probabilistic_svm}.}
    \label{fig:svm-gpdt}
\end{figure}

Most investigations require the ability to perform classification with several classes; several approaches are available to extend SVMs to multi-class classification.  Multiple SVM classifiers can be build, each of which classifies between one of the labels and the rest (\emph{one-vs-all}), or by classifying between pairs of classes (\emph{one-vs-one}), and selecting the final label based on the classifier with the highest output, or by selecting the label based on the highest number of output from the group of classifiers, respectively.  One approach specifically well suited to spatially-dependent classification is SVM-based geometric probabilistic decision trees (SVM-GPDT) \cite{mechbal2015-probabilistic_svm}.  The classifier consists of a tree, where each node contains an SVM which is used to determine the probability that the region represented by the node is classified as positive; The children of each node divide the parent's region into two subregions, as shown in Figure \ref{fig:svm-gpdt}.  The influence parent nodes have on children classifiers results implies than, even in the event of a misclassification, the region with a positive classification would be a neighboring region.

Geometric probabilistic decision could be a particularly promising approach for robotic materials.  As individual leafs of the PDT correspond to specific geometric regions to be classified; the classifier for a given leaf can be implemented on the node corresponding to its region.  Nodes may then communicate with its appropriate neighbor to match the PDT, until the root classification is performed at some specific node.  While this approach distributed computation appropriately among the various nodes, care must be taken to ensure that communication is performed with appropriate nodes.  Additionally, classifiers are required at each level of the tree, so additional nodes will need to be incorporated into the material, or multiple classifiers will need to be implemented on each node, requiring careful assignment of classifiers in the network to ensure no node is responsible for an excessive number of classifiers.

\subsubsection{Density Estimation}

Density estimation approaches estimate a probability density function based on observations.  The probability density function can be used for the applications described above:  the PDF can be used for novelty detection by calculating the probability of observing a particular measurement.  Classification can be performed by including the desired class as a random variable, and performing classification by computing the probability of each class conditioned on a set of observations.  Finally, regression is performed by calculating the expectation of a random variable conditioned on observations.

\textbf{Graphical Models.}  Graphical models are used to describe, learn and perform inference over a multivariate probability distribution \cite{koller2009probabilistic}.  Graphical models include Bayesian networks, which represent the conditional distribution over several random variables as a directed acyclic graph, and Markov networks, which represent random variables as an undirected graph.  The utility of Bayesian networks for robotic materials may be minimal, as the conditional dependences of information of all the nodes are not easily defined.  Markov networks, however, map well to robotic materials, as individual nodes in a material can store and process a clique of local random variables, and messages passed between nodes can be used to perform belief propagation.  Some investigations into distributed learning and inference for Markov networks have been explored in the sensor network literature; see Section ~\ref{sec:sensor_network_pgm}.

A commonly used graphical model is the Na\"{i}ve Bayes model, which are used for classification purposes.  Na\"{i}ve Bayes models are simple models where observations are conditionally independent given a particular class.  The simplicity of these models have made them very attractive for classification tasks, and has been used in several investigations as simple classification models \cite{mannini2010-activity,kerdegari2012-falldetection,ravi2005-activity,luvstrek2009_fall-detection}.  

\textbf{Mixture Models.}  Mixtures models are probabilistic models for representing the distribution of data \cite{russell2009artificial}.  Mixture models consist of $K$ components, where each component is a parametric probability distribution (e.g., Gaussian), and each component has a corresponding mixture weight, i.e. the probability of the particular component, $P(c)$.  Gaussian mixture models (GMMs), where multivariate Gaussians are used for each component, have been used for classification though soft assignment to each category \cite{mannini2010-activity,allen2006-gmm}, or used to detect anomolies or outliers \cite{worden2003-experiment1,sohn2002statistical} based on the likelihood of an observation being generated by the mixture.  

\textbf{Dynamic Bayesian Network.}  A dynamic Bayesian network (DBN) is a class of graphical models which are useful for modeling sequential or temporal data \cite{russell2009artificial}.  DBMs represent a first-order Markov process, where the state variables at a particular time step are only dependent on the state variables in the previous time step, and observation variables are dependent on the current state variables.  Hidden Markov models (HMMs), which have been used for gait recognition \cite{taborri2014-gait_phase} and recognition of affective touch \cite{hughes2015-affective_touch}, are a special case of DBNs which consist of a discrete, hidden state variable and observations conditioned on the current hidden state \cite{rabiner1986introduction}.  HMMs may be build for symbolic or continuous valued observations; in the latter case, observations are assumed to be drawn from some probability density, such as a Gaussian or mixture of Gaussians.  Classification using HMMs involves training a model for each of the target classes, and selecting the class corresponding to the model with the highest likelihood of the sequence of observations.  Like RNNs, DBNs provide a means to consider the temporal aspects of a signal when performing classification or regression, and often produce improved results when compared to systems considering only single frames of data.


\subsection{Features}\label{sec:features}

The major focus of many of the papers referenced in Section \ref{sec:ml_in_materials} is the features extracted from sensors.  As most of the data in the applications described are temporal in nature, there is a high amount of overlap in the features extracted and used--features are typically based on a certain time frame of measured values, rather than a fixed set of predefined characteristics.  There have also been several features of note developed and used for one of the specific domains, for example, transmissibility features in structural health monitoring.  Most of the machine learning tools used require a single vector of features to be provided as input, rather than a sequence of measured values.  Thus, it is common to calculate a set of features by summarizing the properties of measurements over a sliding window (i.e., measurement frame) of relatively short duration.

Features can be broadly categories on how the data points in a frame are processed.  Time domain features perform calculations directly on the data points.  Frequency domain features first perform spectral decomposition, such as a Fourier or wavelet transform, and compute features from the resulting spectrum.  Model based features rely on fitting measurements to some simple model (e.g., autoregressive model), from which features can be extracted based on the parameters of the model or the residual of the measurement.

\subsubsection{Time Domain Features}

Time domain features summarize a measurement frame by performing operations directly on the measurement data in the frame.  Simple statistical features, such as mean, standard deviation and higher-order moments, are commonly used, especially for activity recognition \cite{casale2011-activity,ravi2005-activity,long2009-activity}, and to a lesser extent, impact detection \cite{haywood2005-impact_monitor, staszewski2000-impact_detection}.  In addition to statistical moments, properties such as the minimum, maximum, difference between ordered pairs of peaks (minmax sum) for a measurement window are often used.

Correlation coefficients can be calculated from multiple sensors, assuming the sensor measurements are synchronized.  This can include multiple measurements from the same sensor, such as the three axes of an accelerometer.  Correlation coefficients simply measure the covariance between the two signals, and provide a means for estimating whether the two signals are related by some underlying cause.  Computing correlation coefficients between two signals greatly reduce the dimensionality of the data--the number of samples in both windows are reduced to a single value.  In the wearable domain, correlation coefficients can be useful for identifying activities where motion involves various body parts possibly working in conjunction, such as the motion of the left and right leg when walking or running \cite{mannini2011-accelerometry}.

Certain signals, such as those generated by impacts, are limited in duration.  This allows for various features based on the amount of duration between the onset of the signal and various aspects of the signal, or the delay between the occurrence of specific aspects of signals measured by different sensors.  For example, for detecting impacts in aircraft composite panels, the time at the maximum and minimum value of the both the raw signal and the envelope of the signal, as well as the beginning and end of each envelope, is used as features \cite{haywood2005-impact_monitor, staszewski2000-impact_detection}.  In this investigation, sensor measurements are synchronized, and the onset time of each signal from 12 accelerometers was also used as a feature.  As synchronization is necessary for features based on the time delay between signals from multiple sensors, such features can only be feasibly calculated from signals measured on a single node; variation in clock rates and communication delay introduces unknown deviations from true values of time delays.

Time domain feature have been shown to be very effective in the wearable domain.  In \cite{casale2011-activity}, a 319 feature vector is calculated for 1 second accelerometer data, which include of the first four statistical moments (mean, standard deviation, skewness and kurtosis) of each axis, correlation between accelerometer axis, and minmax sums.  Additionally, the energy level at different frequencies were calculated using a wavelet transform.  Using Random Forest, the 20 features with the highest importance were low-level time domain statistics, namely, the mean, standard deviation, minmax sum, and the RMS velocity (calculated by integrating accelerometer measurements in the data) of each axis, demonstrating the utility of these types of features.  Mean, standard deviation and correlation are the primary features used in \cite{ravi2005-activity}, augmented only by the energy calculated from the frequency components of the signal.  To some extent, time domain components are used in conjunction with other types of features in several studies in the wearable domain \cite{long2009-activity}.  Minimum and maximum values of measured signals have also been used for impact detection \cite{haywood2005-impact_monitor, staszewski2000-impact_detection}.

\subsection{Frequency Domain Features}

Frequency domain features are extracted from signals which have undergone a time-frequency transform, such as a Short Time Fourier Transform (STFT) or Discrete Wavelet Transforms (DWT) \cite{mannini2010-physical_activity}.  Frequency components provide information regarding the amount of energy present at different frequencies in the signal, allowing for a more intuitive representation of periodic signals.  The spectral components calculated by the transforms may be used as features directly, or features may be calculated from the spectrum.  The spectral energy can be calculated as the sum of the squared spectrogram coefficients.  Omitting the DC component, the energy of a spectrogram may help assess the intensity of an activity in the wearable domain \cite{mannini2011-accelerometry}.  Integrating the real and imaginary components separately may also be reasonable in situations where the time delay between multiple sensors can provide useful information, such as in impact detection \cite{haywood2005-impact_monitor, staszewski2000-impact_detection}.  While all frequencies present in the spectrum of a signal are useful, the DC component of the signal, which can be obtained directly from the STFT, can be useful in the wearable computing domain; the DC components of accelerometer values is helpful for determining static postures in activity recognition \cite{mannini2011-accelerometry}.  Finally, spectral entropy, which simply calculates the entropy present in the energy of the spectrogram, has found use for distinguishing between cyclic and acyclic activities \cite{long2009-activity}.

Frequency domain approaches are very common in structural health monitoring, as they easily measure the various modes of vibration present in the structure.  Modal features are useful for detecting defects in structures \cite{kessler2007-patternrecognition, farrar2007-shm}.  Vibrations can be measured using a piezoelectric transducer or accelerometer, and exited from external sources, or directly from a transducer.  The vibrational behavior of structural components is not constant across all frequencies; rather, structures tend to vibrate according to a set of vibration modes, which gives rise to resonant frequencies associated with the structure.  Under normal operating conditions, a structure will demonstrate peaks in the energy spectrum at a set of fixed frequencies, representing the frequencies of the modes of vibration in the structure.  In the presence of a defect, the frequencies at which these peaks will shift.  While useful for detecting defects, they are less useful for localizing or determining the extent of a defect.

A second common feature in structural health monitoring related to vibration of the structure is transmissibility, due to its sensitivity to changes in the stiffness of a component \cite{montalvao2006-review, lang2011transmissibility}.  Transmissibility describes the propagation of a signal between two points on a structure, and are useful for monitoring the natural frequencies or modalities of the structure.  Transmissibility can be measured using piezoelectric elements \cite{worden2007-ml_shm}.  The required excitation needs to be over a range of frequencies.  Excitation can either be generated from an external source, or generated from one transducer and measured from another.  As transmissibility measurements are between multiple measurement points, they provide more information than simply detecting resonant frequencies, and can be useful for localization of defects \cite{worden2007-ml_shm}.  These features have been used to detect defects in aircraft wings \cite{worden2007-ml_shm} and truss structure \cite{worden2001-damage_identification}.

\subsection{Model Based Features}

Model based features involve calculating the coefficients of a model (e.g., autoregressive model) fit to the data in a measurement frame.  Models provide several benefits.  For example, dimensionality of the data can be reduced (e.g., principle components), and linear models (e.g., autoregressive models) can be exploited to detect nonlinearities in a signal.  For the purposes of this paper, this includes common data reduction methods in machine learning, such as principle components analysis.  

Principle components analysis (PCA) and independent components analysis (ICA) are models which are used to map measured signals to a new basis.  In PCA, an orthonormal transformation is perform which results in signals being linearly mapped to a set of uncorrelated components \cite{de2003principal}.  In ICA, signals are also mapped to a set of statistically independent components, though the components observe a non-Gaussian distribution \cite{song2006-shm_ica_svm}.  Both of these approaches are useful for reducing the dimensionality of data, assuming the number of components are fewer than the dimensionality of the original signal, and the mapping does not discard a significant amount of information.  In addition to PCA is also useful for detecting damage in structures.  In \cite{de2003principal}, PCA is used to detect defects in a vibrating plate by reconstructing a measurement using principal components, and calculating the magnitude of the error between the original and reconstructed signals.  This approach is described in more detail in Section \ref{sec:anomaly_detection} as damage indicators based on Singular Value Decomposition.

Autoregressive (AR) models are models which predict the value of a signal based on a linear combination of a fixed history of prior values.  The coefficients of the resulting model may be used as features, though this requires refitting an AR model at each measurement frame.  This type of feature has been used extensively in structural health monitoring \cite{sohn2002statistical, oh2009-damage_svm}, due to their sensitivity to damage, especially damage which results in nonlinear behavior in the signal, and robustness to environmental variability.  Alternatively, an AR model can be trained, and the residual used for novelty detection; see Section \ref{sec:anomaly_detection}.

Domain specific features are commonly used in the wearable domain.   These typically are kinetic and kinematic features, and are utilized for detecting anomalous gait or controlling exosuit activations.  Kinetic features are based on contact with the ground at different phases of the gait cycle.  Heel strike force, push-off force and mid-stance force are included in a feature set for novelty detection in \cite{manap2011anomaly}.  Kinematic features are based on joint angle (e.g., ankle, knee and hip), either measured at different phases of gait, or by measuring the angle at maximum extension and flexion of the joints.  Kinetic features, namely the occurrence of heel strike, are the primary features used for exosuit control \cite{asbeck2015soft,wehner2013lightweight,asbeck2015multi}--actuation occurs at specific phases of a gait cycle, which are estimated from the frequency of the gait and a time delay from last heel strike.


\subsection{Distributed Approaches in Sensor Networks}\label{sec:distribute_approaches}

\begin{figure}[!htb]
  \centering
  \begin{subfigure}[b]{0.33\textwidth}
    \includegraphics[width=\textwidth]{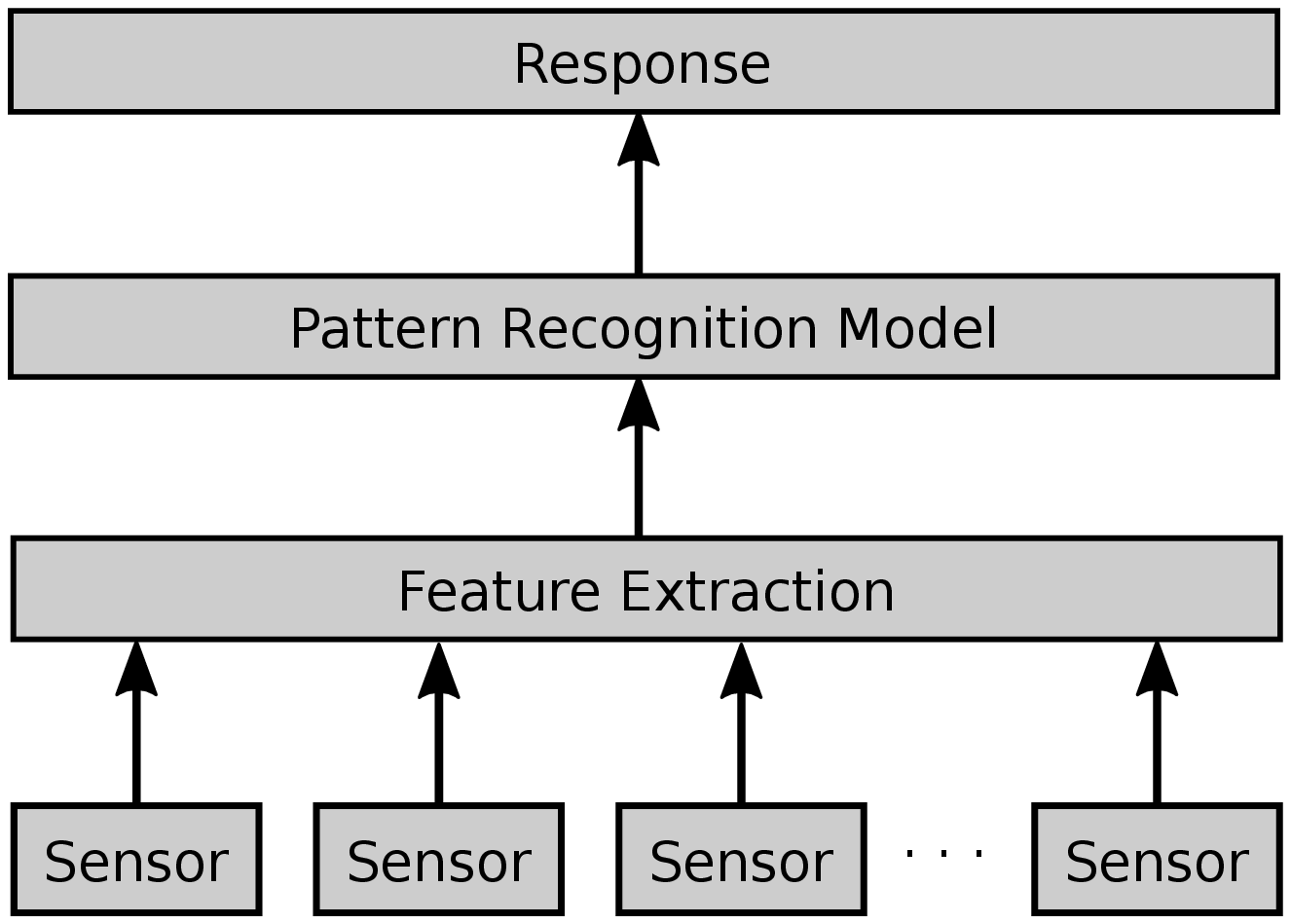}
    \caption{Centralized pattern recognition model}
    \label{fig:ml_model_centralized}
  \end{subfigure}
  \begin{subfigure}[b]{0.6\textwidth}
    \includegraphics[width=\textwidth]{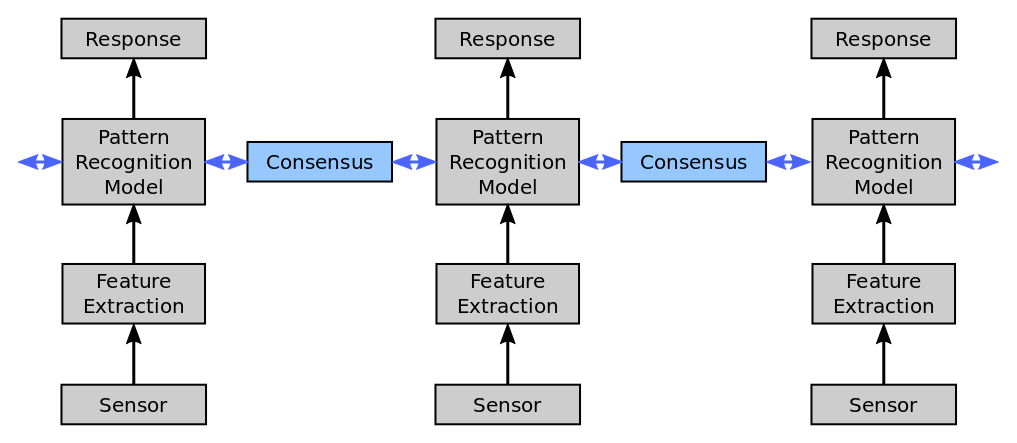}
    \caption{Pattern recognition in sensor networks.}
    \label{fig:ml_model_sensor_network}
  \end{subfigure}
  \caption{Comparison between centralized pattern recognition approaches and pattern recognition in sensor networks.}
  \label{fig:ml_models}
\end{figure}

The general machine learning approaches described above are used in various investigations discussed in Section \ref{sec:ml_in_materials} to solve a variety of tasks in the domains of interest (structural health monitoring, aerospace, wearable computing and robotic skin).  Adapting machine learning applications in the domains discussed requires adapting models to work with a network of computing nodes.  In a centralized approach, values from all sensors are collected, from which features are extracted, as shown in Figure \ref{fig:ml_model_centralized}.  From the calculated features, the desired response (e.g., classification, actuation, etc.) can be calculated using an appropriate model.  Sequential data can either be handled in a static or quasi-static manner, or can utilize a temporal model.  

For robotic materials, this approach becomes infeasible for a variety of reasons.  As the material increases in size, the number of sensors and actuators, and consequently the communication requirements, increases with the size of the material.  Ultimately, this results in a communication bottleneck which could reduce the responsiveness of the material to an unusable level.  This issue is compounded by high-bandwidth signals, such as when measuring vibrations using accelerometers, microphones or piezoelectric transducers.

An alternative approach is to perform feature extraction locally, and alter the machine learning model to incorporate a consensus algorithm or to communicate local state between neighboring nodes.  Figure \ref{fig:ml_model_sensor_network} demonstrate how the approaches in Figure \ref{fig:ml_model_centralized} could be adapted for this.  Approaches typically involve utilizing consensus algorithms or message passing between nodes.  These approaches have been explored in the sensor network community.

The applications described in Section \ref{sec:ml_in_materials} make the assumption that a global understanding of the system is available, i.e., at any point in time, all sensor measurements are available and can be queried in a synchronized manner.  Furthermore, it is assumed that any computation is performed in a centralized manner:  the learning algorithm used to train a model has access to all sensor measurements at any given time.  As described in Section \ref{sec:robotic_materials}, this approach does not scale well as the size of the material / structure increases, or as sensor density increases.  In addition to the communication and computation bottleneck, from a machine learning perspective, increasing the number of sensor measurements quickly leads to curse of dimensionality issues.  To effectively train a centralized model, a prohibitively large amount of training data would be required.  Projecting the machine learning tasks into the framework of robotic materials can alleviate these issues, however, this requires that a learning algorithm be capable of being evaluated in a distributed manner, and ideally also trained in a distributed manner.

This section provides a survey of machine learning techniques used in the field of sensor networks, where sensing nodes have a known location, and local (typically wireless) communication with other sensing nodes.  Robotic materials have similar constraints, though communication is wired and thus more constrained, and location and network topology is less variable.  Machine learning approaches, specifically supervised learning, was first proposed as a unified approach to sensor network applications in 2003 \cite{simic2003-learning}.  The authors suggest kernel methods as an example approach to fit training data to, and suggest plume tracking, object tracking and node localization as potential sensor network specific applications.  Distributed machine learning approaches used in sensor networks include algorithms to train and query support vector machines for classification and regression (Section \ref{sec:sensor_network_svm}), density estimation in Gaussian mixture models (Section \ref{sec:sensor_network_gmm}), and inferences using probabilistic graphical models (Section \ref{sec:sensor_network_pgm}).

\subsubsection{Distributed Support Vector Machines}\label{sec:sensor_network_svm}

Support vector machines typically follow the suggestions presented in \cite{simic2003-learning}, using kernel functions which reflect the limited sensing range of nodes in the network.

A localization scheme is presented in \cite{nguyen2005-kernel_sensor_network}, which describes an approach to determining the location of all sensor nodes in a network when the location of only a small subset of base nodes is initially known.  Both coarse-grained and fine-grained localization of a node is performed, based on the assumption that signals between two nodes decay in a manner similar to Gaussian, polynomial or other kernel with bounded support.  To perform course-grained localization, an SVM is trained to indicate if a node is in an arbitrarily defined region, using the base nodes as training examples.  Once the SVMs are trained, fine-grained localization can be performed.  This consists of performing coarse-grained localization over several regions using the trained classifiers, indicating which regions the node of interest occupies.  The final location is set as the center of the intersection of the region.  As the shape and size of training regions can be arbitrarily set for each SVM, it is possible to achieve a high level of accuracy for fine grained localization.  The main drawback to this algorithm is the need for training at a central station; the number of required values communicated to the central station is quadratic with respect to the number of base nodes.  Once trained, evaluating the course localization using the trained model can be performed in linear or quadratic time with respect to the number of sensor nodes.

The approach given in \cite{nguyen2005-kernel_sensor_network} assumes that the central station can estimate the range to \emph{all} base nodes in the network.  For networks over a wide area, this assumption may not hold.  One approach which does not rely on this assumption is presented in \cite{tran2008localization}.  In this investigation, each beacon node exchanges messages with all other beacon nodes, noting the hop-count distance between the nodes.  The location of each node and hop-count distance of each pair is then sent to the central station.  An SVM-based decision tree, similar to that in \cite{mechbal2015-probabilistic_svm}, is then trained to predict the location of a node based on the hop count distance to beacon nodes.  The SVM model can then be broadcast to all nodes in the network, which non-beacon nodes can then use to estimate their physical location in the network.  The SVM approach is shown to produce location estimates with smaller error than a diffusion-based approache to localization (an approach which has similar assumptions and requirements), while requiring much less communication.

A distributed algorithm for performing regression is presented in \cite{guestrin2004-distributedregression}, where regression model consisting of a linear combination of local measurements over a window of time.  Kernel linear regression is used to minimize the error between the regression model and measured data.  By ensuring that the kernel functions have limited spatial support, the task can be distributed by solving the regression problem for the local region, and communicating kernel weights (i.e., ``regression messages'') with neighboring nodes, which greatly reduces the amount of communication compared with methods that communicate individual measurements or require a centralized training approach.

Algorithms for fully-distributed support vector machine learning has been explored recently in the field of wireless sensor networks.  This includes both training a parallel version of an SVM, with a centralized means of fusing subsets of the training results, as well as fully distributed approaches.  For sensor networks, distributed SVM algorithms are achieved by using kernels with limited support (i.e., sensor measurements only consider events within a certain radius) \cite{gu2008-svr}, by locally summarizing training data to a minimal representation \cite{kim2015-svm}, by performing classification / regression on a subset of measurements in a local neighborhood \cite{kim2014-svdd, kim2013-svr}, and by implementing convergence algorithms \cite{kim2013-svr, gu2008-svr}.

In \cite{gu2008-svr}, algorithms for training and evaluating a support vector regression model is realized by utilizing kernels with finite support and a gradient-based training algorithm.  This approach is based on the assumption that sensor measurements from nodes that are spatially near one another are often highly correlated, while those separated by larger distances are not.  By using a translation-invariant kernel with finite support, and assuming communication ranges with are wider than the support, the authors developed gradient-based SVR training and evaluation algorithms.  The learning algorithm trains a regression function for $N$ sensors, each located at $q_i = (x_i, y_i)$, of the form

\begin{equation}
f(q) = \sum_{i=1}^N (\alpha_i - \alpha_i^*) [K(q_i, q) + \lambda^2]
\end{equation}

The kernel used is a bump function with finite support $B$,

\begin{equation}
K(q_i, q_j) = \frac{1}{2} \left[ 1 + cos \left( \frac{\pi \lVert q_i - q_j \rVert}{B} \right) \right]
\end{equation}

Using this kernel, training can be performed in a parallel manner over local neighborhoods of sensing nodes, where the neighborhood radius is at least the size of the support of the kernel.  Rather than using the full training set, as in the centralized case, each node trains a local SVR based on data available from its local neighborhood.  Each node in the neighborhood shares current training data with neighboring nodes.  Using the shared data and the distances between a node and its neighbors, the local SVR is updated in the same manner as a centralized SVR.  This process is repeated until terminal conditions are met.

To evaluate the function at point $q$, a node first performs a local evaluation of the function

\begin{equation}
f_i(q) = (\alpha_i - \alpha_i^*) [K(q_i, q) + \lambda^2]
\end{equation}

Nodes then iteratively update their estimate by obtaining the approximate function evaluation from its neighbors

\begin{equation}
f_i(q) \leftarrow f_i(q) + \eta \left[ \sum_{j \in \mathbb{N}_i} (f_j(q) - f_i(q)) \right]
\end{equation}

Assuming the sensor network is connected, the consensus algorithm is guaranteed to be globally asymptotically $\epsilon$ stable.

In \cite{kim2015-svm}, a fully distributed algorithm for training an SVM is described which leverages a geometric interpretation of SVMs to minimize memory and communication requirements, and trains a distributed SVM using a gossip protocol.  The geometric interpretation assumes that training examples from a class form a convex set, and that a set of extreme points (which describe the hull of the convex set) can be extracted from this set.  Training examples are spatially dependent, i.e., the position of the measuring sensor node is included as input features.  Inputs are mapped to a Gaussian feature space, so that the effect of measurements / examples have less influence on far away sensors.  The extreme points act as vertices of the convex set, and effectively describe the hull of the set with the fewest number of points from the training examples.  Training is performed using a one-hop gossip protocol.  Individual nodes are provided an initial set of training data, from which the extreme points of the positive and negative examples are extracted.  Nodes maintain a collection of convex sets of positive and negative examples, described by the extreme points of the sets.  Nodes in the network shares extreme points of the current convex hulls for positive and negative training examples with neighboring nodes using single hop communication.  Upon receiving a set of extreme points, a node performs join operations on the convex sets it is aware of, possibly reducing the number of convex sets and extreme points the node must maintain and communicate.  The proposed algorithm converges to the results achieved of a centralized geometric SVM in finite time, and reduces communication and memory requirements compared with other algorithms.  Using a naive algorithm to approximate a convex hull, extreme points can be minimized to allow scalability with respect to the amount of training data and communication packet length.

A method for building an ensemble of SVR predictors for target localization using acoustic data is given in \cite{kim2013-svr}.  The sensor network is subdivided into subnetworks, where subnetworks represents a local region of the physical area being monitored, and training data for each subnetwork is extracted randomly from a global training set.  For each subnetwork, a SVR predictor is trained using a least squares SVR algorithm.  An aggregated predictor can be formed by taking the weighted sum of individual predictors, where the weights are based on the estimated distance from the target to to the location of sensors in the subnetwork.  A distributed distributed algorithm simply predicts and initial position for the target, and then iteratively shares position estimates with other subnetwork predictors, until convergence on a predicted position.  Experiments demonstrate that this ensemble approach is more robust to noise than a conventional SVR predicted using all sensor nodes.

In a similar manner to the ensemble SVR approach, a modified support vector domain description (SVDD) is created used in conjunction with a local classification strategy to track a target in an environment \cite{kim2014-svdd}.  The authors used this approach to localize a target in an environment using acoustic monitoring.  SVDD performs classification by creating a sphere around the data to be classified, and attempts to minimize the radius of the sphere.  This amounts to solving an optimization problem similar to an SVM classifier, except that the optimization problem attempts to minimize the volume of the sphere containing the positive training examples.  Rather than using the measurements from the full set of sensor nodes, the authors instead perform local classification by collecting measurements from a local neighborhood, and performing classification on that reduced dataset.  While this does reduce the amount of computation performed in local nodes, the experimental success relies on the fact that an acoustic measurement is highly dependent on the distance from the source, and consequently distant measurements have very little effect on classification results.  In addition to this signal dependence, there is no proof of global convergence of the algorithm.

A key component to many of the distributed SVM algorithms presented here is the consensus algorithm used during training.  As one major design goal in sensor networks (and robotic materials) is to minimize the amount of communication in the network, selection of gossip algorithm can have a large impact on the overall performance and lifespan of a sensor network.  For instance, nodes share a full set of training data in \cite{gu2008-svr}, whereas only a set of points describing convex hulls of training data are shared in \cite{kim2015-svm}  In \cite{flouri2008distributed}, two gossip protocols for SVM training were developed and compared:  the minimum selective gossip algorithm (MSG-SVM), which attempts to minimize the amount of data diffused in the network, and the sufficient selective gossip algorithm (SSG-SVM), which attempts to maximize the optimality of the resulting trained SVM (i.e., match the performance of a global centralized SVM).  In the MSG-SVM algorithm, each node computes its support vectors based on local training data.  Nodes then communicate support vectors with one-hop neighbors, and update their SVM models and support vectors with those received by neighboring nodes.  At subsequent iterations, nodes communicate only the support vectors which have not been previously shared.  This process continues until a terminal condition is met.  As a minimal number of support vectors are shared, this algorithm is efficient in terms of communication.  However, a sub-optimal discriminant hyperplane is produced--at each iteration, the set of support vectors from training on the separate datasets are not necessarily the same as the support vectors obtained from the union of the datasets.  The SSG-SVM algorithm, in contrast, is designed to converge to a globally optimal solution.  In this algorithm, neighboring nodes exchange the points describing the convex hull of each training set, similar to the approach in \cite{kim2015-svm}.  Experiments demonstrated that the SSG-SVM algorithm resulted in a more accurate classifier than the MSG-SVM algorithm (by ~2\% - 5\%), whereas the MSG-SVM algorithm resulted in many fewer data points communicated between neighbors (by a factor of about 3). 

\subsubsection{Distributed Expectation Maximization Algorithms for Mixture Models}\label{sec:sensor_network_gmm}

A common approach to density estimation (e.g., population of fish, temperature distribution, density of gas present in atmosphere, etc.) in sensor networks is to model measurements of the environment as a mixture of Gaussian or other distributions.  Several algorithms have been developed to determine the parameters of a Gaussian mixture model, usually as a distributed variation of the expectation-maximization (EM) algorithm.  A distributed EM algorithm is developed and analyzed in \cite{nowak2003-distributed_em}.  In this approach, each node in the sensor network maintains mixture parameters (i.e., mixture probabilities, means and covariance matrices) of the global distribution.  Learning the parameters is performed iteratively in a distributed manner.  During an expectation step, each node calculates the sufficient statistic for the mixture model locally based on the current model parameters and local measurements.  The statistic is broadcast through the network, and each node accumulates received statistics to computed the global sufficient statistic.  The maximization step can be performed locally at each node, updating the parameters based on the sufficient statistic and local measurements.  This approach converges to the same distribution achieved by a standard (i.e., non-distributed) EM algorithm.  Multiple maximization steps can also be performed at each iteration, which results in faster convergences with respect to the number of communicated bits.  However, this also results in a higher likelihood of converging to a local maximum.

Newscast EM is a distributed EM algorithm which uses a gossip-based protocol to share mixture parameters and converge to a global solution \cite{kowalczyk2005-newscast_em}.  During the maximization step, a node selects a neighboring node at random, where a neighbor is any node within communication range.  The nodes exchange local versions of their parameters (i.e., mixture probabilities, means and covariance matrices), and perform an update by calculating the average value of each parameter, weighted by the local mixture probability.  The expectation step is performed in parallel, using locally available data, and remains unchanged from the standard EM approach.  This algorithm is shown to probabilistically converge to the global solution in exponentially fast time.  However, this analysis assumes that nodes can communicate arbitrarily with any other node in the network, which is not typically the case in sensor networks, and convergence properties may not hold.

A consensus-based approach to distributed EM is presented in \cite{forero2008-consensus_em}.  This approach relies of a subset of nodes, $b \in \mathcal{B}$ termed \emph{bridge sensors}, which are used to maintain a set of consensus parameters, $\mathbf{\bar{\varphi}}_b$ (i.e., parameters for each pdf in the mixture).  Each node is directly connected to at least one bridge sensor (i.e., single-hop communication), and may be connected to multiple bridge sensors.  Prior to each expectation step, the bridge sensors transmit their consensus variables to each neighboring node.  Each node uses these consensus variables to perform the expectation step locally, assigning a mixture label to each measurement.  The prior distribution over the mixtures, $\pi_j$ are determined locally.  Performing an update on the mixture parameters involves solving the following optimization problem

\begin{equation}
\begin{aligned}
  & \underset{\mathbf{\varphi}}{\text{maximize }} & & -\sum_{j=1}^J \mathbb{E}_c \left\{L^c(\mathbf{y},\mathbf{\theta}_j | \mathbf{x},\mathbf{\theta}^{(i)}_j \right\} \\
  & \text{subject to} & & \mathbf{\varphi}_j = \bar{\mathbf{\varphi}}_b, j \in \mathcal{N}_j, b \in \mathcal{B} \\
  &                   & & 0 \le \pi_{j,k} \le 1, \forall j \in \mathcal{J}, k \in \mathcal{K} \\
  &                   & & \sum_{k=1}^K \pi_{j,k} = 1
\end{aligned}
\end{equation}

which can be solved in a distributed manner using the method of multipliers (MoM), and the dual of which introduces local Lagrange multipliers, $\lambda_j^b$, and penalty coefficients, $\xi_j$.  At each time step, the Lagrange multipliers and local and bridge mixture parameters are updated by

\begin{equation}
\begin{aligned}
\lambda_j^{b(i+1)} & = & \lambda_j^{b(i)} + \xi_j \left(\varphi_j^{(i)} - \bar{\varphi_b}^{(i)} \right) \\
\varphi_j^{(i+1)} & = & arg\min_{\varphi_j} - \sum_{n=1}^N \sum{k=1}^K \hat{c}_{j,n}(k) log [f_k(x_{j,n}; \phi_{j,k})] \\
                 &   & + \sum_{b \in \mathcal{B}_j} \lambda_j^b(\varphi_j - \bar{\varphi}_b) + \sum_{b \in \mathcal{B}_j} \frac{\xi_j}{2} \| \varphi_j - \bar{\varphi}_b \|_2^2 \\
\bar{\varphi}_b^{(i+1)} & = & \sum_{j \in \mathcal{N}_b} \frac{1}{\sum_{\beta \in \mathcal{N}_b} \xi_\beta} \left( \lambda_j^{b(i+1)} + \xi_j \varphi_j^{(i+1)} \right)
\end{aligned}
\end{equation}

where the local Lagrange multiplier and mixture parameters are calculated at each node, and then transmitted to the bridge nodes to calculate the consensus parameters.  The parameters arrived at using this distributed approach also match those formulated from a centralized EM.  In addition, this approach has the advantage of not requiring a path traversing across all sensors, requires only one-hop communication between sensors, and applies to general mixtures.

A distributed EM algorithm similar to the approach in \cite{nowak2003-distributed_em} is presented in \cite{gu2008-distributed_em}.  In this approach, both the expectation and maximization steps are performed locally at each node.  Rather than communicating sufficient statistics throughout the network at each step, as in \cite{nowak2003-distributed_em}, each node maintains an estimate of global summary quantities, $\mathbf{x}_{i,k}^t$.  The statistics calculated by each node at the end of the expectation step, $\mathbf{u}_{i,k}^t$, and the global estimates are communicated to neighboring nodes.  A consensus filter then updates the estimated global summary quantities using the equation

\begin{equation}
\mathbf{x}_{i,k}^{t+1} = \mathbf{x}_{i,k}^{t} + \eta \left[ \sum_{j \in \mathcal{N}_j} (\mathbf{x}_{j,k}^{t} - \mathbf{x}_{i,k}^{t}) + (\mathbf{u}_{i,k}^{t} - \mathbf{x}_{i,k}^{t}) \right]
\end{equation}

where $\eta$ is an update rate.  Each node then performs a maximization state to update the parameters using the estimated global summary quantities.  Rather than performing communication through the entire network, this approach works using an arbitrary hop count to communicate global summary quantities.  This approach is stable and stochastically approximates the global maximum likelihood estimate of the mixture parameters.


\subsubsection{Distributed Algorithms for Graphical Models}\label{sec:sensor_network_pgm}

Given the graphical nature of the communication network in sensor networks, algorithms for learning and inference in graphical models map well to sensor networks.  For example, sensor networks may be represented as a graph, with sensor nodes representing nodes in the graph, and communication representing edges between nodes.  Given this mapping, belief propagation algorithms are easily adapted for use with sensor networks.  

Distributed target tracking using is presented in  \cite{shi2009-sensor_network_mrf}.  By considering only the immediate neighborhood of nodes, the graph forms a Markov random field (MRF), which may be used as a probabilistic model for tracking small targets or determining the boundary of large targets.  To do this, cliques are formed in the network, and the probability that the target is located within the clique or not is determined by nodes in the clique.  Using a message passing algorithm, the probability of the boundary given measurements of all nodes can be determined.  The approach has several advantages, namely allowing a fully distributed approach to tracking, requiring only local communication between cliques, allowing tracking of multiple targets, and allowing prior belief of target boundaries to be used.

A similar belief propagation algorithm is described in \cite{paskin2004-inference, paskin2005-inference} for performing distributed inference of latent environmental variables in a sensor network.  This approach assumes that local sensor measurements can be modeled as the product of a measurement model and environment model.  The measurement model represents the probability of a particular observation, conditioned on local environmental variables of interest, while the environmental model is a factorized prior (e.g., Markov or Bayesian network) over all environmental variables of interest.  To perform distributed inference, four separate distributed algorithms are run concurrently through the network.  A spanning tree formation algorithm computes a tree from the communication graph of the network.  A junction tree formation algorithm determines which environmental variables are shared by different nodes in the tree, and ensures that the running intersection property is enforced in the network.  The running intersection property ensures that if a variable is known by two nodes in the tree, the variable is also known by all nodes in the path between the two nodes.  The junction tree formation algorithm allows each node to maintain a clique of variables for local inference, as in \cite{shi2009-sensor_network_mrf}.  A tree optimization algorithm attempts to reduce the size of cliques in the junction tree by performing legal local edge swaps in a greedy manner.  Finally, belief propagation can be performed over the tree through message passing, where messages involve communicating estimates of variables between nodes in the network.  This approach modifies the belief propagation algorithm by introducing a robust message passing algorithm, allowing belief propagation to be performed without guarantees that the supporting algorithms have completed, and making belief propagation robust to node loss and communication loss.  This is achieved by simply maintaining partial beliefs at each node--a set of model fragments consisting of approximation of environmental variables and the likelihood of an observation are maintained, and are updated based on received messages from other nodes.

A simpler loopy belief propagation algorithm had been described earlier than the junction tree based belief propagation algorithm \cite{crick2003-loopy_belief_propagation}.  Each sensor node maintains a Bayesian graphical model, where sensor measurements are conditioned on environmental state.  The conditioning on environmental state introduces dependencies between neighboring sensor nodes.  The investigation experimentally compares loopy belief propagation performed asynchronously in a sensor network to a synchronous approach, where updates to variables in the graphical model can be performed to a predefined sequence.  The investigation demonstrates that loopy belief propagation can be applied in an asynchronous manner in sensor networks, and still converge to the same belief as a synchronized version of loopy belief propagation.  The approach worked robustly in the presence of non-uniform communication rates between nodes, robust to node failure, and converges even in the presence of rapidly changing environmental conditions.

\textbf{Distributed Dynamic Bayesian Network.}  While there are no distributed temporal models available in the sensor network literature, distributed approaches to dynamic Bayesian networks are demonstrated for fault diagnosis in electronic circuits, and hidden gait recognition.  A multi-stage fault diagnosis system for electronic circuits is presented in \cite{roychoudhury2009-distributed_dbn}, which uses a DBN to represent the dynamics of the circuit based on the energy exchange between circuit components.  The global DBN is factored into a maximal number of conditionally independent DBN Factors (DBN-Fs):  a DBN-F is created such that it is conditionally independent from all other DBN-Fs given its inputs, and one or more state variables are replaced by algebraic functions of measured variables to enable the removal of across-time dependencies.  Nominal diagnosers detect defects in the circuit given inputs and measurements by comparing an estimate of the measurements (calculated using particle filtering) and the actual measurements.  Faults are propagated to other detectors through shared measurements (i.e., the algebraic representation of state variables which can be directly measured); by sharing measurements in this manner, detection of a fault is isolated to a single detector, while detectors sharing measurements with this detector simply treat it as a faulty measurement.  While this approach provides a simple means of detecting defects in robotic materials, the generation of algebraic representation of state variables (which is simply deterministic voltage-current relationships in circuit models) based on observations may not be a trivial task.

Consensus-based algorithms for hidden Markov models have also been explored \cite{taborri2014-gait_phase}, which are more closely aligned with the approaches in Section \ref{sec:sensor_network_pgm}.  In \cite{taborri2014-gait_phase}, individual sensors calculate the most likely sequence of gait phases using a continuous HMM (modeling observations as a multivariate Gaussian).  At each time step, the current gait phase estimate from each sensor node is communicated to a central classifier.  The central classifier then sets the phase estimate to the received values if all agree.  In the case of a disagreement, the probability distribution of gait phases for each sensor node is estimated from the prior state estimate using a nominal transition matrix, from which the most likely gait phase is selected.  This investigation used a small (four sensor node) model, and also assumed a nominal gait transition matrix, which may not be available for all robotic materials applications.


\section{Machine Learning Approaches used in Engineered Materials}\label{sec:ml_in_materials}

Several specific fields have been cited as motivation for robotic materials \cite{mcevoy2015-science}, where robotic materials may extend the potential of the materials of interest in these fields.  In structural health monitoring (SHM) and nondestructive evaluation (NDE), material and structures could self-diagnose and self-repair.  In the field of aerospace, morphing aerodynamic surfaces can improve lift and stability of aerospace vehicles.  In the field of robotics, full-body tactile sensitive skins are of interest, as they provide a rich sensing modality which can augment existing sensors, specifically in safety-critical and highly occluded environments \cite{hughes2014-skin}.  Finally, wearable computing can provide assistive technology and monitor human motion.  Each of these fields have explored using machine learning techniques for domain-specific tasks.  This section highlights the common approaches in each field, and describes some of the challenges faced in each.

\subsection{Structural Health Monitoring}

Structural health monitoring is the process of detecting and identifying damage in structural and mechanical systems, such as civil (e.g., bridges, buildings), aerospace (e.g., aircraft wings or fuselages) or mechanical (e.g., oil platforms) infrastructures \cite{farrar2007-shm}.  During the life cycle of these infrastructures, defects inherent in the underlying material grow until the structure can be considered in a state of damage, in which the performance of the structure is adversely affected.  Once the damage affects the structure beyond some acceptable point, the structure is said to have failed.  The goal of structural health monitoring is to provide a means of online identification and evaluation of damage to the structural system, using a periodic network of sensing elements.  A smart structure may automatically perform monitoring, processing of data, and highlight the need for human intervention or repair.  A more detailed definition of structural health monitoring defines a hierarchy of the main issues \cite{worden2004-intelligentfault}:

\begin{enumerate}
  \item \textbf{Detection:}  provide an indication that damage may be present in the structure.
  \item \textbf{Localization:}  identify the probably position of the damage in the structure.
  \item \textbf{Classification:}  identify the type of damage in the structure.
  \item \textbf{Assessment:}  provide an estimate of the extent of the damage in the structure.
  \item \textbf{Prediction:}  provide information about the safety of the structure, the remaining life of the structure, or a prediction of the growth of damage.
\end{enumerate}

This hierarchy not only describes the main tasks of interest in structural health monitoring, but also the prerequisites for performing each operation.  For example, damage must be detected before being localized, classification can only occur after localization, etc.  

Physics-based analysis of structures involve modeling the response of a particular structure to some external stimulation, such as an impact or vibration.  For simple structures, such as plates and beams with simple geometries, closed form solutions to responses can be easily calculated.  However, when structures are more complex in geometry (e.g., an aircraft), or when defects such as cracks, voids or delaminations are present, an exact solution to stimulus response is not easily obtained.  Finite element method (FEM) \cite{zienkiewicz2005finite} is a powerful and common approach which involves generating a mesh which approximates the structure of interest, approximating the physical response of each element in the mesh, and enforcing continuity of the physical behavior between meshes at their boundary.  For structural health monitoring, the undamaged state of a structure may be simulated using FEM, and the behavior of the actual structure can be compared to this simulation to determine the presence of defects or damage.  Additionally, failure responses under cyclic loading, such as stiffness degradation and crack propagation in composites, can be simulated to provide failure analysis of the structure of interest \cite{zou2000vibration}.  The primary drawback of FEM is the computational requirements are very large:  the physical dimensions of meshes must be small enough to effectively model small defects, while the number of meshes must be sufficient to cover the entire structure under investigation.  Machine learning approaches, in contrast, exchange the complexity of physically modeling the system with gathering a large number of training examples.  For complex objects, this may be more easily achieved.  For damage detection, training examples are only required for situations where the system is operating in an undamaged state, and defects can be simulated in many cases by introducing masses or bumpers where a defect would occur.

A common approach to investigating structures is to measure the response of the structure to some external vibration.  In addition to a wide number of developed features available, large structures are typically stimulated from external sources during operation.  For instance, vibrations can be generated in a multi-story building during an earthquake \cite{song2006-shm_ica_svm, elkordy1993-neuralnetwork}.  Vibrations are naturally induced in aircraft surfaces during flight, and on bridges due to wind or dynamic loading.  Vibrations can also be explicitly generated by external actuators.  Piezoelectric transducers have also been used to generate vibrations in composite panels \cite{mechbal2015-probabilistic_svm}.  Structures are also excited by placing them on large shakers \cite{worden2003-experiment1,manson2003experimental}.  These two approaches are generally reserved for lab investigations, as they require an additional component or are impractical for real-world applications.

As discussed in \cite{worden2007-ml_shm}, each of the structural health monitoring tasks maps well to a specific machine learning objective.  Damage detection is best framed as a density estimation problem:  measurements from undamaged components can be described as some probability distribution, and damage is detected by measuring outliers of this distribution.  Classification could also be considered for damage detection (i.e., classify components as damaged or undamaged), however, this requires a prohibitively large number of damaged samples for training.  Damage localization is easily framed as a regression problem.  Identifying damage is best suited as a classification problem.  In many cases, regression is suitable for damage assessment.  Machine learning techniques, however, are typically not suitable for damage prediction, as tasks such as estimating remaining life or predicting the propagation of a crack require detailed knowledge of the underlying physical system, properties of the material, and physical modeling of the propagation of damage.

\subsubsection{Damage Detection}\label{sec:damage_detection}

Damage detection usually involves constructing a density estimation of features from the undamaged system, and by treating the presence of outliers as an indication of damage, or by projecting feature vectors onto a model trained to represent features from an undamaged system and using the residual error as an indication of damage.  The output of an detection model is referred to as a damage indicator \cite{figueiredo2011-ml_damage_detection}, whose output is a positive scalar which indicates the likelihood that damage is present.  Detection models require a set of training vectors measured from the undamaged system, and outputs a damage indicator, $DI$, from a given input feature vector.  

Framing damage detection as an anomaly detection task is a pragmatic decision.  Alternatively, a classifier could be trained to identify a structure as defective or not defective.  However, this requires generating training examples for cases where defects are present.  This could be prohibitively costly, as several iterations of the same structure would need to be damaged at several locations to collect an appropriate cases.  Despite this, classification approaches have been explored in the past.  Simulated benchmark data has been generated using finite element models \cite{johnson2004-shm_benchmark}, and have used to train classifiers for damage detection \cite{song2006-shm_ica_svm}.

Variability due to operational or environmental changes can adversely affect underlying measurements used to detect damage \cite{figueiredo2011-ml_damage_detection}.  For example, the natural frequency of bridge girders can change by 5-10\% due to variations in traffic or temperature.  Damage-sensitive features have been developed which are robust to these types of variations.  For example, the coefficients of an autoregressive model fit to accelerometer measurements were used to detect cracks in a simulated building during an earthquake, where the load or stiffness of the building is varied \cite{figueiredo2011-ml_damage_detection}.  Using these features, total detection error ranged from 4.0\% to 4.6\%.  In a similar investigation, bumpers were introduced into a vibrating mass-spring system \cite{oh2009-damage_svm}.  Using autoregressive features, defects were detected 67 out of 70 times, despite variations in the level of applied vibration.

MSD-based DIs have been used extensively for damage detection.  In \cite{worden1999-concordancy}, the presence of delamination in composite plates was detected using MSD when a ultrasonic sensor was inline within 20mm of the delamination.  In \cite{figueiredo2011-ml_damage_detection}, non-linear defects were introduced into a three-story frame structure; defects could be detected using MSD with an error rate of 4.0\%.  MSD, KDE, and autoencoder DIs were compared in \cite{worden2003-experiment1}.  Saw cuts were introduced into a panel, whose lengths varied from 10\% to 90\% of the length of the panel, and were measured using accelerometers and vibrating the panel using an external source.  MSD-based DIs could reliably detect cuts 30\% of the panel length or longer, KDE-based DI performed significantly worse, detecting cuts 40\% of the length or longer, while the autoencoder approach produced the best results, detecting defects 20\% of the length or longer.

A similar comparison was performed in \cite{figueiredo2011-ml_damage_detection}, where defects in a four-story structure were simulated and measured using accelerometers at each floor.  Reduction in stiffness of the columns and added mass on the floors were used to simulate environmental variability, though these were not considered defects.  In this study, DIs based on MSD, factor analysis, autoencoders and SVD were used, resulting in an overall detection error of 4.0\%, 4.2\%, 4.3\% and 4.6\%, respectively.  The MSD- and autoencoder based algorithms resulted in lower false negative results, while the factor analysis and SVD-based results resulted in lower false positive results.

While classification approaches are typically infeasible due to the high cost of obtained examples of damaged structures, simulated data has shown that classifiers can produce very accurate predictions of whether damage is present in a structure.  Using simulated data, classification approaches to damage detection have demonstrated a high level of accuracy.  For example, support vector machines and neural networks have classified a simulated building as damaged or not damaged, with accuracies of 99.2\% - 99.8\% and 95.3\% - 98.4\%, respectively \cite{song2006-shm_ica_svm}.  

Finally, damage detection and localization have also been combined into a single classifier, by considering damage location as discrete sections of the structure and incorporating an ``undamaged'' class \cite{mechbal2015-probabilistic_svm}, as part of a SVM-based geometric probabilistic decision tree (SVM-GPDT).

\subsubsection{Damage Localization}

Damage localization is often cast in a regression framework, with output being coordinates of the damage, though classification models have been used when damage is located at discrete position in the structure.  For example, \cite{worden2007-ml_shm} use a classifier to identify which of nine panels are damaged in an aircraft wing.  

Localization of defects are often better suited as classification problems, depending on the underlying structure under investigation or the desired granularity of localization.  For instance, in a truss structure (e.g., certain bridges and buildings), it may be more important to identify which beam is faulty, rather than an exact location of the actual damage.  Similarly, determining if a specific composite panel is defective on an aircraft may be sufficient, as the entire panel would be replaced rather than repaired.  Such a classification approach was given in \cite{manson2003experimental3}.  Transmissibility features were calculated using piezoelectric elements located on a aircraft wing, such that panels were located between transmitting and receiving elements.  A neural network was trained to detect which of nine panels were removed with an accuracy of 86.5\%.

The SVM-GPDT used in \cite{mechbal2015-probabilistic_svm} uses a probabilistic binary tree to determine the location of a defect in a carbon FRP panel, using three piezoelectric transducers (one to emit and two to detect vibrations).  Damage was simulated by placing weights in different regions of the panel.  Features consisted of 200 Fourier components calculated at each detecting transducers.  Using this approach results in classifier which is more robust to misclassification.  Each leaf produces a probability that the defect occurs in a particular subregion of the panel; the geometric decision tree results in misclassifications occurring more often in neighboring nodes when compared to a ``flat'' one-against-all SVM classifier.  The authors report that damage can be localized well in 88\% of example cases, can able to distinguish between two severities with an accuracy of 99.42\%.    

One specific task in structural health monitoring best suited as damage localization is impact monitoring.  Impact monitoring can be very useful in SHM, as impacts are often the cause of serious damage located in the interior of the material, making resulting damage difficult to detect.  This is of specific concern in the aerospace industry, where critical components are constructed of lightweight composite panels, and are subject to high-speed impacts during flight.  Impact detection is naturally framed as a localization problem, either through triangulation or machine learning techniques \cite{haywood2005-impact_monitor}.  Impact monitoring has the advantage of allowing the use of training data collected from real samples--impacts which have low enough energy can be used to collect data without damaging the training samples.

Several investigations involve detecting and localizing impacts in composite aircraft wings.  In \cite{haywood2005-impact_monitor}, a neural network was trained to determine the location of an impact on aircraft composite panels using time-domain features (minimum and maximum values, time of minimum and maximum values, beginning and end of signal detected, etc.).  The average error of the estimated location of impact was 0.85\% of the area of the panel using 12 sensors, and 1.61\% using 4 sensors.  An similar investigation found that introducing noise of 20\% RMS of the measured signal in training data results in localization error of 2.9\% of the area using time after impact of maximum response features, and 6.5\% of the area using magnitude of maximum response \cite{staszewski2000-impact_detection}.  In addition, the force amplitude of the impact was estimated with a mean error of 28\%.  A neural network implemented on an actual wing section section achieved a localization error of 1.76\% using a regression approach \cite{leclerc2007-impact_detection}.  A secondary neural network was trained to classify which portion of the wing (leading edge, center, training edge) was impacted with an accuracy of 3.2\%, with errors near the border between regions.  Combining the two approaches, by training a regression network for each region, resulted in a reduction in localization error to 0.48\%.  In \cite{sharif-khodaei2012-impact_location}, two sensors were used to localize impacts on a simulated composite panel using a neural network.  In addition to the features used in \cite{haywood2005-impact_monitor}, the authors investigated using time of arrival of the signal to each sensor, and information from the discrete wavelet transform (DWT) of the signal, with a localization error was 0.4\%.  Note that these results were from simulated panels, and required time of arrival data, which may be difficult to obtain without synchronization of sensor measurements.

\subsubsection{Damage Assessment}

The final stage of structural analysis involves assessing the severity of damage on the structure.  

In \cite{worden2007-ml_shm}, neural networks were used to determine the percentage of a panel missing.  Training sets consisted of data collected from panels with holes whose area is 25\%, 50\%, 75\% or 100\% (i.e. missing panel), representing various damage severities.  Using neural networks with five to nine hidden units, the damage area was calculated with a mean square error ranging from 0.92\% to 5.6\%, depending on the panel.  As damage severity was set to four discrete values, and as a separate neural network is required for each panel, this approach is not the most satisfactory approach.

Damage localization approaches which use classifiers to identify a discrete location of damage have been extended to assess the level of damage at those locations.  For instance, in \cite{elkordy1993-neuralnetwork}, three neural networks were trained to analyze a simulated building structure using four accelerometers.  The first two detected and identify which floor of the building is defective, while the third was trained to identify the level of damage (e.g., percent reduction of bracing stiffness) of each floor.  

The SVM-GPDT approach used in \cite{mechbal2015-probabilistic_svm} incorporates a simple binary SVM classifier to identify which of two damage severities is present in the FRP panel, with an accuracy of 99.42\%.  Unlike other approaches, this overall approach performs damage assessment prior to localization (as opposed to localization before assessment).  Using this approach, the damage severity can be used as a prior to the probabilistic decision tree to improve localization accuracy.

\subsubsection{Damage Prediction}

As mentioned earlier, machine learning approaches are poorly suited to predict the growth of damage, safety or lifetime of a structure.  Despite this, some machine learning approaches have been used to perform these types of predictions.  Using resistance measurements, monitoring the fatigue of CFRP panels is possible, and neural networks have been used to predict fatigue life (i.e., number of remaining load cycles) and stiffness reduction in the panel \cite{seo1999-cfrp_nn}.  


\subsection{Aerospace Applications}

From the perspective of robotic materials, tasks of interest in aerospace applications consist primarily of monitoring aerodynamic surfaces for defects, and controlling the shape of aerodynamic surfaces to increase lift, provide stability or improve efficiency.  Machine learning approaches to the formal task are explored as structural health monitoring in the prior section.  This section investigates machine learning techniques used for control of morphable aerodynamic surfaces.

\subsubsection{Morphable Aerodynamic Surfaces}

Morphing aircraft describe a wide range of approaches to adaptable aircraft components \cite{weisshaar2013-morphing_aircraft}.  Several advancements in wing capabilities have been realized over the last century.  Leading-edge and trailing-edge flaps and slats enable control over the camber of the wing and allow safer operation at low speeds.  Passive variable camber wings were realized using a compliant structure.  Changing wing area is possible through foldable wings, and variable wing sweep and out-of-plane wing shapes are similarly possible.

More recently, using smart materials, such as shape memory alloys and piezoelectric materials, to construct morphable aircraft has been proposed.  Using such active elements can allow for aircraft that attenuate dynamic loads, suppress flutter and vibration, perform noise cancellation inside the aircraft, and provide adaptive stiffness control \cite{weisshaar2013-morphing_aircraft}.  Large scale research efforts into morphing aircraft include the NASA Langley Morphing Aircraft Project \cite{wlezien1998-aircraft_morphing} and the DARPA Morphing Aircraft Structures Program \cite{weisshaar2006-darpamas}.  This latter research effort has resulted in a push for wings capable of altering its shape in a drastic and continuous manner, similar to the configuration capabilities of bat and bird wings \cite{patel2005-morphing_wings, colorado2012-bat_robot}.  This would allow aircraft which could achieve high level of performance in contradictory situations (e.g., quick maneuvering versus long-duration cruising) through various wing configurations.

Shape memory alloys (SMA) are used extensively as a means of actuation in morphable wings \cite{ruotsalainen2009-shape_control_frp, barbarino2014-sma_review}.  Characterization and modeling of SMA materials is still an active area of research.  SMA materials exhibit two types of effects based on temperature and load, namely a super-elastic effect and a shape memory effect \cite{barbarino2014-sma_review}.  Additionally, the temperature-strain relationship of the material demonstrates hysteresis as temperature is varied, making control difficult \cite{haag2005-sma_characterization}.  To handle this complexity, some recent studies have involved determining an optimal way of controlling the material, including using the SARSA algorithm to approximate a state-action value function based on a K nearest neighbor (KNN) approach \cite{kirkpatrick2009-rl}, and a neuro-fuzzy controller to control the shape of a flexible wing skin \cite{grigorie2010-neurofuzzy_controller}.  Extending this, reinforcement learning has been used to control the shape of an aerodynamic surface built using SMA materials \cite{tandale2004-preliminary_rl, valasek2005-rl, doebbler2005-improved_rl, valasek2008-improved_rl}, or motion of a flapping wing \cite{motamed2006-simulation, motamed2007-experimental} with SMA actuators.

\begin{figure}
\centering
\includegraphics[width=0.8\textwidth]{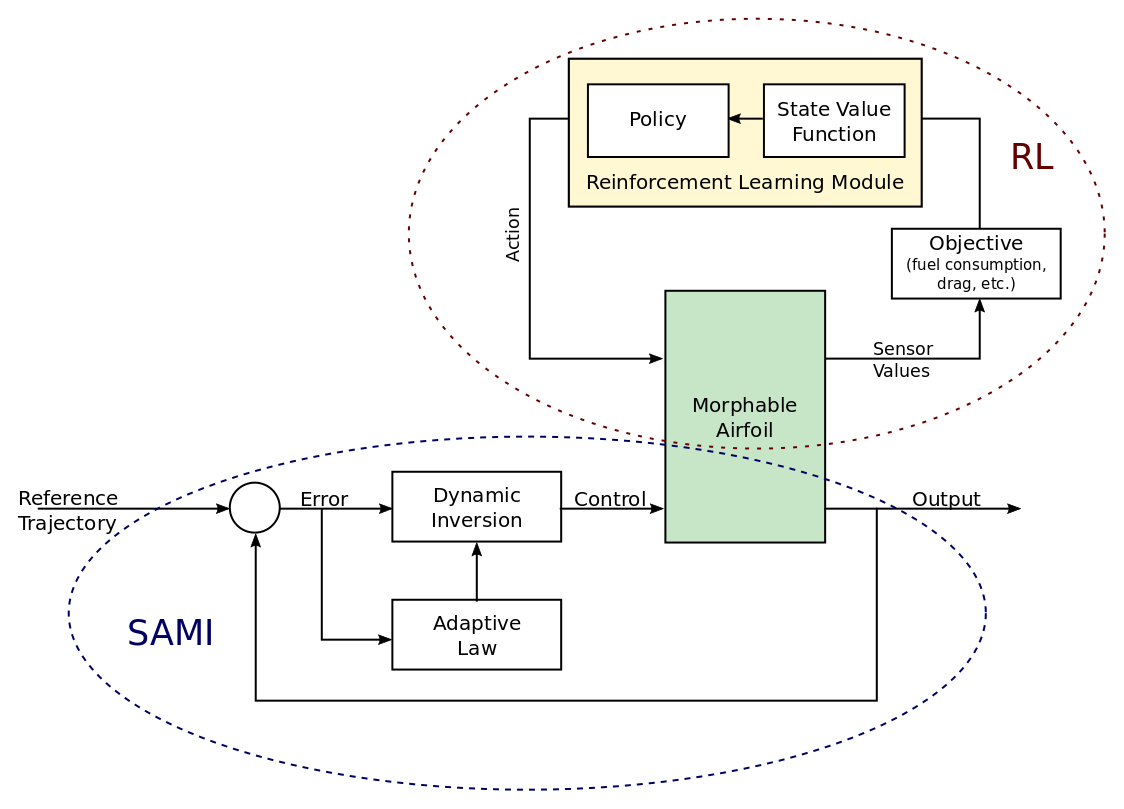}
\caption{Architecture merging an adaptive model with reinforcement learning for controlling morphable aircraft surfaces, as used in \cite{valasek2005-rl}}
\label{fig:RL_SAMI}
\end{figure}

Reinforcement learning has been used in conjunction with a structured adaptive model inversion (SAMI) controller to create an adaptive controller for morphable aircraft, as shown in Figure \ref{fig:RL_SAMI}.  A reinforcement learning module learns to reconfigure a vehicle to a shape which is optimal with respect to flight condition and some cost--for instance, to minimize drag force, increase lift, or minimize fuel consumption.  The SAMI controller is an adaptive controller which allows the aircraft to track a reference trajectory in the presence of changes in dynamic behavior due to vehicle shape.  This architecture decouples the tasks of optimizing shape and maintaining trajectory, and allow shape control to be learned independently.

Using the Q-Learning algorithm, the control of various shapes and use of various approximation functions have been explored.  Initial explorations involved simple geometric shapes \cite{tandale2004-preliminary_rl, valasek2005-rl, valasek2008-improved_rl}, with more recent results determining policies for airfoil cross-sections \cite{lampton2009-rl, lampton2010-rl}.  Preliminary explorations controlled the shape of a constant-volume ellipsoid \cite{tandale2004-preliminary_rl, doebbler2005-improved_rl} or block \cite{valasek2005-rl}, using an arbitrary functions to describe the behavior of the shape with respect to input voltage, and an arbitrary function for the ideal shape for a particular flight condition.  Shape change is described by second-order differential equations, so that the learned policy must be time-dependent.  

The trained policy generates control voltages to adjust the shape based on current flight conditions.  The action-value function (i.e, Q function) is described using a variety of function approximations.  Initially, a k-nearest neighbor (KNN) approach interpolates the value of the Q function from the interpolation of the K nearest state-action pairs \cite{tandale2004-preliminary_rl}, though this suffered when interpolating in regions which have not seen training examples.  A function approximator consisting of a linear combination of tile functions (i.e., tile coding) was used with the shape changing block \cite{valasek2005-rl}.  A Galerkin-based Sequential Function Approximation (SFA) is used in \cite{doebbler2005-improved_rl, valasek2008-improved_rl}, which uses the Petrov-Galerkin approach to select the coefficients for a linear combination of radial basis functions.  The SFA approach resulted in a reduction of RMS error of 10\% to 20\% compared to the KNN approach.

The above approach has been extended to control the cross-sectional shape of an airfoil \cite{lampton2009-rl, lampton2010-rl}.  The model used allowed for a policy to control of the thickness and camber at 100 locations along the length of the airfoil, and optimal configuration is based on the lift, drag and moment coefficient about the leading edge of the airfoil.  As in \cite{tandale2004-preliminary_rl}, the value-action function is approximated using a KNN approach.  To reduce the number of possible state-action pairs, the state-action space is discretized into a lattice, in which admissible actions consist of only increasing or decreasing the thickness or camber by a fixed percentage, resulting in 180 to 13,516 state-action pairs, depending on the discretization.

In addition to controlling the shape of an airfoil, Q-Learning has been used to learn the trajectory of flapping wings for micro aerial vehicles (MAVs) \cite{motamed2006-simulation, motamed2007-experimental}.  While shape control is useful for optimizing a static configuration for various flight parameters, while flapping involves dynamic, though repetitive, actions in a stationary or slowly-varying environment.  In these investigations, the time-varying stroke angle and angle of attack of an insect-like wing is learned which maximizes the lift force of the wing.  Simulation results \cite{motamed2006-simulation} demonstrate that lift force and the frequency of jerk (which could damage wing joints) can be jointly optimized using Q-Learning.  A follow-up experiment finalized the trajectory of a physical wing using Q-Learning after a trajectory was formed using the simulation \cite{motamed2007-experimental}.  


\subsection{Wearable Computing}

Several applications in wearable computing consist of analyzing motion through one or more sensors located about the body.  Activity recognition, also known as context awareness in the wearable community, involves identifying physical activities performed by a wearer, from low-level activities (e.g., standing, walking, and running) to high-level activities (e.g., vacuuming, brusing teeth, and making breakfast).  Accelerometers are typically used as sensors \cite{ravi2005-activity}, as these are typically less expensive than other appropriate sensors and are present in many consumer electronic devices (e.g., smartphone), through gyroscopes or IMUs are also used.  Anomalous gaits, such as freezing gait, are present in individuals with diseases such as Parkinson's disease, or after a stroke \cite{manap2011anomaly}.  Detection of gait anomaly and be used to predict and possibly prevent falls.  Finally, soft, compliant exosuits show potential to assist the wearer with walking, with potential military and rehabilitation applications \cite{cornwall2015pursuit,bae2015soft}.

\subsubsection{Activity Recognition}

Several algorithms have been used to classify activities.  Investigations into classification of activity include single-sensor classification systems, and classification using multiple sensors.  While multiple sensors provide localized information from various positions on the body (e.g., chest, ankle, knee, etc.), such as in \cite{sagha2011benchmarking}, a single sensor approach is much more pragmatic for the wearer, especially if there are no requirements on sensor positioning \cite{long2009-activity}.

Pattern recognition approaches have been shown to be superior to heuristic based approaches in several investigations.  In these investigations, features are extracted from a single time window, and traditional approaches are used are used to perform classification.  In \cite{allen2006-gmm}, features were extracted from a single accelerometer, namely, separating gravity components, and calculating shifted delta coefficients (SDC, 15 features which estimate time derivative of the signal) and signal magnitude area (SMA, which differentiate between active and restful activities) from the activity, for a total of 25 features.  Using a rule-based heuristic system and thresholding, an accuracy of 71.7\% for three postures and five activities, while a GMM achieved an accuracy of 92.2\%.  Random forests were used to classify five activities from one second windows of accelerometer data \cite{casale2011-activity}, with an accuracy ranging from 88.8\% -- 97.7\%.  This approach relied on only a single accelerometer, and used the best 20 of 319 features to build the classifier.  Five activities were classified using a Naive Bayes classifier and decision tree classifiers in \cite{long2009-activity} using time, frequency and orientation features, with accuracies ranging from 49.4\% -- 95.5\%.  While lower than other approaches described here, the sensor was placed without specific orientation on subjects' wrist, sample rate was lower than other approaches (at 20Hz), and longer time windows (16 seconds) were used.  Additionally, the approach reduces the 19 features to 5 features using PCA.

The previous investigations are based on measurements from a single accelerometer.  These investigations typically leverage data from devices such as smartphones or wrist mounted accelerometers.  Multiple accelerometers, gyroscopes and/or IMUs have been distributed in various locations on the body (e.g., lower and upper arms and legs, torso, etc.).  For instance, the Opportunity dataset \cite{sagha2011benchmarking} provides IMU and/or accelerometer measurements at the upper and lower arms and legs, feet, knees, wrists, hands, hip and back, with the goal of providing a dataset to investigate classifying modes of locomotion (standing, sitting, walking and lying), gestures (e.g., opening a door) and high-level activities (e.g., making breakfast).  Such a dataset is of particular interest for investigation as a robotic material, as each sensor represents a potential location of a computing node.  The authors provide baseline classification accuracies using the mean and variance of sensor measurements over 0.5 second windows as features.  K nearest neighbor (1-NN and 3-NN), nearest centroid classifier, linear and quadratic discriminant analysis were used, with classification accuracies reported for four subjects ranging from 54\%--83\% for mode of locomotion, and 39\%--84\% for gesture. 

The above studied do not consider the long-term temporal aspect of activities.  In addition to classification based on short time windows, sequential models, such as hidden Markov models (HMMs), may be used for activity recognition.  In \cite{mannini2011-accelerometry}, five biaxial accelerometers were used to classify seven low-level activities (e.g., walking, running, standing, etc.) using GMM-HMMs.  Features calculated for each accelerometer axis included the DC component, spectral energy, spectral entropy, and correlation coefficients between two measurement channels.  Individual frames are classified using a Gaussian mixture model (GMM), whose parameters are dependent on the hidden state of the HMM.  The classification from the GMM of the feature vector is then used as an observation for a continuous emission HMM, where the hidden states corresponds to activity.  In addition, individual frames were rejected if likelihood scores were below a threshold.  The GMM-HMM model achieved an accuracy of 95.1\% for seven activities, and an accuracy of 99.0\% for a three activity (sit-stand-walk) dataset.  A comparison with state-of-the-art frame-level classification approaches (GMM, k-NN, SVM, Naive Bayes, ANN, etc.); incorporating a cHMM over several frames as a second phase is shown to improve classification results of these approaches \cite{mannini2010-physical_activity}.

Deep learning approaches have been applied for classifying activities and gaits using individual and multiple on body accelerometers, gyroscopes and/or IMUs.  In \cite{plotz2011feature}, a deep belief network (DBN) was trained by stacking four layers of RBMs to learn features from a 64-sample window of sensor data for activity recognition.  The RBM was built with 192 Gaussian input neurons (64 samples, three axes), two binary hidden layers with 1024 units each, and a final 30 unit binary layer, representing the learned features.  The features extracted using this approach, when used with a nearest neighbor classifier resulted in improved accuracy compared with using statistical, spectral and PCA features.  The authors node that, in addition to this improvement, DBN-based features were less reliant on the quantity of training data when compared with PCA features.  In \cite{zeng2014convolutional}, the authors note that the RBM approach does not capture local dependencies in the time series signal.  The authors use a convolutional neural network to classify activities.  The network consisted of a 1D convolutional layer with 12 filters with size of 20 units, a max pooling layer with pool size of 3 units, two fully connected hiddel layers with 1024 and 30 units, and a softmax classifier layer.  This architecture showed an improvement in accuracy of 1.2\% -- 9.02\% over the results obtained in \cite{plotz2011feature}, which, the authors argue, is due to the invariance to location of scale of learned features.  These two approaches attempt to simply classify activities at each frame.  Similarly, in \cite{yang2015deep}, the authors construct a CNN consisting of two convolutional / max pooling layers (50 filters of 5 units, max pool size of 2 units, 40 filters of 9 units, max pool size of 2 units), followed by a convolutional layer with 20 filters of 3 units, a unification layer and softmax output layer.  The authors report an accuracy of 85.8\%--86.7\% and 95.0\%--96.0\% for two datasets, both of which are an improvement over the results presented in \cite{plotz2011feature}.

In \cite{alsheikh2015deep}, the authors build a DBN as observations for a HMM, in order to capture the temporal aspect of activities.  Using only the DBN, a recognition accuracy of 89.39\% was achieved for a 10 activity dataset, using a single accelerometer.  Adding the HMM layer, and considering a 4 second history of measurements results in a significant improvement in accuracy to 99.13\%.  Finally, in \cite{ordonez2016deep}, the authors build a activity classifier by combining a CNN with an LSTM.  The CNN used consists of four convolutional layers, each with 64 filters and 5 units per filter.  This was followed by two LSTM layers with 128 units each, and a softmax classification layer.  A baseline CNN was also built by replacing the LSTM layers with a fully connected, non-recurrent layer, for comparison.  The CNN-LSTM model showed improvement over both the baseline CNN model, as well as for the results reported in the deep network approaches in the previously mentioned investigations (91.5\% and 95.8\% for the datasets considered).  The authors note that the networks were trained and executed on a GPU; as such, the hardware requirements make such a network unsuitable for wearable applications.

One primary issue with activity recognition is that classifiers are highly dependent on the subjects used to generate the training data, and do not generalize to new users, as the dataset is no longer independently and identically distributed.  In \cite{ravi2005-activity}, classification accuracy is in the range of ~92\% -- ~99\% when training and test data were gathered from the same subject, however, this drops to ~46\% -- 73\% when the test data is gathered from a different subject; even when the data is gathered from the same subject on differing days, the test accuracy drops to ~55\% -- ~90\%.  

\subsubsection{Gait Anomaly}

Wearable devices have also been recently proposed to detect gait anomalies due to hemiplegia, Parkinson's disease, pain in the legs or back, or other related issues, as well as detecting falls in patients \cite{pogorelc2010discovery}.  Detecting gait anomaly in Parkinsonian patients has been explored as a classification problem in several studies.  In \cite{manap2011anomaly}, neural networks were trained to discriminate the walking gait of Parkinson's patients and a healthy control group.  This investigation utilized external cameras and floor mounted force plates to collect data, however, it is feasible to collect the features (joint angles and contact pressures at the foot) using wearable sensors.  Classification accuracy of 81.25\% was achieved using basic or kinematic features; this accuracy was increased to 87.5\% by combining basic, kinetic and kinematic features.  Classification accuracy of 98.18\% was achieved using SVMs \cite{manap2011anomalous}.  Principle components analysis was used to enhance the three sets of features in \cite{manap2013performance}, resulting in improved classification rate for neural network, SVM and Na\"{i}ve Bayes classifiers.  Additionally, by calculating Pearson's correlation coefficient for each feature (comparing normal and Parkinson's cases), the authors extracted four significant features and improved neural network performance to a classification accuracy of 95.63\% \cite{manap2011statistical}.  Neural networks were also used in \cite{gagnon2013qualitative} to assess the risk of a fall, using data collected from pressure sensors embedded in the sole of a shoe.  The investigation used features extracted during the stance phase (i.e., foot in contact with the floor), allowing risk to be assessed at each step, and could accurately assess four levels of risk with an accuracy of 76.6\%.

Fall detection and prevention is an interesting application similar to detecting anomalous gait.  Fall-related injuries are common in the elderly population \cite{chen2010reliable}.  Threshold-based detection algorithms, using accelerometer measurements, have been shown useful \cite{sorvala2012two,chen2010reliable} for detecting falls.  Threshold-based approaches typically consider accelerations above a certain threshold as a fall; such accelerations could be generated from normal activities, such as jumping or jogging.  As an alternative to threshold-based approaches, neural networks were trained on the absolute value and integration of acceleration along three axis to detect falls, and achieved an accuracy of 98.4\%, while being able to avoiding mis-identifying normal, high-acceleration activities as falls \cite{vallejo2013artificial}.

As with activity recognition, most investigations are designed around analyzing gait over a short time frame.  As with activity recognition, incorporating sequential models improve gait anomaly detection results.  In \cite{dutta2009automated}, the cross-correlation of a gait signal, measured from pressure sensors located at the sole of the foot, with a reference signal was used to extract a set of 15 features.  A recurrent neural network trained on these features to identify healthy and pathological subjects demonstrated an overall classification accuracy of 90.6\%, while a feed-forward neural network only achieved 81.2\%, and could identify pathology (ALS, Parkinson's disease, Huntington's disease) with an overall accuracy of 89.8\%, while the feed-forward network could only identify specific pathology with an accuracy of 52.1\%.

\subsubsection{Exoskeleton and Exosuit Control}

Control of exosuits presents a final application of machine learning in wearable devices.  Exosuits have been developed to assist gait by applying force during specific phases of walking gait \cite{wehner2013lightweight, asbeck2015soft, asbeck2015multi}, as medical assistance devices \cite{hong2015softgait}, or rehabilitation assistance \cite{bae2015soft}.

Rigid exoskeletons, which consist of rigid linkage and motor-driven joints, have been developed and tested.  While such suits can assist the wearer a carrying large loads, they have not been shown to improve performance, and actually result in the wearer expending more energy to walk \cite{cornwall2015pursuit}.  In contrast, soft exosuits, which consist of cables or pneumatic actuators supported by form-fitting fabric webbing, provide a lightweight alternative which can boost the walking performance of the wearer.  For instance, a cable system attached to a thigh brace and driven by motors located in a backpack can provide a rotational moment about the hip after the heel strike during walking gait, and can contribute up to 30\% of the biological moment for walking \cite{asbeck2015soft}.  An extension of this system also provided additional actuation to provide a rotation moment about the ankle, resulting in torques of 21\% and 19\% of the biological moments about the ankle and hip, respectively \cite{asbeck2015multi}.  Using pneumatically driven McKibben actuators, a lower-body exosuit which can provide plantarflexion and dorsiflexion of the ankle and flexion and extension of the knee and hip is described in \cite{wehner2013lightweight}.  Evaluation of this suit was limited to actuations which provided assistance with plantarflexion of the ankle, which provided a 10.2\% reduction in the metabolic power while walking, though this reduction in power did not compensate for the increased power requirements due to the weight of the suit.

As most of the research available in soft exosuits investigates mechanical aspects, investigation into sensing and control tends to be limited.  For example, in \cite{asbeck2015soft}, actuation is performed during predefined phases of an assumed constant walking gait, which is estimated by timing of heel strike measured by a foot switch.  Similarly, foot switches were used exclusively in \cite{wehner2013lightweight}.  Detecting the peak angular velocity of a gyroscope was added in \cite{asbeck2015multi}.  In \cite{koller2015learning}, control signals for a soft robotic ankle were generated from the wearer's soleus electromyography (EMG) signals; a smoothed envelope of this signal is used as the actuation signal for the pneumatic actuator.

While control is generally a weakness in existing exoskeleton and exosuits, one approach suggested as promising is the adaptive neural oscillators \cite{young2016state}.  Adaptive neural oscillators are recurrent neural networks, with recurrent feedback from the output to the input \cite{doya1989adaptive}.  These models can fall into a forced, synchronized oscillation with the input signal, or be trained to regenerate the input oscillation without the presence of the diving oscillator.  In \cite{ronsse2011human}, an artificial oscillator is used to control a compliant assistive elbow orthosis.  The adaptive oscillator provides assistive torque by attempting to synchronize the frequency, amplitude and phase of the actuation signal with that of the human movement.  This approach has been extended to learn the angular trajectories of a wearer's hips, and provide assisting torques during walking at various speeds \cite{ronsse2011oscillator_model_free, ronsse2011oscillator}, with the exosuit providing up to an average of 26.38 W/kg of assistance to the wearer.  Adaptive oscillators have also been used for high-level control of an active hip orthosis, in conjunction with a low-level PID controller for torque regulation \cite{giovacchini2015light}.


\subsection{Robotic Skin}

The final application of interest for robotic materials is full-bodied robotic skin.  Though high-density tactile sensors have been in use with robotic fingertips and hands for several decades, full body tactile sensitive arrays for artificial skins have only recently been explored \cite{dahiya2010tactile}.  The desire for full-body tactile sensing largely stems from the shift of applications from a controlled environment (e.g., a factory), where a robot can easily plan and execute tasks due to high certainty about the environment, to more human environments:  the presence of humans or other actors, dynamic variations in the world, limited sensing, large variation of objects types and positions, and compliance and safety constraints require a much higher level of sensing and autonomous behavior \cite{kemp2007challenges, argall2010survey}.  

\subsubsection{Full Body Manipulation}

Object manipulation in robotics has been explored extensively when limited to manipulation with end effectors.  Several full-body, modular tactile sensitive skins have been developed in the last several years \cite{mittendorfer2011humanoid,schmitz2011methods,ohmura2006conformable,tajima2002development}, which are designed to conform to arbitrary robotic surfaces and perform sensing, preprocessing and routing of data to a sink computer.  These systems allow for using tactile feedback for manipulation of large objects with humanoid robots \cite{mittendorfer2015realizing, ohmura2007humanoid}.  While there exists several tactile sensors for robotic applications, most are limited to use with fingertips or hands, and do not exhibit a modular design.  



Most research into modular robotic skin involves sensor design, transduction, low-level processing and architectural considerations, and have not investigated tasks such as object or texture recognition.  However, some initial investigations into full-body manipulation of large objects have included attempting to determine surface texture and weight from   Each arms of a NAO robot was covered with seven Hex-o-Skin cells \cite{mittendorfer2011humanoid} to identify five different surface textures (cardboard, glass, sponge, bubble plastic and sand paper) on objects with two separate weights (500g and 1,500g) \cite{kaboli2014humanoids}.  Data was collected using two separate interactions:  grasping an object with a minimum amount of force to prevent sliding, and reducing the force until the object is allowed to slide.  Features were extracted from one second time windows.  For each sensor, Hjorth parameters (activity, mobility and complexity features, based on the variance of the signal and its derivative) were extracted for each accelerometer axis.  The correlation coefficient for each pair of axes was also computed for use as a feature.  SVMs were used to identify surface texture and weight, with surface classification accuracy in the range of 86\% -- 100\% and weight classification accuracy of 84\%--100\%, depending on features used.  Given the parallels between grasping with a robotic hand and full-body grasping, using machine learning approaches for object recognition, texture recognition and determining grasp stability are interesting areas of research for modular tactile-sensitive skins.

\subsubsection{Affective Touch}

Performing touch-based affective sensing, such as identifying the toucher's emotional state, degree of expressiveness of touch, and/or social context of interaction, has found application in social and therapeutic robots \cite{altun2014-affectivetouch}.  Specific touch-sensitive robots include Robovie-IIS, a humanoid robot utilizing full-body tactile sensor elements \cite{taichi2006automatic}, Huggable, a robot capable of detecting temperature, force, light touch and kinesthetics \cite{stiehl2005design}, and the Haptic Creature \cite{flagg2013affective}, which uses pressure sensitive fabric and conductive and stroke sensitive fur.  Affective touch has also been investigated on humanoid robots, using full-body sensor arrays \cite{kaboli2015humanoids}.

One of the first investigations in recognizing affective touch involved identifying touch using a 44 x 44 element pressure sensitive interface \cite{naya1999recognizing}.  The total load and contact area are calculated at each frame; the peak value of the total load and corresponding total area are used as features.  Using k-nearest neighbor approach, five gestures could be represented with accuracies ranging from 27.3\% (tickle) to 100.0\% (slap).  Tickle and stroke gestures were difficult to distinguish, consequently, the slope at of the total load and area at the time of the peak load was calculated, which provide temporal information enabling these two gestures to be distinguished.

In \cite{taichi2006automatic}, clusters were generated from a set of haptic interactions, using the output of tactile sensors at each time step during an interaction as a point in an individual cluster.  Hierarchical clustering of the interactions was then performed, which allows for the generation of an arbitrary number of clusters; the resulting clusters can be used to describe future interactions with the robot.  Gestures were labeled from video recordings, and the clustering method demonstrated that common gestures (e.g. ``poke the chest'', ``stroke the head'', etc.) tended to fall into a common cluster.  Additionally, gestures with similar locations and intent (e.g., ``poke the chest'' and ''push the shoulder'') were often assigned the same cluster.  This approach is interesting in that it does not assume any predefined set of gestures; rather, gestures could be automatically detected based on previous experiences.

A semantic description tree was developed for 31 possible contact verbs was developed in \cite{koo2008online,kim2010robust}, classifying the actions based on contact time, incidence of a repeat, force, purpose and object used to perform the touch.  Based on this, the authors describe four features (force level, contact time, repeat and change in contact area) for use with classifying four touch patterns.  Measurements from nine contact sensors and an accelerometer, using measurement windows of three different durations (10ms, 40ms and 200ms), were used to extract these features.  A temporal decision tree is used to classify the features in an online manner, producing a classification once sufficient information has been collected, resulting in recognition accuracies of 79.62\% -- 87.83\%.  The same features were used in \cite{yan2013tactile}, and were able to identify three touch patterns in a robotic pet using three pressure sensors.  Classification accuracy of 98.67\% was achieved using a Na\"{i}ve Bayes classifier.

In \cite{tawil2012interpretation}, a pressure sensitive fabric skin was constructed and placed upon an artificial arm.  Features used included the minimum, maximum and mean pressure intensity, minimum and maximum rate of pressure change, and the area, duration, center and displacement of the touch.  Using LogitBoost with decision stumps, eight touch modalities could be identified with an average accuracy of 70.7\% for individual participants; human classification averaged at 90.1\%.

More recently, Hex-o-Skin \cite{mittendorfer2011humanoid} was applied to the chest and back of a NAO robot for identifying nine touch modalities \cite{kaboli2015humanoids}.  Hjorth parameters and correlation coefficients were calculated from accelerometer data, collected over the duration of a gesture, and averaged over cells which were contacted, similar to \cite{kaboli2014humanoids}.  Experiments involved identifying touches at a single location (chest or back) or multiple locations, and while the robot was stationary or in motion.  Classification accuracies were generally high, with some variation based on experimental design:  single touch classification accuracy was 96.79\% with the robot stationary, 94.4\% with the robot in motion, and 92.52\% when trained on stationary data and evaluated on data while in motion.  Multiple touch classification, which relied on training from single touch data, achieved an accuracy of 93.03\%.  These results demonstrate the robustness of the features calculated to touch location and robot motion:  when compared with results using features adapted from \cite{tawil2012interpretation,koo2008online,naya1999recognizing}, results were significantly higher, with reported accuracies of 58.1\% -- 67.3\% for the adapted features.

Using the statistics of a two second time window (minimum, maximum, mean, variance, etc.), 28 features were calculated to identify nine gestures using data gathered from 16 participants \cite{flagg2013affective}.  Gestures were classified using a random forest, neural network, logistic regression and Bayes classifiers, with accuracies of 86\%, 75\%, 72\% and 68\%, respectively.  A preliminary investigation into classifying the CoST dataset used a variety of features, including temporal statistic, size of contact area, histogram-based features, and motion-based features \cite{van2014neural}.  Using these features, neural network was able to achieve an accuracy of 60.2\%.

The increased interest has resulted in two major social gesture corpora being made available:  the Corpus of Social Touch (CoST) dataset \cite{jung2014towards, jung2014touching}, consisting of 14 gestures gathered from a pressure-sensitive fabric, and the Human-Animal Affective Robot Touch (HAART) dataset \cite{cang2015different}, consisting of 7 gestures gathered from the Haptic Creature.  While most of the work has been in the development of inexpensive sensor suites for use in social robotics, there has been some work in automatically identifying the affect social touch.

Several approaches to recognizing gestures in the CoST and HAART datasets from several groups were recently reported \cite{jung2015touch}, with the goal of identifying approaches which may be appropriated from well established fields, such as speech recognition and video analysis.  Features were generally extracted from individual frames (i.e., sensor array data at a single time point), individual channels (i.e., the values of a single sensor measurement over the duration of the gesture), and global features (i.e., a summary of the sensor array data over time).  Several histogram features were developed.  In \cite{gaus2015social}, a coarse histogram with three bins (no touch, low pressure, high pressure) at the frame level and global level, global statistical features of the frames (minimum, maximum, mean, area, variation, etc.), and a multiscale motion histogram were calculated.  In \cite{balli2015recognizing}, individual frames were summarized with the mean, coordinate of the centroid and maximum pressure, and values summarizing an elliptical approximation of the touched area.  From these values, self-similarity and complexity was calculated using Hurst exponent and Hjorth parameters, and coefficients of autoregressive models (2nd to 10th order) were computed to form a feature set.  Global and channel-based statistics, as well as spectral components of the average pressure of each frame, were used as features in \cite{ta2015grenoble}.  In \cite{hughes2015-affective_touch}, autoencoders were used to automatically generate features at the frame level.  Frame level features were also calculated from the first four orders of geometric moments, and global features were computed from the mean, variance and spectrum of the global pressure.

From the computed features, several classification approaches were employed.  Random forests were commonly used to determine the quality of each feature in addition to classifying gestures \cite{balli2015recognizing,gaus2015social,ta2015grenoble}.  Additionally, SVMs were used in \cite{ta2015grenoble}, and HMM models were built for each gesture in \cite{hughes2015-affective_touch} using the autoencoder and geometric moment features as observations.  Final classification accuracies ranged from 26.0\%--61.3\% for the CoST dataset, and 61.0\%--70.9\% for the HAART dataset.  It has been noted that these accuracies may be artificially low, as there is a large amount of confusion between very similar gesture pairs (e.g., scratch and tickle) \cite{jung2015touch}.


\section{Discussion}\label{sec:discussion}

The investigations cited in the previous sections shows that machine learning approaches have found various levels of use in the domains of interest.  In structural health monitoring, fault detection, classification and localization is commonly performed in truss structures, composite panels and multi-story buildings by monitoring the structure's response to vibrations, typically from an external source (e.g., buildings during an earthquake, aircraft panels during flight, etc.).  In addition to monitoring for defects, in the aerospace domain, there is rising interest in applying reinforcement learning to controlling morphable airfoils.  Detecting falls and classifying gait and activities have been explored with on-body, wearable computers; control of soft exosuits presents an application similar to shape control of airfoils.  Finally, full-body sensitive robotic skins can be utilized to assist in navigation and manipulation, and can be used to identify affective gestures in human-robot interactions.  This section describes several potential areas of investigation for intelligent robotic materials, and highlights areas of specific recent research which could be effectively applied to all domains investigated.

\subsection{Data Processing Approaches}\label{sec:data_processing}

\begin{figure}[!htb]
  \centering
  \includegraphics[width=0.5\textwidth]{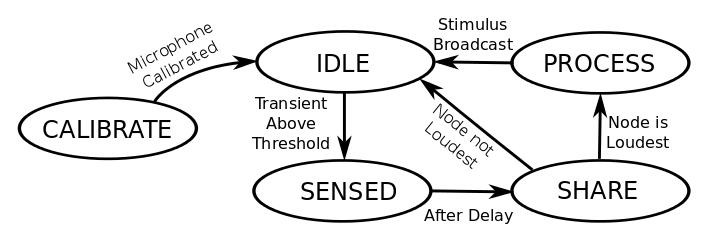}
  \caption{State machine utilized in \cite{hughes2015-skin} to continually detect and localize texture events in a robotic skin.}
  \label{fig:rm_state_machine}
\end{figure}

The literature cited in the previous sections focused primarily on features extracted and pattern recognition algorithms used in the various domains.  While features and pattern recognition algorithms are of primary interest for the investigations in the previous sections, the investigations typically assume knowledge of sensor measurements, and are concerned with performing recognition tasks for a fixed window of time.  With regards to robotic materials, such an approach is not as well suited, for several reasons:  individual nodes need to continually monitor and detect when a signal of interest occurs, nodes perform calculations locally and communicate only with neighboring nodes, nodes which detect this signal need to aggregate measured data (likely from an unknown number of neighboring nodes), and a universal decision needs to be made and communicated to a sink node.  This requires algorithms to determine a means of reasonably executing a task of interest using information from a subset of the nodes.  For instance, individual nodes in a tactile sensitive skin utilize the state machine shown in Figure \ref{fig:rm_state_machine} \cite{hughes2015-skin}.  The nodes in this skin monitor a microphone for signals of interest in an idle state.  When a signal is sensed, it is broadcast through the network after a short delay, and nodes which have also detected a signal of interest rebroadcast this information.  This ensures that the location of the stimulus is only computed by a small, regional group of nodes.  This also has the advantage of allowing a single node to perform the localization task, based on which node detected the signal with the most energy.

\begin{figure}[!htb]
  \centering
  \begin{subfigure}[b]{0.45\textwidth}
    \includegraphics[width=\textwidth]{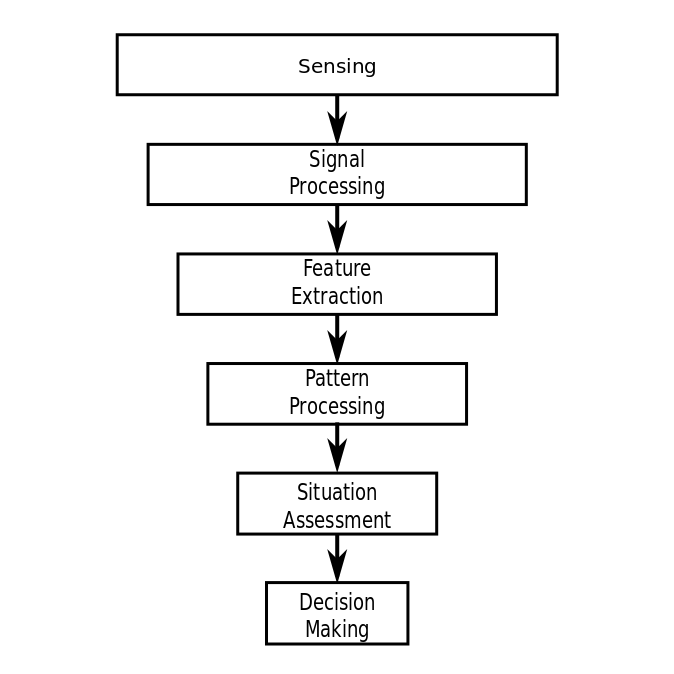}
    \caption{Waterfall model}
    \label{fig:waterfall_model}
  \end{subfigure}
  \begin{subfigure}[b]{0.45\textwidth}
    \includegraphics[width=\textwidth]{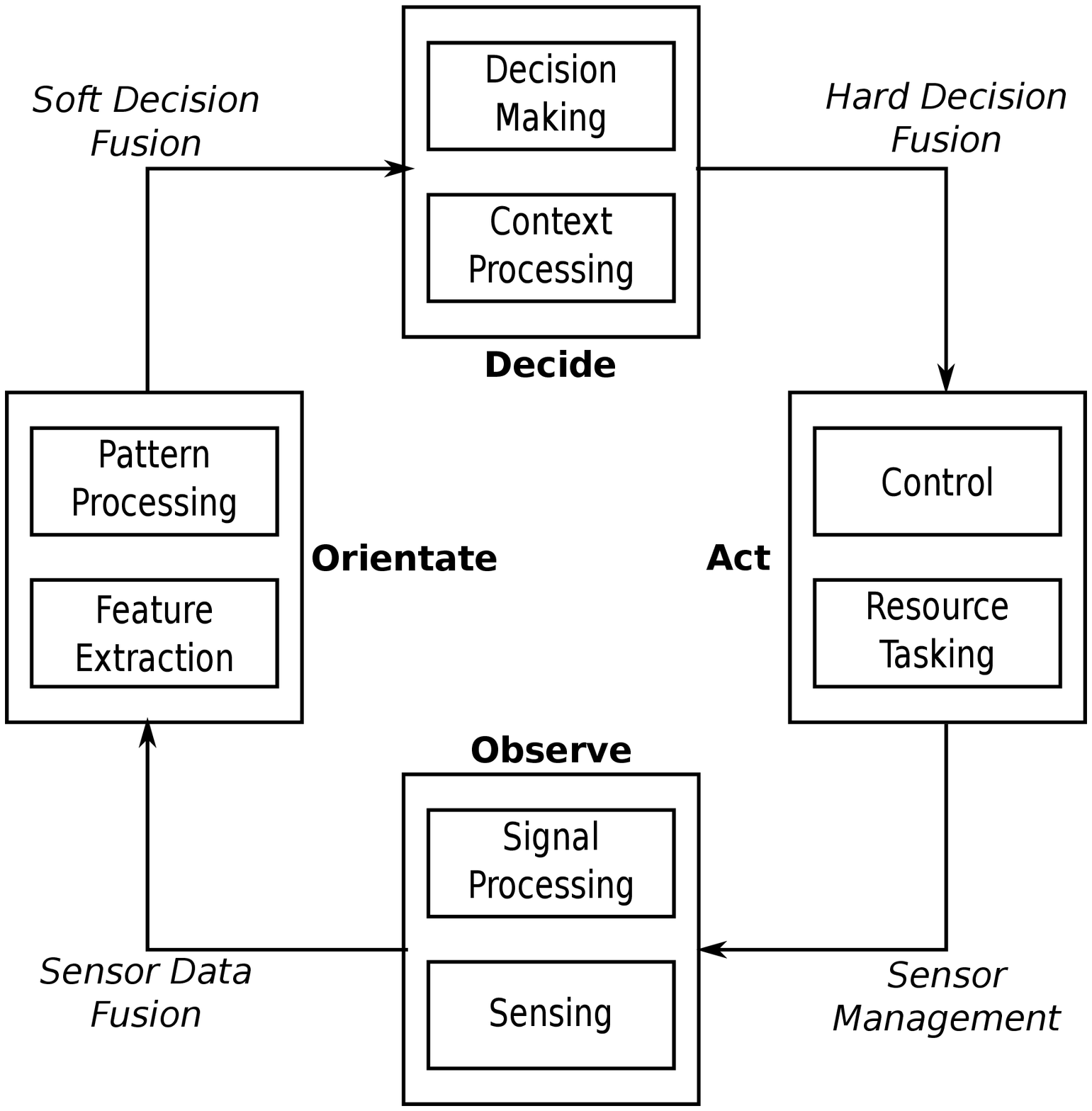}
    \caption{Omnibus model}
    \label{fig:omnibus_model}
  \end{subfigure}
  \caption{Data processing models proposed in \cite{worden2004-intelligentfault}}
  \label{fig:fusion_models}
\end{figure}

As mentioned in \cite{worden2007-ml_shm}, machine learning steps are simply one part of analysis in structural health monitoring.  The full pipeline involves several steps, from low-level features to decision making.  The waterfall model, shown in Figure \ref{fig:waterfall_model}, describes a simple structure which is well suited to several of the tasks described in the literature cited.  This viewpoint assumes operation on a single time window.  \emph{Signal Processing} performs low-level filtering and transforms, such as an FFT, in preparation for calculating features.  \emph{Feature Extraction} computes a vector of features, such as those described in Section \ref{sec:features}.  \emph{Pattern Processing} involves the application of a machine learning algorithm to perform anomaly detection, classification and/or regression.  The final \emph{decision} performed should be reached with higher confidence than using any of the individual sensors.

As an alternative to the waterfall model, the omnibus model, shown in Figure \ref{fig:omnibus_model}, is proposed in \cite{worden2004-intelligentfault}.  This model provides additional advantages over the waterfall model.  The model explicitly includes an \emph{action} stage.  In the context of robotic materials, this may involve various domain-specific actuations:  in structural health monitoring, the underlying material may change stiffness or perform repairs; in morphable aircrafts, SMA skins may be morphed to change the overall shape of an airfoil; artificial muscles may be activated in a wearable exosuit.  A more subtle activity of this stage may involve assessing the quality of individual sensors, and detecting failures, recalibrating or adjusting the relative importance of individual measurements.  The cyclic nature of the model implies that sensor monitoring is performed continuously, is not interrupted by decision or activity, and actions may enhance future measurements.  

As robotic materials are designed to be reactive only to interesting stimuli, the omnibus model provides a much more suitable framework for designing algorithms to run on robotic materials.  The state machine in Figure \ref{fig:rm_state_machine} provides a motivating example for using the omnibus model as an operational framework for the node.  For monitoring and classification / localization applications (such as in \cite{hughes2015-skin}), the model allows nodes to dynamically switch between particular tasks.  During the \emph{Observe} phase, nodes monitor connected sensors and perform low-level processing.  The \emph{Orientation} phase may perform different tasks, depending on which mode of operation the node is in:  anomaly detection may be performed at each time frame until a signal of interest is detected, or localization performed to determine the source of a signal.  The \emph{Decide} phase can be used to switch between different tasks (observation or localization), while the \emph{Act} phase can be used to communicate with neighboring nodes or generate an event packet to be broadcast to a sink node.  For applications which involve changing the underlying structures (e.g., morphable aircraft and exosuit control), the \emph{Act} phase enables control of actuation, and direct feedback to the node and neighboring nodes via the next observation.

\subsection{Use of Temporal Models}\label{sec:temporal_model}

Temporal models, such as dynamic Bayesian networks (including HMMs) and recurrent neural networks (including LSTMs), have been shown to improve classification results for gait recognition \cite{alsheikh2015deep, ordonez2016deep, mannini2011-accelerometry, mannini2010-physical_activity} and affective touch \cite{hughes2015-affective_touch}.  In the domains of structural health monitoring and morphable aircraft, as well as in many of the investigations into recognition of gait and affective touch, many of the investigations rely solely on features extracted from an individual time window, ignoring previous measurements.  In the structural health monitoring domain, structures such as buildings and aircraft panels are excited from continuous, external sources, such as vibrations generated from earthquakes, flights, or dynamic loads (such as traffic on a bridge).  As multiple measurements can be made over time in these domains, improvements in the tasks of interest may be realized in these applications by incorporating temporal models.

In addition to providing improved classification results, temporal models provide state information (either explicitly, as in HMMs, or encoded in hidden units, as in RNNS), which evolves after each measurement based on the current state and observations.  From the perspective of the omnibus model, state information can also be used in the \emph{Decide} and \emph{Act} phases, while the \emph{Orientate} phase may rely solely on observations.  This could have several benefits.  For instance, features can be automatically extracted from training data for each individual frame independent of pattern processing, decision making or actions.  Additionally, determining correct decisions and actions can be learned based on the current state.  Separating these two phases in this manner then enables the spatial and temporal aspects of learning to be encapsulated solely in local state.  For example, local state could be encoded using a mixture model or a dynamic Bayesian network, and at each time step, nodes communicate local state with neighboring nodes, and update its own state using current observations.  State can be trained using a distributed approach, as described in Section \ref{sec:sensor_network_gmm} or Section \ref{sec:sensor_network_pgm}, for example, while the remaining training could be performed using non-distributed approaches described in Appendix ~\ref{sec:appendix_machine_learning}.

\subsection{Automatic Feature Extraction}

The features used in most of the investigations are based on either common mathematical operations (e.g., statistical or spectral features), or domain-specific heuristics (e.g., time of heel strike).  Recent investigations in gait recognition and affective touch recognition have used deep learning approaches to automatically extract features from provided training data, including restricted Boltzmann machines, \cite{plotz2011feature}, convolutional neural networks \cite{zeng2014convolutional,yang2015deep}, and autoencoders \cite{hughes2015-affective_touch}.  These approaches, explored extensively in image classification \cite{krizhevsky2012imagenet} and speech recognition \cite{graves2013speech}, have provided approaches to extract features from training data which are robust to variations in the scale and spatial or temporal position of features.  Image classification and speech recognition applications demonstrate the applicability of automatic feature learning for spatial and temporal aspects of signals, respectively.  Such approaches to feature learning could find application in the various domains investigated in this paper.

\begin{figure}[!htb]
  \centering
  \includegraphics[width=0.8\textwidth]{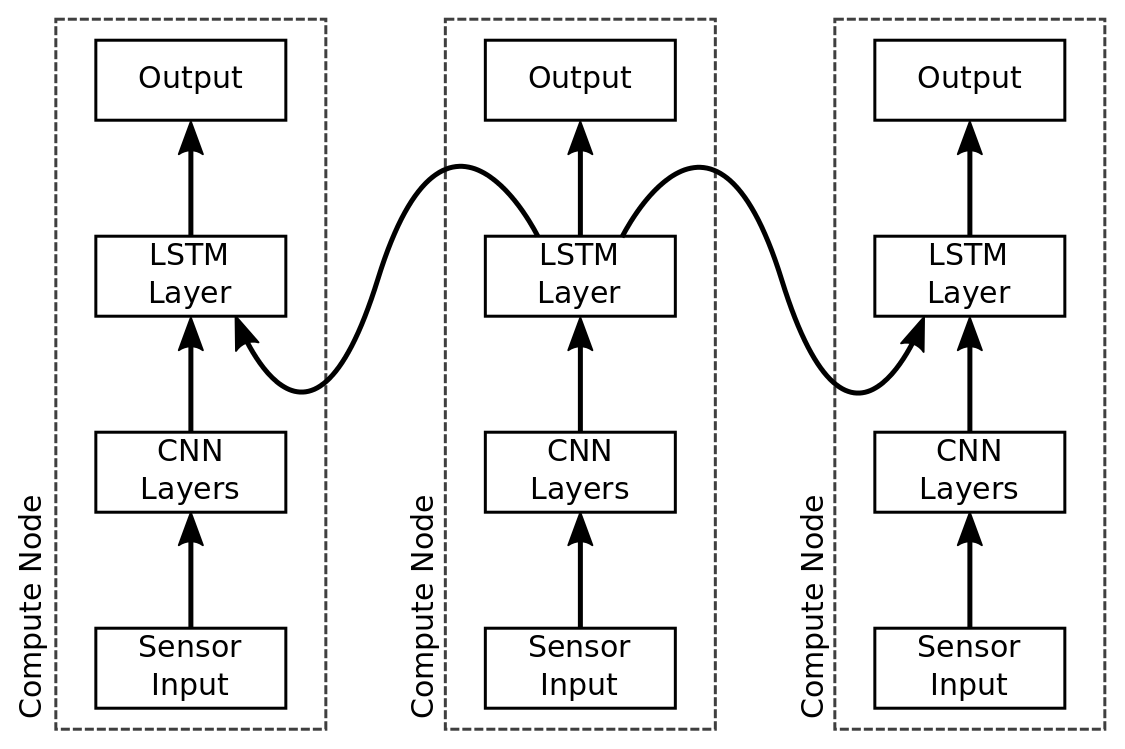}
  \caption{Potential implementation of CNN-LSTM model in a robotic material}
  \label{fig:rm_cnn_lstm_model}
\end{figure}

The CNN-LSTM model described in \cite{ordonez2016deep} provides a particularly interesting architecture for robotic materials.  In addition to providing a means of learning features directly from a training set, the LSTM portion of the model allows for continual monitoring of temporal signals, as described in Section \ref{sec:temporal_model}.  This architecture is not readily adapted to sensor networks, as the number of neighboring nodes in a sensor network varies based on the local network density and communication range.  In contrast, nodes in robotic materials have a fixed number of neighbors, determined at time of construction.  The current state (LSTM nodes) of the LSTM model in a node may be communicated to neighboring nodes.  A node may then use the received states as input to its own LSTM model, in addition to the measurements.  The proposed implementation is shown in Figure \ref{fig:rm_cnn_lstm_model}.  

\subsection{Training}

Traditionally, training algorithms involved building or adjusting a model given a training dataset.  In robotic materials, each individual node maintains a model locally.  Consequently, training may not be performed in as direct a manner:  the data from all sensors are not available for each node, communication introduces delays between measurements from one node and processing in another, and packet loss and node failure may result in missing or erroneous inputs.  Training sets can be collected from physical simulations, if easily performed, or from a physical implementation of the material.  Additionally, training can be performed on-line or off-line.

If a material or structure, and corresponding stimulus, are easily simulated, then training can be performed based on these simulations.  This has been performed in several of the investigations previously mentioned, for instance \cite{johnson2004-shm_benchmark}.  Simulations allow for rapid generation of large amount of data for many situations, including those which are prohibitively costly.  This approach suffers from several drawbacks.  Primarily, simulated data may suffer from a ``reality gap'', where important aspects of the system are not properly simulated.  Additionally, measurements among several nodes in robotic materials are typically not synchronized, which may not be easily represented in simulated data.  However, for iterative training algorithms (e.g., gradient-based optimization), simulated datasets may provide an approach to 

Alternatively, training data can be collected from each node under a variety of situations, from which training can be performed off-line on a centralized computer.  This becomes plausible as the computing power (memory, clock speed, etc.) of individual nodes in a robotic material are several orders of magnitude below a computer on which training would be performed.  For materials with a large number of nodes, collecting data can be hindered by communication bandwidth to sink nodes or the amount of memory available on the node.  If sensor values cannot be collected by a sink node before the next set of measurements are collected, then nodes need to buffer measurements, limiting the length of time frames in sequential datasets.  However, this does provide more accurate training data than simulated results.  Additionally, several frameworks have been developed for learning using graphical processing units (GPUs), including GPU implementations of neural networks \cite{bahrampour2015comparative}, SVMs \cite{catanzaro2008fast} and Bayesian networks \cite{linderman2010high}.  This approach allows for a \emph{global} model to be built as a collection of modules, where each module would be implemented on an individual node, and trained using an appropriate framework.

On-line approaches, where training is performed directly on the network of nodes, have been discussed in Section \ref{sec:ml_robotic_materials}, providing algorithms for distributed learning for SVMs, mixture models and graphical models.  On-line learning in robotic materials has two major drawbacks, the severity of which is based on the microcontroller used.  First, microcontrollers are much slower than typical microprocessors (several dozen MHz versus several GHz), resulting in training time taking much longer (in absolute time) in the robotic material.  Second, the amount of RAM available on microcontrollers are typically much less than the available EEPROM (e.g., 8kB RAM versus 120kB EEPROM on XMega128 chips).

\begin{figure}[!htb]
  \centering
  \includegraphics[width=0.5\textwidth]{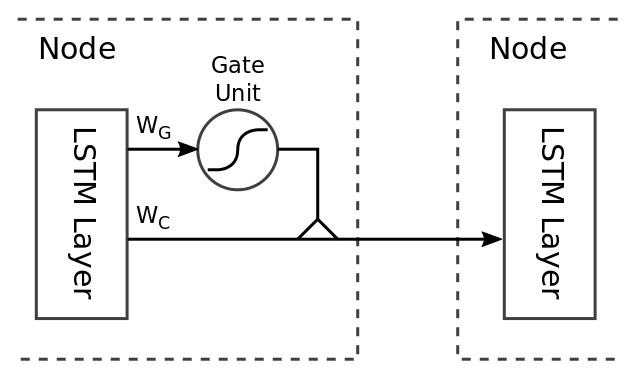}
  \caption{Communication architecture for off-line CNN-LSTM models.}
  \label{fig:communication_architecture}
\end{figure}

In order to apply off-line training to a network of nodes, the training model needs to consider various aspects unique to a network of nodes.  For instance, each node's clock cannot be easily synchronized with other nodes in the network.  The sensing and update rate of the models described in the literature are much slower than the clock rates of the microcontrollers used in the node.  However, it still may be worthwhile to investigate the effect of performing measurement and model updates with differing clock rates in the nodes, as the difference between the clock rates will ultimately cause a potentially significant shift in measurements or state updates.  A second consideration is modeling communication.  As mentioned previously, communication could be unreliable, or have a limited communication rate.  An off-line model would need means to simulate such communication aspects.  Building upon the CNN-LSTM model described earlier, Figure \ref{fig:communication_architecture} demonstrates a potential approach using gates similar to those used for learning, recall and forgetting in LSTM layers.  Using a KL-divergence objective, such as that used with sparse autoencoders \cite{ngiam2011optimization}, the gate can be trained to be active a certain percentage of the time, and consequently control the rate of communication.  

\subsection{Hardware Constraints}

Hardware constraints provide a final limitation to algorithms running on robotic materials.  With regards to machine learning approaches, memory limitations are the primary limitation to the capabilities of a trained model.  For instance, the maximum number of parameters in neural networks or the number of random variables and dependencies in a graphical model will be limited by the available memory in the microcontroller.  This limitation has been discussed in \cite{ordonez2016deep}, where a CNN-LSTM model can be trained to classify gaits with a high level of accuracy, but the size of the resulting model is too large to be reasonably used in a wearable system.  In contrast, by distributing a similar model among a network of nodes as a robotic material, the size of input for an individual node is greatly reduced.  In this way, robotic materials may be viewed as a potential approach to addressing the ``curse of dimensionality''.  This viewpoint provides an interesting concept for investigation:  to what extent can a model be modularized and parameters reduced while maintaining a reasonable level of quality compared with a model built on the full input feature?

\subsection{Robotic Applications}

A final consideration is to incorporate the suggested applications in Section \ref{sec:ml_in_materials} into robotic applications.  Essentially, this involves offloading several robotic tasks of interest into the material itself, allowing an external controller to query the material or have the material push pertinent information to the controller in an event driven manner, such as with the robotic skin in \cite{hughes2015-skin}.  For instance, a robotic skin can perform structural health monitoring tasks (anomaly detection, localization and assessment) to assess wear and identify failure of joint actuators in the robot.  The benefit to this approach is that the skin can perform this task passively during normal robot application by monitoring vibrations induced from the actuators, for example, as opposed to requiring the robot to actively performing some sequence of actions and observations to determine the state of the actuators.  Gait recognition, fall detection and exosuit control have direct correlation with generating stable gaits for bipedal robots.  In addition, The reinforcement learning approaches used for optimal shape control of airfoils has potential for use in the field of soft robotics \cite{rus2015design}.  While there has been much research into the fabrication and system level applications of soft robots, modeling and controlling the kinematics and dynamics of a soft robot remains challenging.  Enabling the robot to learn a correct sequence of actuations to achieve a pose or trajectory may solve these difficulties.  As robotic nodes have been successfully embedded in soft robotic skins, the same hardware can be leveraged to learning and executing control tasks.


\section{Conclusion}\label{sec:conclusion}

This paper investigates the potential of incorporating machine learning approaches with inexpensive computing nodes as a design approach for intelligent robotic materials.  As the investigation is interested in applying machine learning techniques to currently existing and proposed robotic material applications, much of the paper is based on machine learning literature in various domains, namely structural health monitoring, aerospace, wearable computing, human activity monitoring, and robotic skin.  The paper provides the background to and an overview of the current state of research in robotic materials in Section \ref{sec:robotic_materials}.  Section \ref{sec:ml_robotic_materials} describes features and machine learning algorithms used in the reviewed literature, and which show promise for use with robotic materials.  Specifically, approaches to learning and using a machine learning model in a distributed manner are explored in Section \ref{sec:distribute_approaches}.  While much of this research is based on investigations in the wireless sensor network domain, they still are applicable for intelligent robotic materials, either directly, or as insight into similar algorithms.  Section \ref{sec:ml_in_materials} reviews machine learning applications in structural health monitoring (detection, localization and identification of defects), morphable aircraft (reinforcement learning of optimal shape), wearable computing (gait and activity recognition, fall detection, exosuit control) and robotic skin (full body manipulation and affective touch recognition).  Finally, potential areas of research are discussed in Section \ref{sec:discussion}.

The reviewed literature focused specifically on applications which would be well suited for a robotic material paradigm, namely, in situations where a large number of small, networked computing elements are more suited to the application than a centralized system.  Primarily, this is due to the large number of sensors and actuators required to monitor and adjust the underlying structure.  This may be due to the scale of the structure, such as with buildings, bridges or aircraft surfaces, due to the desired sensor / actuator density, such as with robotic skins, or due to the bandwidth of individual sensors, such as measuring vibration using piezoelectric elements or microphones.  A secondary reasoning involves the need to minimize weight or power consumption, such as with wearable computing.  The underlying commonality with all these domains is that the actual computing elements are easily embedded into the underlying material.

The main focus of this paper is to demonstrate the need for a common underlying approach to incorporating machine learning with robotic materials.  This has been guided by the machine learning approaches and tasks present in the reviewed literature, as well as machine learning algorithms adapted for wireless sensor networks.  The distributed network approach and reactive nature of robotic materials implies that machine learning approaches will have both spatial and temporal aspects to processing.  This reflects the underlying physical process:  many of the tasks described in the literature involves responding to vibrations which propagate through the structure; monitoring the movement and contact area of an individual's hand on robotic skin is also a common tasks where the spatiotemporal aspect of the signal is highly evident.  Consequently, this paper argues for using a temporal model, such as dynamic Bayesian models or recurrent neural networks, as a critical component for use in individual nodes, as discussed in Section \ref{sec:temporal_model}, with communication between nodes being used to share preprocessed information to incorporate in state updates.  

Data processing approaches are an ancillary consideration and are presented in Section \ref{sec:data_processing}.  Much of the reviewed literature consider only performing a machine learning task in isolation, e.g., classifying a gait or gesture from a fixed duration of measurements.  Unfortunately, from a system-level perspective, data is not presented in such a direct manner to robotic materials.  Rather, nodes must constantly monitor and decide when and how to act and communicate with neighboring nodes.  This section argues for a system-level architecture which monitors data for novel or anomalous local measurements, upon which the node utilizes the more complex temporal model which shares information with neighboring nodes described in Section \ref{sec:temporal_model}, allowing for an approach which involves local monitoring of sensors, and performing responses in an event-based manner.

Finally, the literature reviewed shows that a common approach to machine learning--extracting a feature set from measurements, and perform classification or regression on static or quasi-static data using existing data mining software (such as Weka \cite{hall2009weka})--is employed in many of the papers in all the domains.  However, some papers demonstrate an advantage to employing approaches that have not been explored in other domains.  For instance, geometric probabilistic decision trees used in \cite{mechbal2015-probabilistic_svm} demonstrates improved performance over a ``flat'' multi-class classifier.  Several activity recognition papers demonstrate improved performance when temporal models (specifically, HMMs and LSTMs); the same approaches may find use with robotic skin applications.  The anomaly detection approaches used in Section \ref{sec:damage_detection} could also be useful for detecting anomalous gaits.  Finally, convolutional neural networks were used to automatically extract features for activity recognition; such approaches can be used in the other domains to extract shift and scale invariant features from time-series data.  Incorporating these approaches could provide significant advances in using machine learning approaches in the other domains.


\appendix

\section{Overview of Machine Learning Concepts}\label{sec:appendix_machine_learning}

Several machine learning approaches have been applied to the various fields described in Section \ref{sec:ml_in_materials}.  While specific tasks, models and feature selection vary among the applications, the desired tasks fall broadly into one of the following categories:  classification (identifying which class a set of measurements belongs to), regression (fitting a function to example input-output pairs), and anomaly detection (identifying when a particular measurement is abnormal).  The following subsections describe potential models for use in robotic materials.  These include models and algorithms used in the fields in Section \ref{sec:ml_in_materials}, as well as models and algorithms which have been adapted for sensor networks, as discussed in Section \ref{sec:distribute_approaches}.  An exhaustive survey of the following approaches is beyond the scope of this paper.  However, the basic approaches are summarized here, and described where used in the literature.  All approaches described assume that the data is described by an $N$ dimensional feature vector, $\mathbf{x} = [x_1, x_2, ... x_N]^T$, and produce an output $\mathbf{y} = [y_1, y_2, ... y_M]^T$.  

\subsection{Novelty Detection}\label{sec:appendix_anomaly_detection}

Novelty detection involves a system learning to identify data or features which have not been previously experienced.  Novelty detection approaches are based on collecting data under normal operating condition or state, and determining if novel measurements lie outside some threshold of normal operation or state.  The variation from normal operation is referred to as novelty indicator (NI), which is a scalar measurement indicating the deviation from nominal state.  

\subsubsection{Mahalanobis Squared Distance}

Mahalanobis squared distance (MSD) assumes the training features can be accurately represented by a multivariate Gaussian distribution.  The mean and covariance matrix of the training data, $\mathbf{\hat{x}}$ and $\Sigma$, are calculated.  From this, the NI for a new feature vector, $\mathbf{z}$ is defined as

\begin{equation}
  NI(\mathbf{z}) = (\mathbf{z} - \mathbf{\hat{x}})^T \Sigma^{-1} (\mathbf{z} - \mathbf{\hat{x}})
\end{equation}

\subsubsection{Kernel Density Estimate}

Kernel density estimate (KDE) estimates the probability density function of the nominal state, and determines the probability that new feature vectors are outside some threshold of this distribution.  The probability of a feature vector being generated from a nominal state is

\begin{equation}
  p(\mathbf{z}) = \frac{1}{N h} \sum_{i=1}^N K \left( \frac{\mathbf{z} - \mathbf{x}_i}{h} \right)
\end{equation}

where the kernel function, $K$, is a localized function, with multivariate Gaussian kernels being a common choice.  $h$ represents a smoothing factor, the value of which can be computed by cross-validation.

\begin{equation}
  K(\mathbf{x}) = \frac{1}{(2 \pi)^{d/2}} e^{-\lVert \mathbf{x} \rVert^2 / 2}
\end{equation}

\subsubsection{Autoencoder}

Training data can be used to build an autoencoder, such that the output of the autoencoder, $\mathbf{\hat{X}}$, accurately reconstructs the training data.  Once trained, the reconstruction of a new feature vector, $\mathbf{\hat{z}}$, can be computed from new features.  The NI is then simply the norm of the error (i.e., difference between the original and reconstructed feature vector)

\begin{equation}
  NI(\mathbf{z}) = \lVert \mathbf{z} - \mathbf{\hat{z}} \rVert
\end{equation}

\subsubsection{Factor Analysis}

Factor analysis describes correlations between observed features and a small number of unobserved independent variables (factors) \cite{figueiredo2011-ml_damage_detection}.  A set of training data, $\mathbf{X}$, can be represented as the product of a matrix of factor scores $\mathbf{\Xi}$, and a matrix of factor loadings, $\mathbf{\Lambda}$

\begin{equation}
  \mathbf{X} = \mathbf{\Lambda} \mathbf{\Xi} + \mathbf{E}
\end{equation}

and $\mathbf{E}$ represent error terms.  The factor loadings can be calculated from covariance of the training data, $\mathbf{\Sigma}$

\begin{equation}
  \mathbf{\Sigma} = \mathbf{\Lambda} \mathbf{\Lambda}^T + \mathbf{\Psi}
\end{equation}

where $\mathbf{\Psi}$ is a diagonal matrix representing the variance of each element in the training data.  $\mathbf{\Lambda}$ can be calculated using several methods, such a maximum likelihood estimation.

The error term for a measurement made from a novel state is assumed to be much larger than one made from a nominal state.  The factor loadings of a novel measurement, $\mathbf{\hat{\xi}}$, can be computed using linear regression

\begin{equation}
  \mathbf{\hat{\xi}} = \mathbf{\Lambda}^T \left( \mathbf{\Psi} \right)^{-1} \mathbf{z}
\end{equation}

From which the residual error can be calculated

\begin{equation}
  \mathbf{e} = \mathbf{z} - \mathbf{\Lambda} \mathbf{\hat{\xi}}
\end{equation}

from which the NI can be calculated as the norm of the residual error, as with the autoencoder NI.

\subsubsection{Singular Value Decomposition}\label{sec:appendix_SVD}

Singular value decomposition (SVD) decomposes a matrix of measurements, $\mathbf{M}$ into the product of two orthonormal matrices, $\mathbf{U}$ and $\mathbf{V}$, and a diagonal matrix of singular values, $\mathbf{\Sigma}$, as

\begin{equation}
  \mathbf{M} = \mathbf{U} \mathbf{\Sigma} \mathbf{V}^T
\end{equation}

Novelty detection can be performed using nominal measurements (i.e., training data) by estimating the rank of the measurements using SVD \cite{ruotolo1999using}.  Measurements from nominal states maybe be assumed to be a linear combination of a finite number of characteristic vectors, which represent particular properties of the underlying system which generated the measurements.  The rank of the training matrix is simply the number of non-zero singular values.  In the presence of additive noise, the training data matrix would full rank, the noise-free rank can be estimated by only considering singular values above a predefined threshold.

To detect the presence of novel measurements in a new feature vector, the feature vector is appended to the training data, i.e., $\mathbf{M} = [ \mathbf{X}, \mathbf{z} ]$, after which the rank is estimated using SVD.  Without a novel characteristic, the new feature vector can be considered a linear combination of the characteristic vectors in the training data.  Novel measurements would contain a novel characteristic vector in the matrix, increasing the estimated rank due to the novelty.

A general novelty indicator can be formulated as

\begin{equation}
  NI(\mathbf{z}) = \alpha \left( \prod_{i=1}^{r+1} \sigma_k(\mathbf{M}) \right)
\end{equation}

where $\alpha$ is a normalizing factor, and $\sigma_k$ are the non-zero (or above threshold) singular values.  Unlike using the rank of the matrix as a novelty indicator, this approach also accounts for the presence of multiplicative noise in the measurement data.

\subsection{Neural Networks}

\begin{figure}[!htb]
\centering
\includegraphics[width=0.75\textwidth]{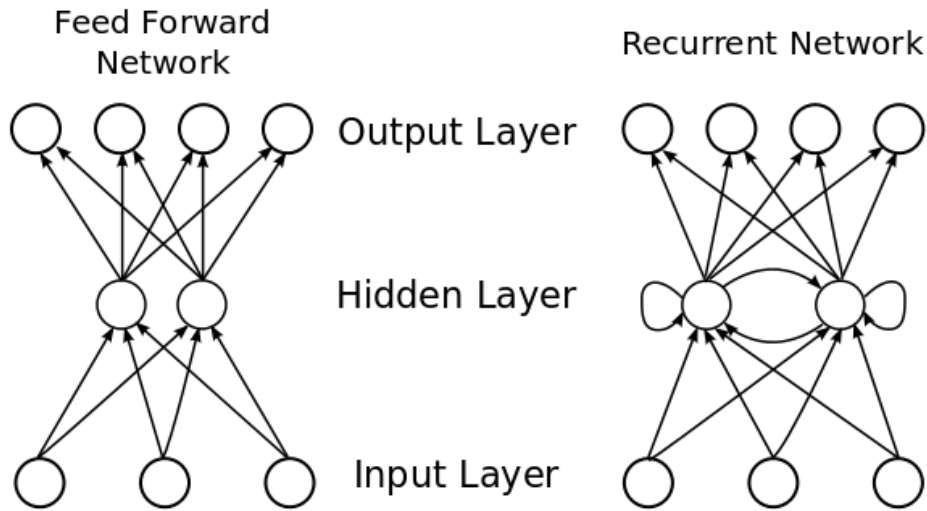}
\caption{Feed forward and recurrent neural networks.}
\label{fig:neural_network}
\end{figure}

\begin{figure}[!htb]
  \centering
  \includegraphics[width=0.4\textwidth]{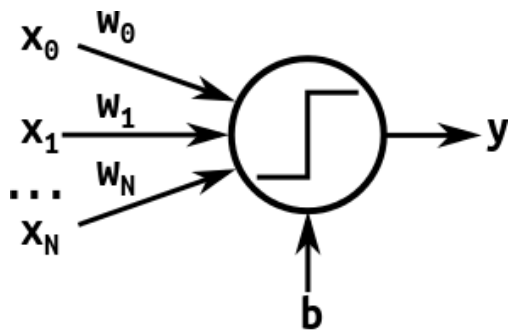}
  \caption{Individual neuron in a neural network.}
  \label{fig:perceptron}
\end{figure}

Artificial neural networks are computational models inspired by the behavior of biological neurons.  Neural networks consist of layers of neurons and directed connections between them, as shown in Figure \ref{fig:neural_network}.  An individual neuron is shown in Figure \ref{fig:perceptron}.  Given an input vector, $\textbf{x}$, the output of a neuron, $y$, is calculated by computing the net input to the neuron (the weighted sum of the input and a bias term), and applying the activation function, $\phi$, of the neuron to the net input, i.e.,

\begin{equation}\label{eq:net_input}
  y = \phi(b + \sum_{i=1}^N w_i x_i)
\end{equation}

\begin{figure}[!htb]
  \centering
  \includegraphics[width=0.5\textwidth]{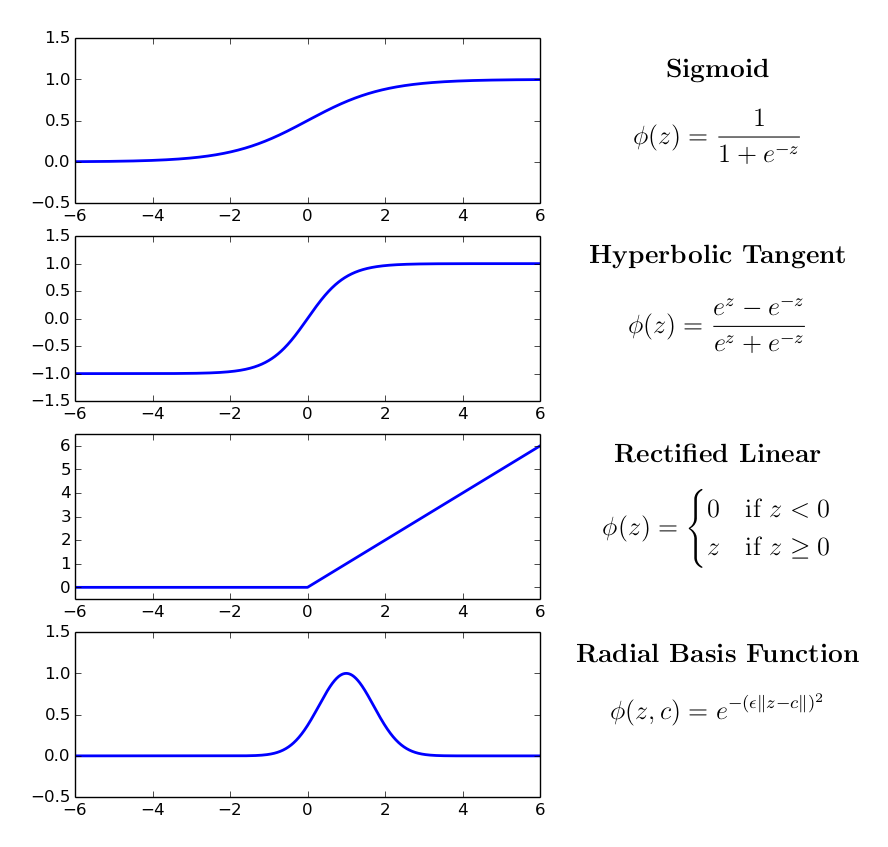}
  \caption{Common neural network activation functions.}
  \label{fig:activation_functions}
\end{figure}

The output of neurons may then be used as the input of other neurons.  Common activation functions, shown in Figure \ref{fig:activation_functions}, include sigmoidal functions, such as the logistic or hyperbolic tangent functions, radial basis functions, and rectified linear functions.

Depending on the activation function used in the output layer, the neural network can be used for regression or classification.  For regression, using linear activation functions can be used, rectified linear units can be used for strictly positive outputs, or sigmoid or hyperbolic tangent layers can be used for bounded outputs.  Classification is typically performed using a sigmoid function to represent the probability of belonging to a particular class.  Softmax layers, which represent a multinormal probability distribution and provides a useful output for classification tasks, are constructed by normalizing the output of a layer of logistic units.

Training neural networks involves minimizing an objective function, such as the mean squared error (MSE) between the neural network output and target values, or the cross-entropy error between the target class and predicted class.  Using the backpropagation algorithm \cite{werbos1974-backprop, rumelhart1986-backprop}, the gradient of the error with respect to the parameters (weights and biases) of the network can be calculated, and updated using gradient descent or other update rule.

Three particular architectural components are interesting for the applications described in this paper:  convolutional networks provide an efficient means to process spatial data, recurrent networks allow for processing of temporal data, and autoencoders provide allow trainable, nonlinear compression of data.

\textbf{Recurrent Neural Networks.}  Feedforward neural networks are commonly used for tasks which perform classification or regression using only a single input vector.  For sequential or time-series data, a finite number of frames in the sequence can be concatenated for use as input.  This solution requires a fixed history duration, and increases the size of the input (and consequently, the number of network parameters)  Recurrent Neural Networks (RNNs) consist of networks which contain cycles, which provide models which are more suitable for sequential or time-series models.  Recurrent connections allow the network to maintain and update an internal state at each step in the sequence, providing a suitable architecture to handle temporal aspects of a signal.

\begin{figure}[!htb]
  \centering
  \includegraphics[width=0.5\textwidth]{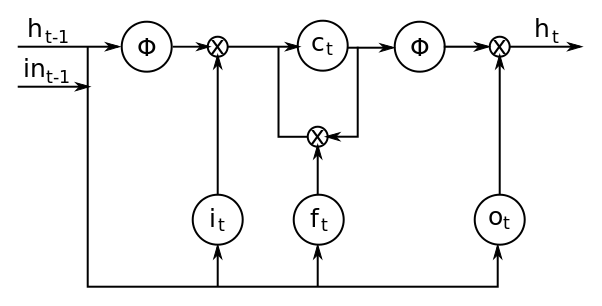}
  \caption{A long-short term memory (LSTM) cell.}
  \label{fig:LSTM_layer}
\end{figure}

A major limitation to recurrent neural networks is the ability to learn tasks where there exists a long time delay between the effect from some input signal and the corresponding output.  The network is unable to learn with large time delays due to the nature of backpropagaing the error signal through time during training \cite{hochreiter2001-rnngradient}.  Essentially, as the error is propagated backward through time, it either vanishes or blows up, depending on the recurrent weights.  A promising solution to this problem is to incorporate Long Short Term Memory (LSTM) \cite{hochreiter1997-LSTM}.  A LSTM component, shown in Figure \ref{fig:LSTM_layer}, can store values using a recurrent neuron and perform operations on the value of this neuron using input, forget and output gate neurons.  These values may then be read, written to, or forgotten, based on the activation of input, output or forget gates.  Using these, a network can learn both \emph{what} to write into memory, and \emph{when} to read and write it.

\begin{figure}[!htb]
  \centering
  \includegraphics[width=0.8\textwidth]{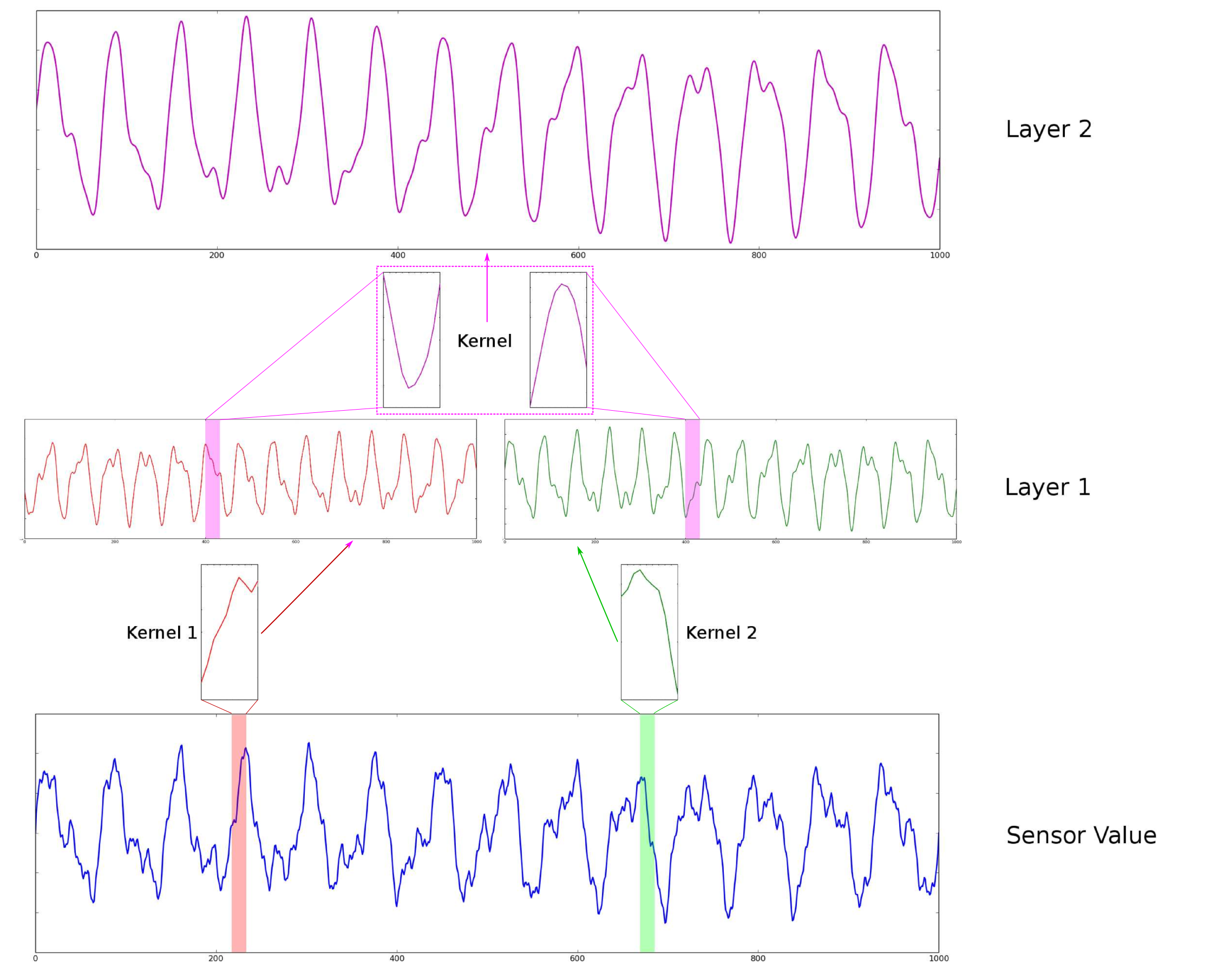}
  \caption{Temporal convolution over a single sensor.}
  \label{fig:convolutional_network}
\end{figure}

\textbf{Convolutional Neural Networks.}  Convolutional layers in neural networks utilize a weight sharing scheme to allow a network to learn local features which are invariant to translation and scaling.  This is achieved through the use of convolutional layers, which convolve a feature map with an input space, and pooling layers, which reduces the size of a layer by selecting the activation from a small region.  An example of convolutional layers applied to a temporal signal is shown in Figure \ref{fig:convolutional_network}.  In this example, the second layer is generated by convolving the original signal with two kernels, while the third layer is generated using both channels from the second layer.  Convolutional networks have been shown to be especially useful for image recognition \cite{lecun1995-cnn}.  The main advantage of convolutional networks is the use of weight sharing:  sets of weights are replicated among multiple connections.  Replicating weights in this way enables the network to detect features regardless of the position of the feature in signal.  Additionally, shared weights reduce the number of parameters in a model, which reduces the training time and required number of training examples, and allows the model to generalize better when compared to fully connected networks.  Finally, convolutional networks utilize the structure of input data--images have strong 2D local structure and time series have strong 1D local structure, where nearby variables are highly correlated.  In contrast, the input of feed-forward models can be permuted in an arbitrary (though constant) manner without affecting the behavior of the network.

\textbf{Autoencoders.}  One application of feed-forward neural networks which has demonstrated a variety of uses is the autoencoder.  The purpose of this type of network is to generate an efficient encoding of samples in a dataset, typically by generating a representation with smaller dimensionality than the original dataset.  Autoencoders are built and trained to perform the identity function--the output of the network is trained to reproduce the provided input.  A ``code'' layer of neurons, whose size is smaller than the input, provides an means of reducing the dimensionality of the data.  In this way, autoencoders behave in the same way as principle components analysis (PCA).  Because of the nonlinearity of the architecture, autoencoders with a single hidden layer have been shown to provide superior reconstruction of the input signal, when compared with principle components analysis (PCA) \cite{japkowicz2000-nonlinear}.  

Deep autoencoders, which contain several nonlinear hidden layers between the input and encoding layer, have been used for several applications \cite{hinton2006-autoencoders}.  Deep autoencoders have shown the ability to compress data to very small sizes.  For images, 625 pixel (25 x 25) gray scale images of faces were compressed to a 30-unit code, the reconstructed images showed great improvement over encoding using PCA; similarly, 784 pixel (28 x 28) pixel digits from the MNIST dataset can be compressed to a 10-unit code, with a reconstruction error of 1.2\%.

\subsubsection{Support Vector Machines}

\begin{figure}[!htb]
\centering
\includegraphics[width=0.5\textwidth]{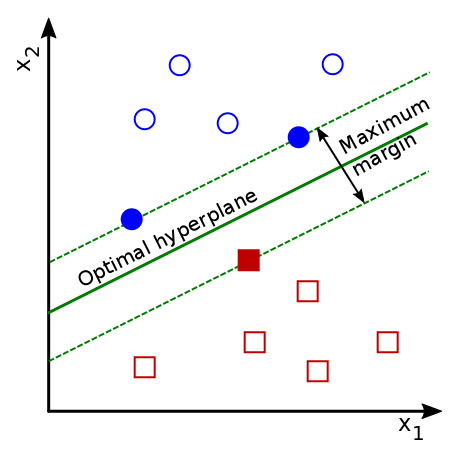}
\caption{Support vectors and max-margin hyperplane.}
\label{fig:svm_max_margin}
\end{figure}

Support Vector Machines have been extensively used for classification and regression tasks.  In the original formulation, an SVM is a binary classifier trained on a linearly separable dataset, which is constructed by finding an optimal hyperplane which separates the two classes in the dataset \cite{cortes1995-svm}.  The optimal hyperplane is constructed to maximize the margin between positive and negative training examples, as shown in Figure \ref{fig:svm_max_margin}.  The max-margin hyperplane is determined by a set of \emph{support vectors}, the subset of the training data which is nearest to the separating hyperplane.  Nonlinear separation is possible by using nonlinear kernel functions to map the original features to a kernel space, and performing linear separation on the mapped features.  Several approaches are available to extend SVMs to multi-class classification.  Multiple SVM classifiers can be build, each of which classifies between one of the labels and the rest (\emph{one-vs-all}), or by classifying between pairs of classes (\emph{one-vs-one}), and selecting the final label based on the classifier with the highest output, or by selecting the label based on the highest number of output from the group of classifiers, respectively.

\begin{figure}[!htb]
\centering
\includegraphics[width=0.5\textwidth]{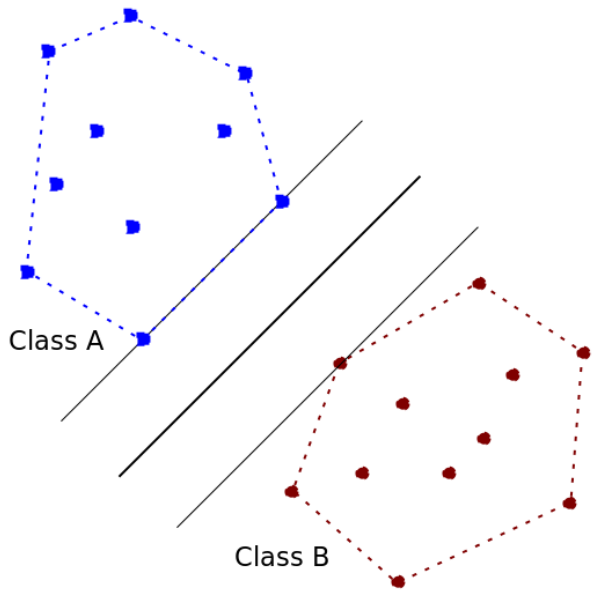}
\caption{Geometric interpretation of SVM with convex hull of each class.}
\label{fig:geometric_SVM}
\end{figure}

In the sensor network literature, a geometric interpretation of SVMs is often employed.  In the geometric interpretation, a convex hull is constructed for each class such that all the data is contained within the set, as shown in Figure \ref{fig:geometric_SVM}.  The separating hyperplane can then be reduced to finding the closest points between the two convex hulls, and selecting a hyperplane which bisects and is orthogonal to the line connecting these two points \cite{bennett2000duality}.

\textbf{Classification.}  Determining the equation for the optimal hyperplane involves solving a convex optimization problem, where the dual of the optimization problem is often used to allow for easy use of kernels.  Given a training set consisting of $N$ examples, $(\mathbf{x_i}, y_i)$, Lagrange variables $\alpha_i$ are determined by solving the optimization problem

\begin{equation}
\begin{aligned}
& \underset{\mathbf{\alpha}}{\text{maximize }} & & -\frac{1}{2}\sum_{i,j=1}^N \alpha_i \alpha_j y_i y_j k(\mathbf{x_i}, \mathbf{x_j}) + \sum_{i=1}^N y_i \alpha_i \\
& \text{subject to} & & \sum_i^N \alpha_i y_i = 0 \\
&                   & & \alpha_i \in [0,C] 
\end{aligned}
\end{equation}

where C is a parameter representing the tradeoff between maximizing the margin and minimizing the training error.  $k(\bullet, \bullet)$ is a kernel function which maps the input into a (possibly) nonlinear feature space.  Finally, a classification function can be constructed

\begin{equation}
f(\mathbf{x}) = \sum_{i=1}^N \alpha_i k(\mathbf{x_i}, \mathbf{x}) + b
\end{equation}

where $b$ is calculated by selecting a suitable training example from each class

\begin{equation}
  \begin{aligned}
  & b = -\frac{1}{2} \sum_{i=1}^N \alpha_i k(\mathbf{x_i}, \mathbf{x_r} + \mathbf{x_s}) \\
  & \text{where } \alpha_r, \alpha_s > 0, y_r = -1, y_s = 1
  \end{aligned}
\end{equation}

Novel examples can be classified from the sign of the classification function

\begin{equation}
  y = 
    \begin{cases}
      1 & \quad \text{if } f(\mathbf{x}) \ge 0 \\
      -1 & \quad \text{if } f(\mathbf{x}) \le 0 \\
    \end{cases}
\end{equation}

\textbf{Regression.}  To create a support vector regression (SVR) machine, the problem formulation is modified to minimize the distance between training examples and the optimal hyperplane \cite{smola2004-tutorial_svr}.  A soft margin of distance $\epsilon$ is set, such that training examples which fall within this margin have a cost of zero.  The reformulation of the problem results in the following dual optimization problem with dual variables $\alpha_i, \alpha_i^*$

\begin{equation}
\begin{aligned}
  & \underset{\alpha_i, \alpha_i^*}{\text{maximize }} & & -\frac{1}{2}\sum_{i,j=1}^N (\alpha_i - \alpha_i^*) (\alpha_i - \alpha_i^*) k(\mathbf{x_i}, \mathbf{x_j}) \\
  & & & -\epsilon \sum_{i=1}^N (\alpha_i + \alpha_i^*) + \sum_{i=1}^N y_i (\alpha_i - \alpha_i^*) \\
  & \text{subject to} & & \sum_i^N (\alpha_i - \alpha_i^*) = 0 \\
  &                   & & \alpha_i, \alpha_i^* \in [0,C] 
\end{aligned}
\end{equation}

The regression function is given by

\begin{equation}
f(\mathbf{x}) = \sum_{i=1}^N (\alpha_i - \alpha_i^*) k(\mathbf{x_i}, \mathbf{x}) + b
\end{equation}

where $b$ is calculated from suitable training examples by

\begin{equation}
\begin{aligned}
  b = y_i - \sum_{j=1}^N (\alpha_j - \alpha_j^*) k(\mathbf{x_j}, \mathbf{x_i}) - \epsilon & & \text{for $\alpha_i \in (0,C)$} \\
  b = y_i - \sum_{j=1}^N (\alpha_j - \alpha_j^*) k(\mathbf{x_j}, \mathbf{x_i}) + \epsilon & & \text{for $\alpha_i^* \in (0,C)$} \\
\end{aligned}
\end{equation}

\textbf{Kernels.}  Kernels allow for nonlinear regression and decision boundaries by projecting the input space into a nonlinear feature space.  Several admissible kernels are given in \cite{smola2004-tutorial_svr}, which include

\begin{equation}
\begin{aligned}
  k(\mathbf{x}, \mathbf{x'}) &= \mathbf{x}^T \mathbf{x'} & & \text{linear kernel} \\
  k(\mathbf{x}, \mathbf{x'}) &= (\mathbf{x}^T \mathbf{x'})^p & & \text{homogenous polynomial kernel} \\
  k(\mathbf{x}, \mathbf{x'}) &= (\mathbf{x}^T \mathbf{x'} + c)^p & & \text{inhomogenous polynomial kernel} \\
  k(\mathbf{x}, \mathbf{x'}) &= tanh(\rho \mathbf{x}^T \mathbf{x'} + \nu) & & \text{hyperbolic tangent kernel} \\
  k(\mathbf{x}, \mathbf{x'}) &= e^{ \left( -\frac{\| \mathbf{x} - \mathbf{x'} \|^2}{2 \sigma^2} \right)} & & \text{Gaussian radial basis function} \\
  k(\mathbf{x}, \mathbf{x'}) &= e^{ \left( -\frac{\| \mathbf{x} - \mathbf{x'} \|}{2 \sigma^2} \right)} & & \text{exponential radial basis function} \\
\end{aligned}
\end{equation}

In addition to allowing for nonlinear functions and decision boundaries, kernels can also be selected which provide bounded support over the domain, which has been exploited in several of the approaches used in the sensor network literature.

\subsection{Bayesian Models}

\begin{figure}[!htb]
\centering
\includegraphics[width=0.5\textwidth]{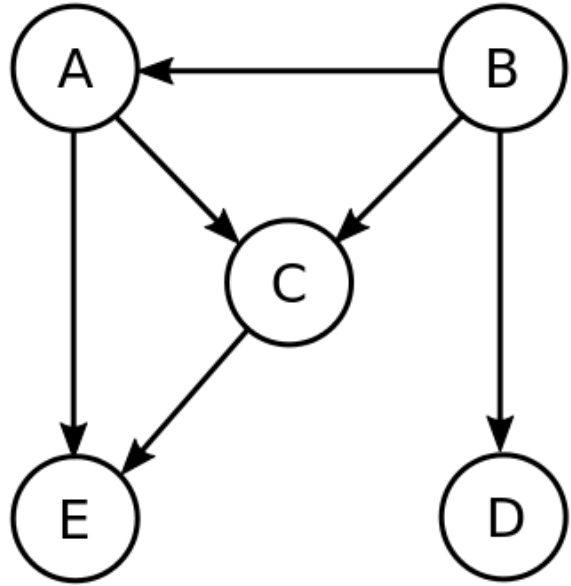}
\caption{Bayes model.}
\label{fig:bayes_model}
\end{figure}

Bayesian networks are probabilistic graphical models which represents a probability distribution as a directed acyclic graph:  each random variable in the distribution is represented as a node in a graph, and conditional dependencies are represented as directed edges in the graph.  Figure \ref{fig:bayes_model} shows an example Bayesian network over five random variables.

The structure of the graphical model represents a factorization of the joint distribution, i.e.

\begin{equation}
  P(X_1,X_2,...X_N) = \prod_{i=1}^N P \left( X_i | Pa(X_i) \right)
\end{equation}

where $Pa(X_i)$ are the parents of variable $X_i$.
\subsubsection{Belief Propagation}

Inference in graphical models involve determining the probability distribution or most likely assignment of variables given a set of observations.  Belief propagation is one approach based on message passing between nodes in the graph \cite{murphy1999loopy}.  The goal of belief propagation is for each node to compute the belief that a variable is assigned a specific value, given evidence, i.e., $BEL(x) = P(X = x|E)$.  The belief can be expressed as

\begin{equation}
   BEL(x) = \alpha \lambda(x) \pi(x)
\end{equation}

which consists of the combination of messages received from the node's children, 

\begin{equation}
  \lambda^{(t)}(x) = \lambda_X(x) \prod_j \lambda^{(t)}_{Y_j}(x)
\end{equation}

and messages received from its parents

\begin{equation}
  \pi^{(t)}(x) = \sum_u P(X = x | U = u) \prod_k \pi^{(t)}_X (u_k)
\end{equation}

At each iteration, a node passes the message to its parent $U_i$

\begin{equation}
  \lambda^{(t+1)}_X(u_i) = \alpha \sum_x \lambda^{(t)}(x) \sum_{u_k; k \neq i} P(x | u) \prod_{k \neq i} \pi^(t)_X(u_k)
\end{equation}

and the message to its child, $Y_j$

\begin{equation}
  \pi^{(t+1)}_{Y_j}(x) = \alpha \pi^{(t)}(x) \lambda_X(x) \prod_{k \neq j} \lambda^{(t})_{Y_k}(x)
\end{equation}

Message passing continues until belief updates are below a certain threshold for all variables.  This algorithm is used extensively in the sensor network literature, as the messages passed between neighboring nodes require only communication with immediate neighbors.

\subsubsection{Na\"{i}ve Bayes Classifier}
\begin{figure}[!htb]
\centering
\includegraphics[width=0.5\textwidth]{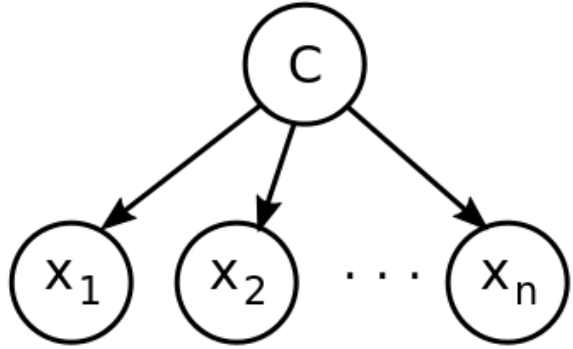}
\caption{Na\"{i}ve Bayes model.}
\label{fig:naive_bayes_model}
\end{figure}

Na\"{i}ve Bayes models are simple Bayesian models where observations are conditionally independent given a particular class, shown in Figure \ref{fig:naive_bayes_model}.  The simplicity of these models have made them very attractive for classification tasks.  The probability of a particular class, $C$, given an observation of $n$ features, $\mathbf{x}$, is proportional to the product of each feature conditioned on the class

\begin{equation}
  p(C | \mathbf{x}) \propto p(C) \prod_{i=1}^n p(x_i | C)
\end{equation}

\subsection{Markov Network}

Markov networks, also referred to as Markov random fields, are graphical models similar to Bayesian networks, except that the probability distribution is represented using a undirected graph.  The probability distribution can be expressed as a product of factors, $\phi_i(\mathbf{X})$, 

\begin{equation}
  P(\mathbf{X}) = \frac{1}{Z} \prod_{i=1}^N \psi_i(\mathbf{X})
\end{equation}

where each factor represents the affinities between pairwise variables (nodes connected by an edge) or clique in the graph.  The partition function, $Z$ is a normalizing factor to ensure the product results in a valid probability distribution.  As with Bayesian networks, belief propagation can be utilized to efficiently determine the joint probability distribution given observations.  The algorithm simply needs to be modified such that messages are passed between clusters, after summarizing messages from neighboring clusters (which share one or more common variables).

\subsection{Mixture Models}

Mixtures models are probabilistic models for representing the distribution of data \cite{russell2009artificial}.  Mixture models consist of $K$ components, where each component is a parametric probability distribution (e.g., Gaussian), and each component has a corresponding mixture weight, i.e. the probability of the particular component, $P(c)$.  The probability of a data point, $\mathbf{x}$, being generated by the mixture model is simply the the sum of the probability of the data point being generated by the individual components, weighted by the mixture weights

\begin{equation}
  P(\mathbf{x}) = \sum_{i=1}^K P(C_i) P(\textbf{x} | C_i)
\end{equation}

Gaussian mixture models (GMMs), where multivariate Gaussians are used for each component, have been used for classification though soft assignment to each category \cite{mannini2010-activity,allen2006-gmm}, or used to detect anomolies or outliers \cite{worden2003-experiment1,sohn2002statistical} based on the likelihood of an observation being generated by the mixture.  

\subsection{Dynamic Bayesian Network}

\begin{figure}[!htb]
\centering
\includegraphics[width=0.5\textwidth]{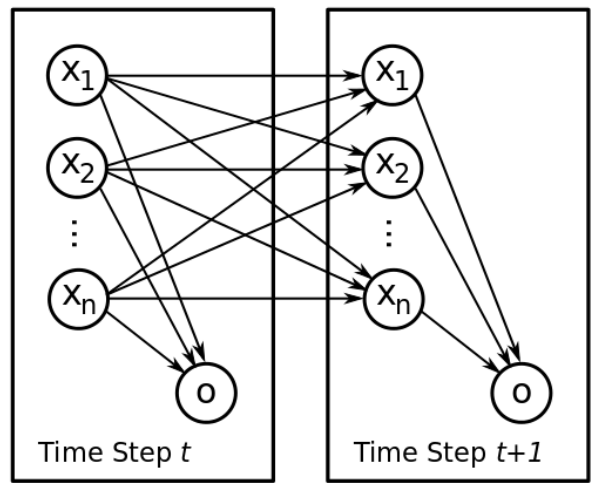}
\caption{Dynamic Bayes network.}
\label{fig:dynamic_bayes_network}
\end{figure}

A dynamic Bayesian network (DBN) is a class of graphical models which are useful for modeling sequential or temporal data \cite{russell2009artificial}.  DBMs represent a first-order Markov process, where the state variables at a particular time step are only dependent on the state variables in the previous time step, and observation variables are dependent on the current state variables.  Figure \ref{fig:dynamic_bayes_network} shows a dynamic Bayes model at two time steps, consisting of state variables, $x_1, x_2, ... x_n$ and an observation variable, $o$.  The model is represented as a transition model which follows a Markov assumption, $P(\mathbf{X}^t | \mathbf{X}^{t-1})$, and an observation model, $P(\mathbf{O}^t | \mathbf{X}^t)$ \cite{koller2009probabilistic}.

\subsubsection{Hidden Markov Models.}  Hidden Markov models (HMMs) are a special case of DBNs which consist of a single discrete, hidden state variable and observations conditioned on the current hidden state \cite{rabiner1986introduction}.  HMMs may be build for symbolic (discrete) or continuous valued observations; in the latter case, observations are assumed to be drawn from some probability density, such as a Gaussian or mixture of Gaussians.  Classification using HMMs involves training a model for each of the target classes, and selecting the class corresponding to the model with the highest likelihood of the sequence of observations.


\section{Overview of Common Features}\label{sec:appendix_features}

The major focus of many of the papers referenced in Section \ref{sec:ml_in_materials} is the features extracted from sensors.  As most of the data in the applications described are temporal in nature, there is a high amount of overlap in the features extracted and used.  However, there have also been several features of note developed and used for one of the specific domains, for example, transmissibility features in structural health monitoring.  Most of the machine learning tools used require a single vector of features to be provided as input, rather than a sequence of measured values.  Thus, it is common to calculate a set of features by summarizing the properties of measurements over a sliding window (i.e., measurement frame) of relatively short duration.

Features can be broadly categories on how the data points in a frame are processed.  Time domain features, perform calculations directly on the data points.  Frequency domain features first perform spectral decomposition, such as a Fourier or wavelet transform, and compute features from the resulting spectrum.  Model based features rely on fitting measurements to some simple model (e.g., autoregressive model), from which features can be extracted based on the parameters of the model or the residual of the measurement.

\subsection{Time Domain Features}

Time domain features summarize a measurement frame by performing operations directly on the measurement data in the frame.  

\textbf{Statistical Moments.}  Statistical features have been extracted from windows of measurements in many of the applications described.  Statistical moments (e.g., mean, standard deviation, skewness, kurtosis, etc.) are calculated from measurement windows.  The \emph{m}-th moment is calculated by

\begin{equation}
  \begin{aligned}
  \mu_1 &= \frac{1}{N} \sum_i^N x \\
  \mu_m &= \frac{1}{N} \sum_i^N (x - \mu_1)^m
  \end{aligned}
\end{equation}

\textbf{Correlation Coefficients.}  The correlation between two elements or axes of measurements is also used

\begin{equation}
  corr(x,y) = \frac{\frac{1}{N} \sum_i^N (x_i - \mu_x) (y_i - \mu_y)}{\sigma_x \sigma_y}
\end{equation}

The correlation coefficients are easily extracted by calculating the data covariance matrix.

In addition, minimum and maximum values of windows, as well as the sum of the difference between ordered pairs of peaks (i.e., minmax sum), are common features.

\textbf{Time Delay.}  Certain signals, such as those generated by impacts, are limited in duration.  This allows for various features based on the amount of duration between the onset of the signal and various aspects of the signal.  For example, for detecting impacts in aircraft composite panels, the time at the maximum and minimum value of the both the raw signal and the envelope of the signal, as well as the beginning and end of each envelope \cite{haywood2005-impact_monitor, staszewski2000-impact_detection}.  In this investigation, sensor measurements are synchronized, and the onset time of each signal from 12 accelerometers was also used as a feature.  

\subsection{Frequency Domain Features}

Frequency domain features are extracted from signals which have undergone a time-frequency transform, such as a Short Time Fourier Transform (STFT) or Discrete Wavelet Transforms (DWT) \cite{mannini2010-physical_activity}.  Frequency components provide information regarding the amount of energy present at different frequencies in the signal, allowing for a more intuitive representation of periodic signals.  The spectral components calculated by the transforms may be used as features directly, or features may be calculated from the spectrum.

\textbf{Modal Features.}  Modal features are useful for detecting defects in structures \cite{kessler2007-patternrecognition, farrar2007-shm}.  Vibrations can be measured using a piezoelectric transducer or accelerometer, and exited from external sources, or directly from a transducer.  The vibrational behavior of structural components is not constant across all frequencies; rather, structures tend to vibrate according to a set of vibration modes, which gives rise to resonant frequencies associated with the structure.  Under normal operating conditions, a structure will demonstrate peaks in the energy spectrum at a set of fixed frequencies, representing the frequencies of the modes of vibration in the structure.  In the presence of a defect, the frequencies at which these peaks will shift.  While useful for detecting defects, they are less useful for localizing or determining the extent of a defect.

\textbf{Transmissibility.}  Transmissibility describes the propagation of a signal between two points on a structure, and are useful for monitoring the natural frequencies or modalities of the structure.  Transmissibility is a feature commonly used in structural health monitoring, due to its sensitivity to changes in the stiffness of a component \cite{montalvao2006-review,lang2011transmissibility}.  The transmissibility between two points is defined as the ratio of the spectra measured at two points, $i$ and $j$

\begin{equation}
  T(\omega) = \frac{A_j(\omega)}{A_i(\omega)}
\end{equation}

Transmissibility can be measured using piezoelectric elements \cite{worden2007-ml_shm}.  The required exitation needs to be over a range of frequencies.  Exitation can either be generated from an external source, or generated from one transducer and measured from another.  As transmissibility measurements are between multiple measurement points, they provide more information than simply detecting resonant frequencies, and can be useful for localization of defects \cite{montalvao2006-review, lang2011transmissibility}.

\textbf{DC Component.}  The DC component of the signal, which can be obtained directly from the STFT, can be useful for discrimination purposes.

\textbf{Energy.}  The spectral energy can be calculated as the sum of the squared spectrogram coefficients.  Omitting the DC component, the energy of a spectrogram may help assess the intensity of a signal \cite{mannini2011-accelerometry}.  Integrating the real and imaginary components separately may also be reasonable in situations where the time delay between multiple sensors can provide useful information, such as in impact detection \cite{haywood2005-impact_monitor, staszewski2000-impact_detection}.

\textbf{Entropy.}  The spectral entropy can be calculated from the normalized energy of the spectrogram.  The spectral energy is normalized for each bin, $p_i$, after which the spectral entropy can be calculated

\begin{equation}
  H = - \sum_i^N p_i ln p_i
\end{equation}

\subsection{Model Based Features}

Model based features involve calculating the coefficients of a model (e.g., autoregressive model) fit to the data in a measurement frame.  Models provide several benefits.  For example, dimensionality of the data can be reduced (e.g., principle components), and linear models (e.g., autoregressive models) can be exploited to detect nonlinearities in a signal.  

\textbf{Principal Components Analysis.}  Given a set of measurement vectors which demonstrate correlation among the elements in the vectors, principle components analysis (PCA) determines an orthogonal transformation which results in linearly uncorrelated values \cite{de2003principal}.  The transformation can be represented as a matrix $\mathbf{P}$, which transforms a measurement vector $\mathbf{x}$, to a vector of principal components, $mathbf{y} = \mathbf{P} \mathbf{x}$.  One approach to computing the transformation matrix is to perform a singular value decomposition (SVD) of a set of training data.  SVD has the advantage of sorting the principle components by the amount of variance of each component in the data.  The principle components with the highest variance can be retained, while those with low variance discarded as statistical noise, thus providing a means of dimensionality reduction.  The individual principal components may then be used as a feature set.

\textbf{Independent Components Analysis.}  Independent components analysis (ICA) can be used to describe a time series as a linear combination of components, where each component is statistically independent, and observe a non-Gaussian distribution \cite{song2006-shm_ica_svm}.  The set of measurements is represented by $\mathbf{X} = \mathbf{A} \mathbf{S}$, where $\mathbf{A}$ is a mixing matrix, and $\mathbf{S}$ is the set of components.  Given a set of measurements, the components can be calculated by $\mathbf{S} = \mathbf{A}^{-1} \mathbf{X} = \mathbf{W} \mathbf{X}$ (where $\mathbf{W} = \mathbf{A}^{-1}$).  $\mathbf{W}$ is selected such that the non-Gaussianity of the resulting components are maximized.  Non-Gaussianity is measured by different methods, such as kurtosis, negentropy, or a nonlinear function, such as hyperbolic tangent.  Future signals can then be represented as a vector of the mixing coefficients of each component.

\textbf{Autoregressive Model.}  Autoregressive (AR) models construct an input signal from $r$ autoregressive terms as

\begin{equation}
  x(t) = \sum_{i=1}^r \alpha_i x(t-i) + e_x(t)
\end{equation}

where $e_x(t)$ is the error between the measured values and prediction from the AR model \cite{sohn2002statistical, montalvao2006-review}.  Assuming $e_x(t)$ is generated by some unknown input, a second prediction stage can be generated to model this error, referred to as an autoregressive with exogenous inputs (AR-ARX) model

\begin{equation}
  x(t) = \sum_{i=1}^p \alpha_i x(t-i) + \sum{j=1}^q \beta_j e_x(t-j) + \epsilon_x(t)
\end{equation}

where $\epsilon_x(t)$ is the residual error.  The autoregressive terms, $\alpha_i$ and $\beta_j$, are then used as features for a measurement frame.


\bibliographystyle{plain}
\bibliography{prelim,shm,morphable_aircraft,gait_anomaly,robot_skin}

\end{document}